%% file: main.tex
\documentclass{article}
\usepackage{log_2022}         % for camera-ready version
\usepackage{booktabs}           % professional-quality tables
\usepackage{multirow}           % tabular cells spanning multiple rows
\usepackage{amsfonts}           % blackboard math symbols
\usepackage{graphicx}           % figures
\usepackage{duckuments}         % sample images

\usepackage[utf8]{inputenc} % allow utf-8 input
\usepackage[T1]{fontenc}    % use 8-bit T1 fonts
\usepackage{url}            % simple URL typesetting
\usepackage{nicefrac}       % compact symbols for 1/2, etc.
\usepackage{microtype}      % microtypography
\usepackage{xcolor}         % colors
\usepackage{amsmath}
\usepackage{amssymb}
\usepackage{mathtools}
\usepackage{amsthm}

\usepackage{diagbox}
\usepackage{nicematrix}
\usepackage{subfigure}
\usepackage{booktabs}
\usepackage{caption}
\theoremstyle{plain}
\newtheorem{theorem}{Theorem}[section]
\newtheorem{proposition}[theorem]{Proposition}

\theoremstyle{definition}
\newtheorem{definition}[theorem]{Definition}

\theoremstyle{remark}

\usepackage{url}
\usepackage{xspace}
\usepackage{booktabs}
\usepackage{longtable}
\usepackage{adjustbox}
\usepackage{amsthm}
\usepackage{xcolor}
\usepackage[ruled,lined,noresetcount]{algorithm2e}
\usepackage{algpseudocode}
\usepackage{wrapfig,lipsum,booktabs}

\newcommand{\name}{GARNET\xspace}

\usepackage[numbers,compress,sort]{natbib}

\title{\name: Reduced-Rank Topology Learning for Robust and Scalable Graph Neural Networks}
\author[C. Deng et al.]{%
Chenhui Deng\\
\institute{Cornell University}\\
\email{cd574@cornell.edu}\And
Xiuyu Li\\
\institute{UC Berkeley}\\
\email{xiuyu@berkeley.edu}\AND
Zhuo Feng\\
\institute{Stevens Institute of Technology}\\
\email{zfeng12@stevens.edu}\And
Zhiru Zhang\\
\institute{Cornell University}\\
\email{zhiruz@cornell.edu}
}

\begin{document}

\maketitle

\input{abstract}

\input{introduction}

\input{background}

\input{method}

\input{experiments}

\input{conclusions}

%\section*{Author Contributions}
%Authors of accepted papers are \emph{encouraged} to include a statement that declares the individual contribution of every author, especially when there are co-authors that made equal contributions to the research.
%You may adopt the \href{https://credit.niso.org/}{Contributor Roles Taxonomy (CRediT)} methodology for attributing contributions.
%Do not include this section in the version for blind review.
%This section does not count towards the page limit.

\input{acknowledgements.tex}

% For natbib users:
\bibliographystyle{unsrtnat}
\bibliography{reference}
% For bibLaTeX users:
% \printbibliography

\newpage
\appendix

\input{appendix}

\end{document}

%% file: abstract.tex
\begin{abstract}
Graph neural networks (GNNs) have been increasingly deployed in various applications that involve learning on non-Euclidean data. However, recent studies show that GNNs are vulnerable to graph adversarial attacks. Although there are several defense methods to improve GNN robustness by eliminating adversarial components, they may also impair the underlying clean graph structure that contributes to GNN training. In addition, few of those defense models can scale to large graphs due to their high computational complexity and memory usage. In this paper, we propose GARNET\footnote{Source code of GARNET is freely available at: \href{https://github.com/cornell-zhang/GARNET}{github.com/cornell-zhang/GARNET}.}, a scalable spectral method to boost the adversarial robustness of GNN models. GARNET first leverages weighted spectral embedding to construct a base graph, which is not only resistant to adversarial attacks but also contains critical (clean) graph structure for GNN training. Next, GARNET further refines the base graph by pruning additional uncritical edges based on probabilistic graphical model. GARNET has been evaluated on various datasets, including a large graph with millions of nodes. Our extensive experiment results show that GARNET achieves adversarial accuracy improvement and runtime speedup over state-of-the-art GNN (defense) models by up to $10.23\%$ and $14.7\times$, respectively.
\end{abstract}

%% file: introduction.tex
\section{Introduction}

Recent years have witnessed a surge of interest in graph neural networks (GNNs), which incorporate both graph structure and node attributes to produce low-dimensional embedding vectors that maximally preserve graph structural information~\citep{hamilton2020graph}. GNNs have achieved promising results in various real-world applications, such as recommendation systems~\citep{ying2018graph}, 
self-driving car~\citep{casas2020spagnn}, 
%protein structure predictions~\citep{senior2020improved}, 
and chip placements~\citep{mirhoseini2021graph}. However, recent studies have shown that adversarial attacks on graph structure accomplished by inserting, deleting, or rewiring edges in an unnoticeable way, can easily fool the GNN models and drastically degrade their accuracy in downstream tasks (e.g., node classification)~\citep{zugner2018adversarial, zugner2019adversarial}.

In literature, one of the most effective ways to defend GNNs is to purify the graph by removing adversarial graph structures.
%there are a few attempts to defend GNNs against graph adversarial attacks. 
%One effective way is to purify the graph by removing adversarial edges. 
\citet{entezari2020all} observe that adversarial attacks mainly affect high-rank graph properties; thus they propose to first construct a low-rank graph by performing truncated singular value decomposition (TSVD) on the graph adjacency matrix, which can then be exploited for training a robust GNN model. Later, \citet{jin2020graph} propose Pro-GNN to jointly learn a new graph and a robust GNN model with the low-rank constraints imposed by the graph structure. While prior methods using low-rank approximation largely eliminate adversarial components in the graph spectrum, they involve dense adjacency matrices during GNN training, leading to a much higher time/space complexity and prohibiting their applications in  large-scale graph learning tasks.

In addition, due to the high computational cost of TSVD, existing low-rank based methods can only preserve top $r$ singular components (e.g., $r=50$). Consequently, as shown in Figure \ref{figure:rank_cp}, these methods may lose a wide range of clean graph spectrum that corresponds to %\fixme{the important clean graph structure} 
%\zz{important structures of the clean graph} 
important structures of the clean graph in the spatial domain. This is confirmed in Figure \ref{figure:clean}, where the clean accuracy of the TSVD-based method largely increases when preserving more spectral information via increasing the graph rank $r$. In other words, prior low-rank approximation methods eliminate high-rank adversarial components at the cost of inevitably impairing the important (clean) graph structure, which degrades the overall quality of the reconstructed graph and therefore limits the performance of GNN training.

\input{figure/intro_fig}

In this work, we propose \name, a novel spectral approach to learning the 
%\zf{base graph} 
underlying clean graph
topology of an adversarial graph via combining spectral embedding with probabilistic graphical model (PGM), where the learned graph structure encodes the conditional dependence   among low-dimensional node representations (spectral embedding vectors)~\citep{dong2019learning}.
%Specifically, we propose \name, a spectral method for constructing GNN models that are robust to graph adversarial attacks 
%for both homophilic and heterophilic graphs, 
%with the aid of graphical models. In addition, \name scales comfortably to large graphs due to its nearly-linear algorithm complexity. 
More concretely, given an adversarial graph, \name first constructs a   base graph topology by leveraging weighted spectral embeddings that are resistant to adversarial attacks, which is followed by an effective and efficient graph refinement scheme for pruning noncritical edges in the base graph by exploiting PGM. 
%As a result, \name produces a graph preserves the critical graph structure for GNN training and recovers most of clean graph spectrum, as shown in Figures \ref{figure:rank_cp} and \ref{figure:clean}.

By recovering the clean graph structure, Figures \ref{figure:rank_cp} and \ref{figure:singular} show that the adversarial graph purified by \name largely restores the rank of the underlying clean graph. Thus, \name can be viewed as a reduced-rank topology learning approach that slightly reduces the rank of the input adversarial graph, which is fundamentally different from the prior low-rank based defense methods (e.g., TSVD and ProGNN).
%\zz{also need to bring up topology learning} 
Moreover, \name scales comfortably to large graphs due to its nearly-linear algorithm complexity, and produces a sparse yet high-quality graph that improves GNN robustness without involving any dense adjacency matrices during GNN training. 
%\zz{counterintuitive -- i suppose the input graph is already sparse? producing a sparse graph sounds weird. does ProGNN produce a dense graph? I guess not?}
%performs spectral embedding with a few dominant singular components of the adversarial adjacency matrix. The spectral embeddings are then used to construct a nearest neighbor graph, followed by a novel sparsification scheme, which removes unimportant edges based on PGM. As a result, \name produces a sparse yet high-quality graph that effectively mitigates the effects of adversarial attacks via ignoring highest rank components while preserving the key graph structure for GNN training. 
As a byproduct, unlike existing defense methods (e.g., ProGNN) that assume graphs to be homophilic, i.e, adjacent nodes in a graph tend to have similar attributes~\citep{zhu2020beyond}, \name does not have such an assumption and thus can protect GNNs against adversarial attacks on both homophilic and heterophilic graphs.
%The \stageb kernel aims to learn a polynomial graph filter whose coefficients are trainable; the learned graph filter can  adapt to the homophilic/heterophilic properties of the underlying graph and thus will work effectively for both homophilic and heterophilic graphs. 
%The \stagec stage will leverage the learned adaptive graph filter to guide the label propagation phase, which can further improve the adversarial accuracy by enhancing the quality of node representations.

We evaluate \name on both homophilic and heterophilic datasets
%two high-homophily datasets: Cora and Pubmed, as well as two low-homophily datasets: Chameleon and Squirrel, 
under strong graph adversarial attacks such as Nettack~\citep{zugner2018adversarial} and Metattack~\citep{zugner2019adversarial}. Moreover, we further show the nearly-linear scalability of our approach on the ogbn-products dataset that consists of millions of nodes~\citep{hu2020open}. Our experimental results indicate that \name largely improves both clean and adversarial accuracy over baselines in most cases. 
Our main technical contributions are summarized as follows:
\noindent $\bullet$ To our knowledge, we are the first to exploit spectral graph embedding and probabilistic graphical model for improving robustness of GNN models, which is achieved by learning a reduced-rank graph topology for recovering the underlying clean graph structure from the input adversarial graph.
%\vspace{-2pt}

% preserves critical structural information while ignoring those spectral components induced by adversarial attacks. 
%\noindent $\bullet$ We theoretically show that the edges, whose insertions to the graph mitigate large spectral embedding distortions, are critical and contribute to the maximum likelihood estimation for learning the reduced-rank graph structure in PGM. 
%\zz{combine with the first contrib?}
\noindent $\bullet$ By recovering the critical edges that contribute to maximum likelihood estimation in PGM 
%\zf{additional edges are introduced, so we should not use "preserving"?} 
while ignoring adversarial components, \name produces a high-quality graph on which  existing GNN models can be trained to achieve high accuracy. Our experimental results show that \name gains up to $10.23\%$ adversarial accuracy improvement over state-of-the-art defense baselines.

\noindent $\bullet$ Our proposed reduced-rank topology learning method has a nearly-linear  complexity in time/space and produces a sparse graph structure for scalable GNN training. This allows \name to  run up to $14.7\times$ faster than prior defense methods on popular data sets such as Cora and Squirrel. 
%\zz{the first line is about small graphs}
In addition, \name scales comfortably to very large graph data sets with millions of nodes, while prior defense methods run out of memory even on a graph with $20$k nodes.

%% file: figure/intro_fig.tex
%\begin{figure*}
%\centering
%\begin{minipage}{0.4\textwidth}
%  \centering
%  \includegraphics[width=\linewidth]{figure/rank_comp.pdf}
%  \vspace{-15pt}
  %\caption{Rank comparison of graphs with different perturbation ratios under Metattack on Cora. ``TSVD Graph'' and ``\name Graph'' denote the adversarial graph purified by TSVD and \name, respectively.}
%  \label{figure:rank_cp}
%\end{minipage}%
%\hspace{0.1\textwidth}
%\begin{minipage}{0.4\textwidth}
%  \centering
%  \includegraphics[width=\linewidth]{figure/clean_acc.pdf}
%  \vspace{-15pt}
  %\caption{Accuracy comparison on Cora with different $r$-rank approximations for GCN-TSVD.}
%  \label{figure:clean}
%\end{minipage}
%\caption{ff}
%\end{figure*}
%\vspace{-6pt}
\begin{figure*}[t!]% [hpbt] what you need
    \centering
     \subfigure[]{%
        \includegraphics[width=0.32\textwidth]{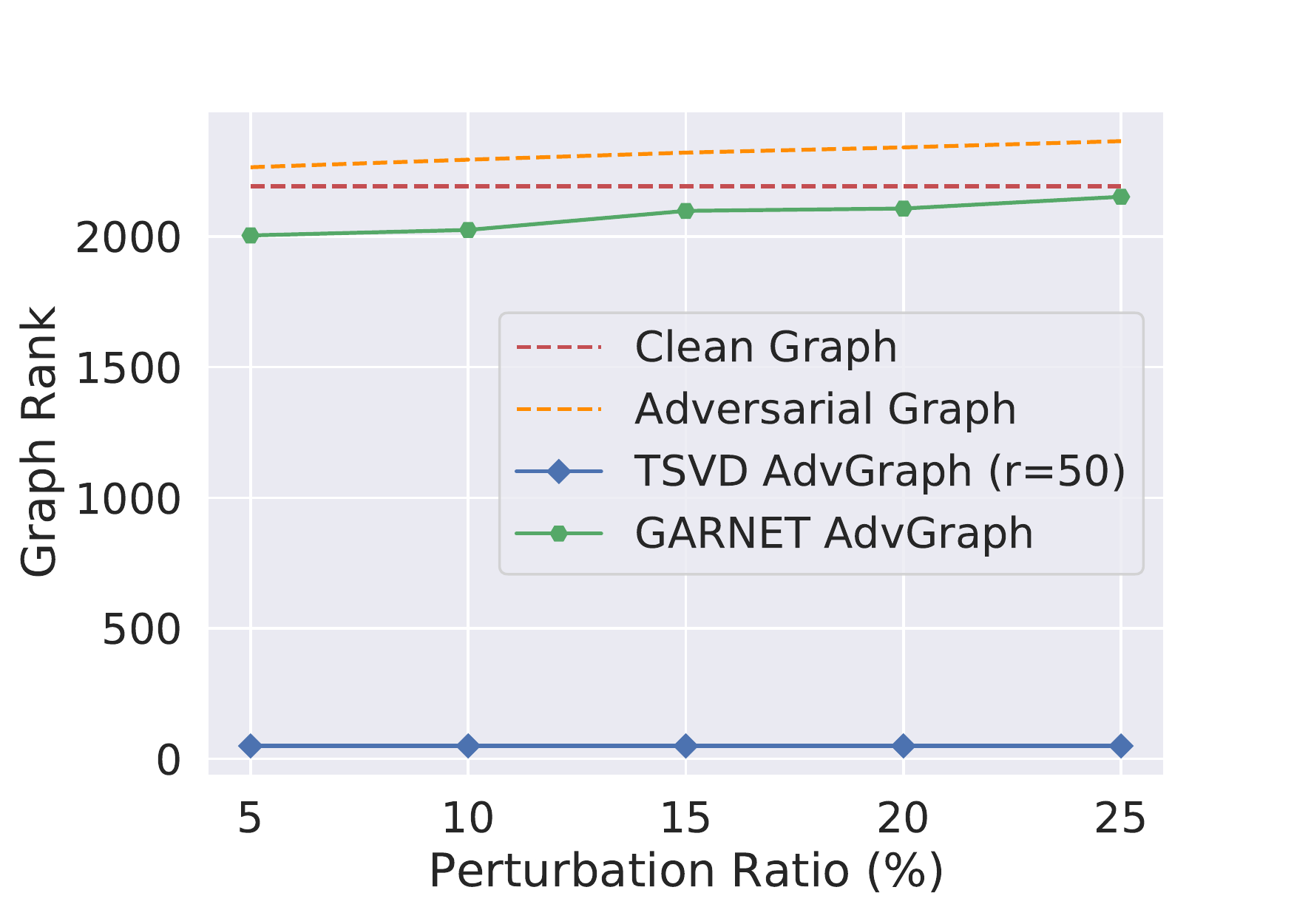}
        \label{figure:rank_cp}}
    %\hspace{0.1\textwidth}
    \subfigure[]{%{0.3\textwidth}
        \includegraphics[width=0.32\textwidth]{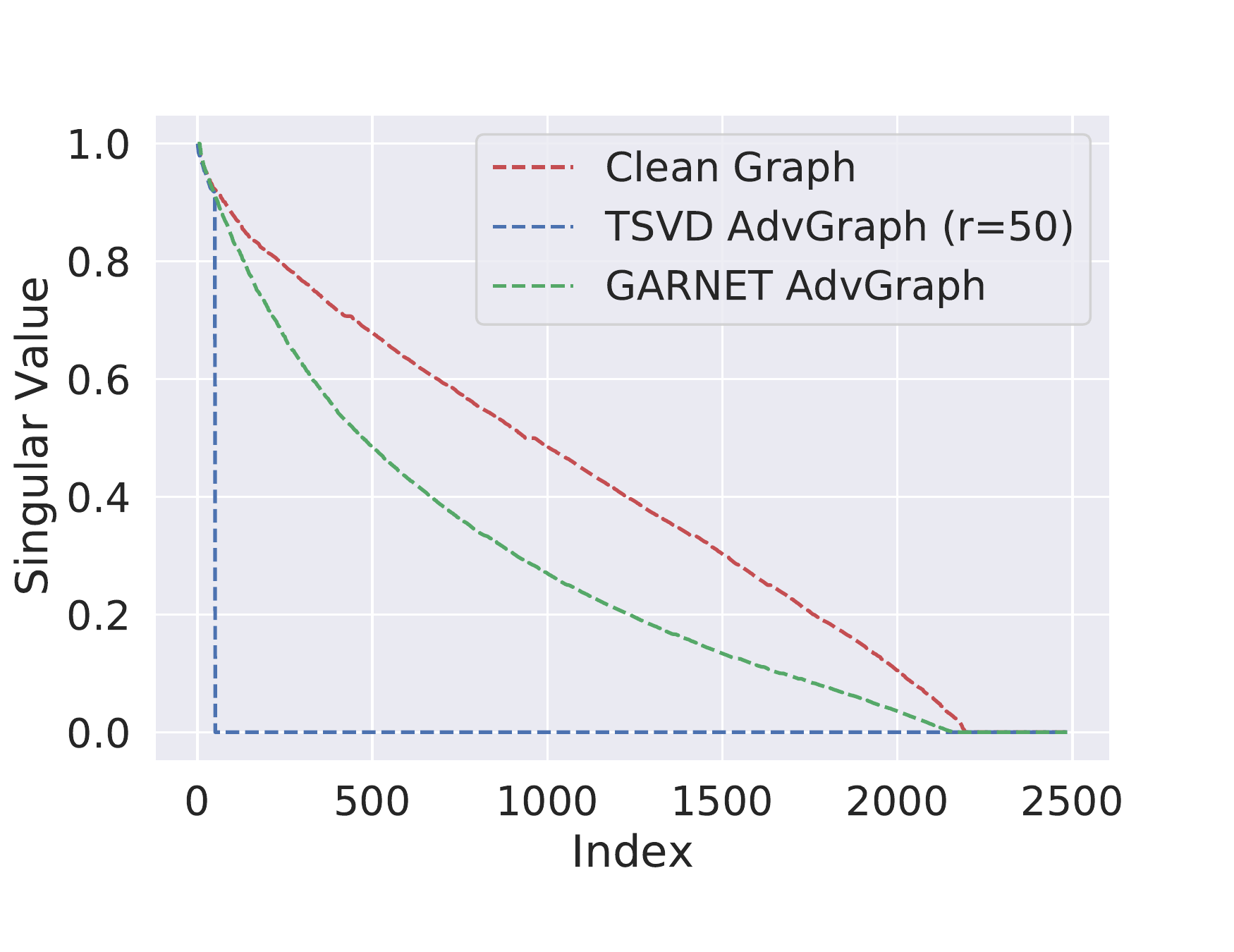}
        \label{figure:singular}}
     \subfigure[]{%{0.3\textwidth}
        \includegraphics[width=0.32\textwidth]{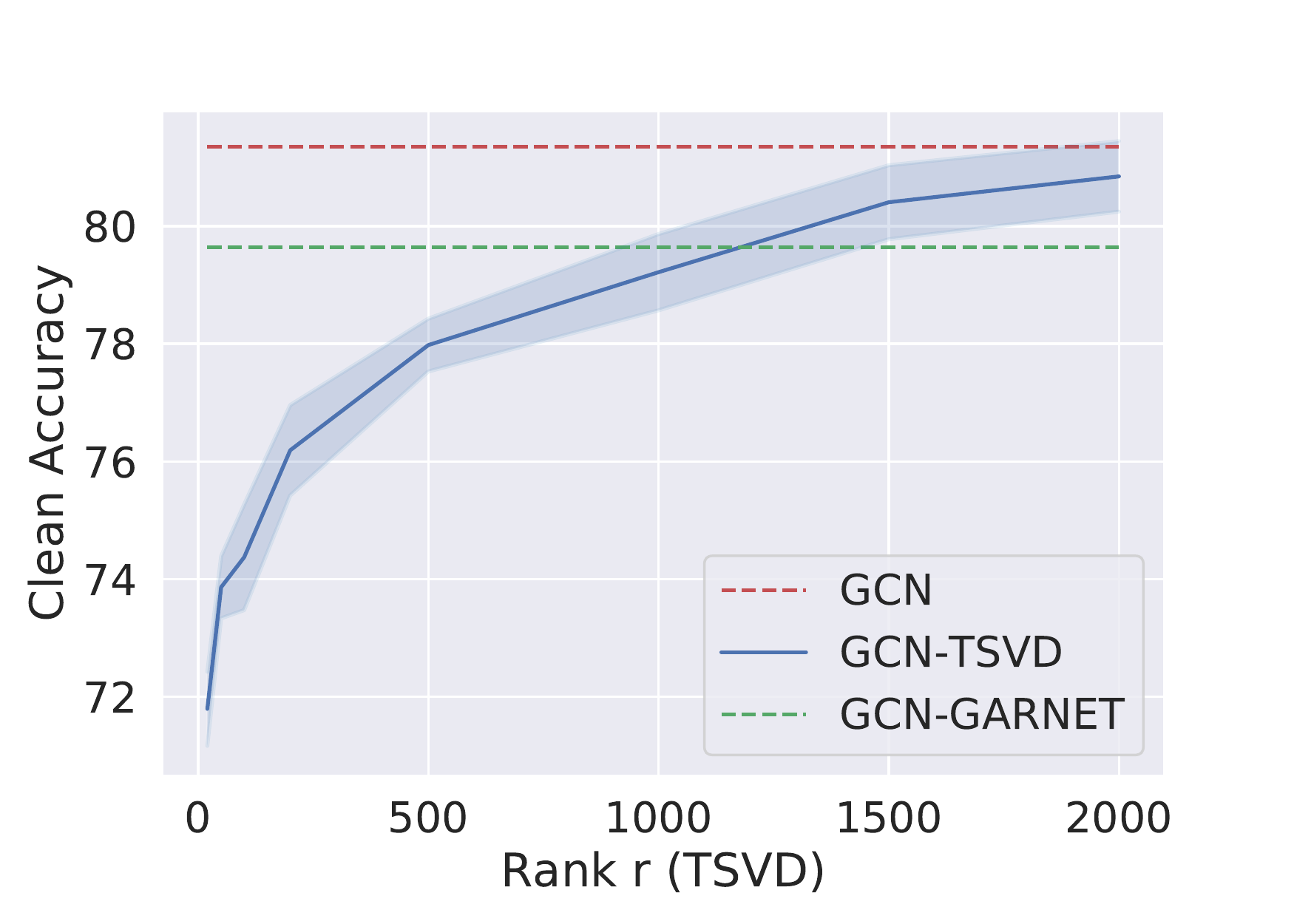}
        \label{figure:clean}}
    \vspace{-5pt}
    \caption{``TSVD AdvGraph'' and ``\name AdvGraph'' denote adversarial graphs purified by TSVD and \name, respectively. (a) Graph rank comparison on Cora under Metattack with different perturbation ratio. (b) Singular value comparison of different normalized graph adjacency matrices on Cora. (c) Accuracy $\pm$ std. of GCN-TSVD on Cora with different $r$-rank approximation via TSVD.}
    \vspace{-22pt}
    \label{figure:intro}
    %\vspace{-1.6cm}
\end{figure*}

%% file: background.tex
%\vspace{-10pt}
\section{Background}
%\vspace{-5pt}
\subsection{Undirected Probabilistic Graphical Models}
%\vspace{-5pt}
\label{pgm_work}
Consider an $n$-dimensional random vector $x$ that follows a multivariate Gaussian distribution  $x\sim N(0,\Sigma)$, %with probability density:
%\vspace{-5pt}
  %\begin{equation}
  %$$
   %   f(x)=\frac{\exp{\left(-\frac{1}{2}x^\top \Sigma^{-1}x\right)}}{(2\pi)^{n/2}\det(\Sigma)^{(1/2)}}\propto \det(\Theta)^{1/2}\exp{\left (-\frac{1}{2}x^\top \Theta x\right )},
      %\vspace{-5pt}
%      $$
  %\end{equation}
  %\vspace{-1pt}
 where $\Sigma=\mathbb{E}[xx^\top]\succ 0$ represents the covariance matrix, and $\Theta=\Sigma^{-1}$ represents the precision matrix (inverse covariance matrix). 
%  Consider each data point as a node in an underlying graph $\mathcal{G}$ and introduce   edges  to encode the conditional dependence relationship among random variables represented by nodes~\cite{dong2019learning}. 
 Given a data matrix $X \in R^{n \times d}$ that includes $d$ i.i.d (independent and identically distributed) samples $X=[x_1,...,x_d]$, where $x_i\sim N(0, \Sigma)$ has an $n$-dimensional Gaussian distribution with zero mean, the goal of probabilistic graphical models (PGM) is to learn a precision matrix $\Theta$ that corresponds to an undirected graph structure $\mathcal{G}$ for encoding the conditional dependence between variables of the observations on columns of $X$ ~\citep{banerjee2008model, friedman2008sparse}. Specifically, the classical graphical Lasso method aims at estimating a sparse   $\Theta$ through  maximum likelihood estimation (MLE) of $f(x)$ leveraging convex optimization \citep{friedman2008sparse}.
 In this work, we focus on one increasingly popular type of Gaussian graphical models, which is also known as attractive Gaussian Markov random fields (GMRFs). Attractive GMRFs restrict the precision matrix to be a Laplacian-like matrix $\Theta = L + \frac{I}{\sigma^{2}}$, where $L=D-A$ denotes the set of valid graph Laplacian matrices with $D$ and $A$ representing the diagonal degree matrix and adjacency matrix of the underlying undirected graph, respectively, $I$ denotes the identity matrix, and $\sigma^2$ is a constant denoting prior data variance. Similar to the graphical Lasso method \citep{friedman2008sparse}, recent  methods for estimating attractive GMRFs   leverage emerging graph signal processing (GSP) techniques  to   solve the following  convex  problem \citep{dong2016learning,egilmez2017graph,dong2019learning, kalofolias2019large, feng2021sgl}:
 %\vspace{-4pt}
\begin{equation}\label{lasso}
\max\limits_{\Theta}\log \det \Theta - \frac{1}{d}tr(XX^T\Theta) - \alpha \|\Theta\|_1
%\vspace{-3pt}
\end{equation}
%\vspace{-5pt}
%\vspace*{-5mm}
where $det(\cdot)$ and $tr(\cdot)$ denote the determinant and trace operators, respectively, $\alpha$ is a hyperparameter to control the regularization term. The first two terms together can be interpreted as log-likelihood under a GMRF. The last $\ell_1$ regularization term is to enforce $\Theta$ (and the corresponding graph) to be sparse. {If  $X$ is non-Gaussian},  Equation \ref{lasso} can be regarded as
Laplacian estimation based on minimizing the Bregman divergence between positive definite matrices   induced by the function $\Theta \mapsto -\log \det(\Theta)$ \citep{slawski2015estimation}.

%GNNs have received an increasing amount of attention in recent years due to its ability of learning on non-Euclidean (graph) data. In contrast to developing powerful GNN models on natural graph data, there is an active body of research focusing on adversarial attacks as well as defenses for GNNs. We summarize some of the recent efforts for graph adversarial attacks and defenses as follows.
%\vspace{-10pt}
\subsection{Graph Adversarial Attacks} 
%\vspace{-5pt}
Most existing  graph adversarial attacks aim at degrading the accuracy of GNN models by inserting/deleting edges in an unnoticeable way (e.g., maintaining node degree distribution)~\citep{sun2018adversarial}. The most popular graph adversarial attacks fall into the following two categories: (1) targeted attack, (2) non-targeted attack. The targeted attacks attempt to mislead a GNN model to produce a wrong prediction on a target sample (e.g., node), while the non-targeted attacks strive to degrade the overall accuracy of a GNN model for the whole graph data set. \citet{dai2018adversarial} first formulate the targeted attack as a combinatorial optimization problem and leverages reinforcement learning to insert/delete edges such that the target node is misclassified.  \citet{zugner2018adversarial} propose another targeted attack called Nettack, which produces an adversarial graph by maximizing the training loss of GNNs. \citet{zugner2019adversarial} further introduce Metattack, a non-targeted attack that treats the graph as a hyperparameter and uses meta-gradients to perturb the graph structure. It is worth noting that graph adversarial attacks have two different settings: poison (perturb a graph prior to GNN training) and evasion (perturb a graph after GNN training). As shown by ~\citet{zhu2021improving}, the poison setting is typically more challenging to defend, as it changes the graph structure that fools GNN training. Thus, 
%in this work, 
we aim to improve model robustness against attacks 
%against (non)targeted 
%and untargeted attacks 
under the poison setting.
%\vspace{-10pt}
\subsection{Graph Adversarial Defenses} 
%\vspace{-5pt}
To defend GNN against adversarial attacks, 
%\citet{wu2019adversarial} propose to modify the edge weights by computing the Jaccard similarity score per edge based on node attributes, such that the adversarial edges are likely to have small weights. Later, 
\citet{entezari2020all} first observe that Nettack, a strong targeted attack, only changes the high-rank information of the adjacency matrix. Thus, they propose to construct a low-rank graph by performing truncated SVD to undermine the effects of adversarial attacks. Later, \citet{jin2020graph} propose Pro-GNN that adopts a similar idea yet encourages nodes with similar attributes to be connected when jointly learning the low-rank graph and GNN model.
%jointly learns the low-rank graph and GNN model. 
Although those low-rank approximation based methods achieve state-of-the-art results on several datasets, they produce dense adjacency matrices that correspond to complete graphs, which would limit their applications for large graphs. Moreover, they only preserve a small region of the graph spectrum and thus may lose too much important information corresponding to the clean graph structure in the spatial domain, which limits the performance of GNN training. Recently, ~\citet{chang2021not} exploit Laplacian eigenpairs to guide GNN training, which produces a robust model with quadratic time complexity and is thus not scalable to large graphs. In addition to the aforementioned spectral-based defense methods, GCNJaccard~\cite{wu2019adversarial} and RS-GNN~\cite{dai2022towards} purify the adversarial graph by connecting nodes with similar attributes or same labels. However, those defense methods explicitly (or implicitly) assume the underlying graph to be homophilic, which results in rather poor performance when defending GNN models on heterophilic graphs.
%Another line of research strives to purify the adversarial graph by assigning edge weights. Specifically, 
%which is followed by \citet{zhang2020gnnguard} that propose GNNGuard to learn node attribute similarity score per edge through a trainable linear layer. Nonetheless, such approaches assume that nearby nodes should have similar attributes (i.e., graph homophily assumption), which is not valid for heterophilic graphs that have adjacent nodes with dissimilar attributes~\citep{zhu2020beyond}. 
In contrast to the prior arts, \name achieves highly robust yet scalable performance on both homophilic and heterophilic graphs under adversarial attacks by leveraging a novel graph purification scheme based on spectral embedding and graphical model.
% Moreover, it performs well on both homophilic and heterophilic graphs with the aid of adaptive graph filter.

%% file: method.tex
%\vspace{-10pt}
\section{The \name Approach}
%\vspace{-5pt}
\label{method}
\input{figure/overview}
 %The adjacency matrices of many real-world graphs (e.g., social network and biological network) are naturally low-rank and sparse, 
 %\zz{must use noun for a property -- rank and sparsity}, 
 %as the nodes typically tend to form communities and have a small number of neighbors~\citep{zhou2013learning}. As a result, graph adversarial attacks can be viewed as compromising
 %damaging \zz{compromising} 
 %these properties by inserting edges that connect nodes from different communities~\citep{zugner2018adversarial, zugner2019adversarial}. 
 %Similar to adversarial attacks on images, graph adversarial attacks typically perturb the graph structure in an unnoticeable way, i.e., $\|A_{clean} - A_{adv} \|_0 \leq \Delta$, where $A_{clean}$ and $A_{adv}$ are adjacency matrices of the clean and adversarial graphs, respectively, and $\Delta$ is a small constant as the perturbation budget. As a result
 Recently, \citet{entezari2020all} and \citet{jin2020graph} have shown that the well-known 
 %\zz{i would avoid using strongest without defining the metric; just use well-known} 
 graph adversarial attacks (e.g., Nettack and Metattack) are essentially high-rank attacks, which increase graph rank by enlarging the smallest singular values of adjacency matrix when perturbing the graph structure, while rest of the graph spectrum remains almost the same. 
 %We empirically confirm that the graph rank indeed increases under adversarial attacks in Appendix 
 %\ref{graph_rank}. 
 Consequently, a natural way for improving GNN robustness is to find the low-rank approximation of the adversarial adjacency matrix.
 %purify an adversarial graph by eliminating the high-rank components of its spectrum. 
 %\zz{is it a commonly-used term? low/high-rank info doesn't sound right. maybe high-rank singular components?}

%\textbf{Rank reduction via truncated SVD.} 
%\textbf{Graph purification via low-rank approximation.} To enhance GNN robustness, 
 %a graph that only preserves the lowest rank components of its spectrum to mitigate the effects of adversarial attacks, via performing truncated SVD on the adjacency matrix. 
 \textbf{Low-rank topology learning (prior work).} Given an adversarial 
 %graph and its 
 adjacency matrix $A_{adv} \in R^{n \times n}$, \citet{entezari2020all} propose to reconstruct a low-rank approximated adjacency matrix via performing TSVD: $\hat{A} = U {\Sigma} V^T$, where $\Sigma \in R^{r \times r}$ is a diagonal matrix consisting of $r$ largest singular values of $A_{adv}$. $U \in R^{n \times r}$ and $V \in R^{n \times r}$ contain the corresponding left and right singular vectors, respectively. As the largest singular values are hardly affected by graph adversarial attacks, the reconstructed 
 low-rank adjacency matrix 
 $\hat{A}$ is resistant to adversarial attacks.

However, due to the high computational cost of TSVD, $\hat{A}$ is typically computed by only using top $r$ largest singular values and their corresponding singular vectors, where $r$ is a relatively small number (e.g., $r=50$). Consequently, the rank of $\hat{A}$ is only $r=50$, which is two orders of magnitude smaller than the rank of the clean graph, as shown in Figure \ref{figure:rank_cp}. Since these low-rank methods are overly aggressive in reducing the graph rank, $\hat{A}$ may lose too much important spectral information corresponding to the clean graph structure. As shown in Figure \ref{figure:clean}, the clean accuracy of the TSVD-based method is largely improved by increasing the graph rank $r$, which indicates the low-rank graph obtained with a small $r$ loses the key graph structure contributing to GNN training. Note that the adversarial and clean graphs share most of the graph structure, as adversarial attacks perturb the clean graph in an unnoticeable way. Consequently, losing those important clean graph structures will also limit the performance of GNN on the adversarial graph.
%which limits the performance of GNN training on $\hat{A}$. 
%In addition, $\hat{A}$ is typically a dense matrix with $O(n^2)$ nonzero elements, which may result in prohibitively expensive storage as well as GNN training, where $n$ represents number of nodes. Consequently, existing low-rank based defense methods are not scalable to large graphs~\citep{entezari2020all, jin2020graph}.
%Thus,  , which restricting the application scope of those methods
%For instance, Table \ref{table:nettack} shows that a state-of-the-art low-rank based method (i.e., ProGNN) even runs out of GPU memory on Pubmed, a graph with only 20k nodes. (2) Since adversarial attacks mainly change the smallest singular values~\citep{entezari2020all}, most singular components (except those smallest ones) of the adversarial graph are almost the same as those of the clean graph, which correspond to the clean graph structure in spatial domain. Nonetheless, due to the high computational cost of SVD, 

\textbf{Reduced-rank topology learning (this work).} Given the adversarial graph $\mathcal{G}_{adv}$ and its adjacency matrix $A_{adv}$, our goal is to learn a reduced-rank graph, which slightly reduces the rank of $\mathcal{G}_{adv}$ to mitigate the effects of adversarial attacks, while retaining most of the important graph spectrum corresponding to the clean graph structure. As adversarial attacks mainly affect the least dominant singular components of $A_{adv}$~\citep{entezari2020all}, one straightforward way for constructing such a reduced-rank graph is to utilize all the singular components except those least dominant ones via TSVD. Nonetheless, computing such a large number of singular components is computationally expensive 
%even with efficient numerical solvers
\citep{baglama2005augmented}, and is thus not scalable to large graphs. 

To learn the reduced-rank graph in a scalable way, in this work, we leverage only the top few (e.g., 50) dominant singular components of $A_{adv}$ to restore its important graph spectrum, via recovering the corresponding clean graph structure with the aid of PGM. 
%\zz{don't define an acronym twice}
%recover the underlying clean graph structure from a given adversarial graph, while ignoring the adversarial components. By recovering the clean graph structure, Figure \ref{figure:rank_cp} shows that \name largely restores the rank of the clean graph, indicating that \name contains more important spectral information corresponding to the clean graph structure than prior low-rank based methods. From the perspective of graph spectrum, \name can be viewed as learning a reduced-rank graph, as the clean graph that \name attempts to recover has slightly smaller rank than the input adversarial graph~\cite{entezari2020all}.
Figure \ref{figure:overview} gives an overview of our proposed approach,  
%\name aims to recover the underlying clean graph structure from a given adversarial graph. Specifically, 
\name, which consists of three major phases. The first phase constructs a base graph by exploiting spectral embedding and a scalable nearest-neighbor graph algorithm. The second phase further refines the base graph by pruning noncritical edges based on PGM. The last phase trains existing GNN models on the refined base graph to improve their robustness.
%(1) base graph construction, (2) graph refinement via edge pruning, (3) downstream GNN training. 
%The first two steps together aim to recover the underlying clean graph structure from a given adversarial graph, which is then used to train a robust GNN model in the last step. 
%aims to construct a base graph ${\mathcal{G}}_{base}$ that satisfies two properties: first, ${\mathcal{G}}_{base}$ should ignore the high-rank adversarial components of ${\mathcal{G}}_{adv}$; second, ${\mathcal{G}}_{base}$ should contain most of important (clean) edges in ${\mathcal{G}}_{adv}$, such that it can serve as a pool of candidate edges for the second step, which is to improve the quality of the base graph by pruning those noncritical edges based on the probabilistic graphical model. As a result, \name produces a sparse graph that is not only immune to adversarial attacks, but also retains the important graph structure for GNN training. 
%Figure \ref{figure:rank_cp} shows that \name maintains most of the clean graph spectrum. 
Next, we will first describe our notion of clean graph recovery via PGM as well as the scalability issue of prior PGM-based work in Section \ref{graph_recover}, which motivates us to develop scalable \name kernels described in Sections \ref{bgc} and \ref{refine}. We further provide the overall complexity of \name in Section \ref{comp}.

%\vspace{-5pt}
\subsection{Graph Recovery via Graphical Model}
%\vspace{-5pt}
\label{graph_recover}
A general philosophy behind PGM is that there exists an underlying graph $G$, whose structure determines the joint probability distribution of the observations on the data entities, i.e., columns of a data matrix $X \in R^{n \times d}$, where $n$ is the number of data points, $d$ the dimension per data point. To recover the underlying graph structure from the data matrix $X$, one common way is to leverage MLE by solving Equation \ref{lasso} in Section \ref{pgm_work}. As the top few dominant singular components of the adjacency matrix capture the corresponding graph structure, 
we can naturally construct the data matrix $X$ based on those dominant singular components, and then adopt PGM to recover an underlying graph via MLE. To this end, we define a weighted spectral embedding matrix as follows:
\begin{definition}
\label{def}
Given the top $r$ smallest eigenvalues ${\lambda}_1, {\lambda}_2, ..., {\lambda}_r$ and their corresponding eigenvectors $v_1, v_2, ..., v_r$ of normalized graph Laplacian matrix $L_{norm} = I - {D}^{-\frac{1}{2}} A {D}^{-\frac{1}{2}}$, where $I$ and $A$ are the identity matrix and graph adjacency matrix, respectively, and $D$ is a diagonal matrix of node degrees, the \textbf{weighted spectral embedding matrix} is defined as $V \overset{\mathrm{def}}{=}\left[ {\sqrt{ \left| 1-{\lambda_1} \right|}}{{v}_1} ,..., {\sqrt{ \left| 1-{\lambda_r} \right|}}{{v}_r} \right]$,
%\begin{equation}\label{spec_embed}
%V \overset{\mathrm{def}}{=}\left[ {\sqrt{ \left| 1-{\lambda_1} \right|}}{{v}_1} ,..., {\sqrt{ \left| 1-{\lambda_r} \right|}}{{v}_r} \right],
%\end{equation}
whose $i$-th row ${V}_{i,:}$ is the \textbf{weighted spectral embedding} of the corresponding $i$-th node in the graph.
\end{definition}

\begin{proposition}%[Informal]
 \label{theory_embed}
 Given a normalized graph adjacency matrix ${A}_{norm} = {D}^{-\frac{1}{2}} A {D}^{-\frac{1}{2}}$ and weighted spectral embedding matrix $V$ of an undirected graph, let $\hat{A}$ be the rank-$r$ approximation of ${A}_{norm}$ via TSVD. If the top $r$ dominant eigenvalues of ${A}_{norm}$ are non-negative, then we have ${\hat{A}} = VV^T$.
 %corresponds to the dot product score between the weighted spectral embeddings of node $i$ and $j$.
 %\vspace{-15pt}
\end{proposition}

Our proof for Proposition \ref{theory_embed} is available in Appendix \ref{proof_1}. Proposition \ref{theory_embed} shows the connection between weighted spectral embedding and the low-rank adjacency matrix $\hat{A}$ obtained by TSVD. Specifically, the weighted spectral embedding matrix $V$ can be viewed as an eigensubspace matrix consisting of a few dominant singular components of the corresponding adjacency matrix. Thus, we can use $V$ to recover the underlying clean graph via PGM. However, obtaining $V$ requires the knowledge of the clean graph structure, which seems to create a chicken and egg problem. 

Fortunately, since the dominant singular components are hardly affected by adversarial attacks~\citep{entezari2020all}, the weighted spectral embedding is therefore also resistant to adversarial attacks, indicating that the underlying clean graph $\mathcal{G}_{clean}$ and its corresponding adversarial graph $\mathcal{G}_{adv}$ share almost the same weighted spectral embeddings. As a result, we can exploit the weighted spectral embedding matrix $V$ of $\mathcal{G}_{adv}$ to represent that of $\mathcal{G}_{clean}$. By replacing the data matrix $X$ with $V$ in Equation \ref{lasso}, we have the following objective function:
%\vspace{-4pt}
\begin{equation}\label{pgm}
\max\limits_{\Theta} F := \log \det \Theta - \frac{1}{r}tr(VV^T\Theta) - \alpha \|\Theta\|_1
%\vspace{-4pt}
\end{equation}

%\vspace{-10pt}
More discussions on Equation~\ref{pgm} are available in Appendix~\ref{eff_resist}. By finding the optimizer $\Theta^*$ 
%in Equation \ref{pgm}
, we can recover the underlying graph that maximizes the likelihood given the observation on the weighted spectral embedding %matrix 
$V$. 
%The edges in $\mathcal{G}_{pgm}$ capture significant associations among nodes, based on the observation of their weighted spectral embeddings under a GMRF. As a result, we can identify those edges in $G_{base}$ to preserve the critical graph structure. 
However, solving Equation \ref{pgm} requires at least $O(n^2)$ time/space complexity per iteration with the most efficient algorithms, which thus cannot scale to large graphs~\citep{friedman2008sparse, hsieh2013sparse, hsieh2014quic}. 
%Instead of directly solving Equation \ref{pgm}, \citet{wang2020learning} first construct a complete graph and then iteratively remove edges between low correlation nodes to produce a sparse graph that 

As $\Theta$ is constrained to be a Laplacian-like matrix, finding the optimizer $\Theta^*$ in Equation \ref{pgm} is equivalent to searching for critical edges from a complete graph,
%\fixme{In contrast to the prior work, we consider solving  as finding the optimizer $\Theta^*$ from a set of valid Laplacian matrices.} \zz{we need to concisely describe what we mean by the ``search space''} From the graph perspective, this is equivalent to identifying critical edges from a complete graph, 
which would involve all possible (i.e., $O(n^2)$) edges. Here we say an edge is critical (noncritical) if including it to the graph significantly increases (decreases) $F$ in Equation \ref{pgm}. 
Hence we can recover the underlying graph by pruning noncritical edges from the complete graph.
However, storing a complete graph is still expensive. % requires quadratic complexity. 
To have a near-linear algorithm for clean graph recovery, instead of searching in the complete graph, we limit our search within an initial base graph $\mathcal{G}_{base}$ that is much sparser but containing sufficient information for identifying the candidate edges critical to recover the clean graph. Subsequently, the final graphical model (graph Laplacian) can be obtained by further pruning noncritical edges from $\mathcal{G}_{base}$.
%Consequently, we can recover the underlying graph by pruning noncritical edges from the $\mathcal{G}_{base}$. 
%that is much sparser than the complete graph while containing most edges in the underlying graph. Consequently, we can recover the underlying graph by first cons and then refine $\mathcal{G}_{base}$ by pruning additional noncritical edges, such that only edges contributing to maximizing $F$ are preserved in the refined graph. 
%We will illustrate details of base graph construction and identifying critical edges in the following two subsections.

%\vspace{-5pt}
\subsection{Base Graph Construction}
%\vspace{-5pt}
\label{bgc}
%\zz{can we refer back to Figure \ref{figure:overview} in this section and maybe other places two}
%To restrict the search space of graph recovering without sacrificing the quality, we aim to 
During the first phase of \name (shown in Figure \ref{figure:overview}), our goal is to build a base graph $\mathcal{G}_{base}$, which greatly reduces the search space by not constructing a complete graph while preserving the critical candidate edges that are key to clean graph recovery. 
%A Na\"ive way is to construct a complete graph $K_n$, which has the same number of nodes as ${\mathcal{G}}_{adv}$. Apparently, $K_n$ is adversarially robust in the sense that no matter how ${\mathcal{G}}_{adv}$ is perturbed by adversarial attacks, $K_n$ always remains the same graph structure. In addition, $K_n$ certainly contains all the critical edges in ${\mathcal{G}}_{adv}$ that are useful for GNN training. 
%Thus, we can produce a high-quality and robust graph by removing unimportant/harmful edges in $K_n$ based on the graph refinement method introduced in Section \ref{refine}. 
%Nonetheless, $K_n$ involves a dense adjacency matrix that results in quadratic space complexity, which cannot scale to large graphs.
%To have a near-linear time/space complexity for building the base graph, 
To this end, we give the following theorem: 
\begin{theorem}%[Informal]
 \label{embed_bound}
 Given a graph $\mathcal{G}=(\mathcal{V}, \mathcal{E})$ and its normalized Laplacian matrix $L_{\mathcal{G}}$, let $V_i$ denote the weighted spectral embedding of node $i$ by using top $r$ eigenpairs of $L_{\mathcal{G}}$. Suppose 
 %$V_i$ is normalized, i.e., $\|V_i\|_2=1$, and 
 a relatively small $r$ is picked such that $\lambda_r \leq 1$, where $\lambda_r$ is the $r$-th smallest eigenvalue of $L_{\mathcal{G}}$, then we have $\sum_{(i,j) \in \mathcal{E}} \|V_i - V_j\|_2^2 \leq 0.25 r$.
 %\vspace{-8pt}
 %\begin{equation}\label{emb_bnd}
%\sum_{(i,j) \in \mathcal{E}} \|V_i - V_j\|_2^2 \leq 0.25 r
%\end{equation}
%\vspace{-20pt}
\end{theorem}

%\vspace{-5pt}
Our proof for Theorem \ref{embed_bound} is available in Appendix \ref{proof_2}. Note that $r$ is a small constant, which is independent of the graph size. 
%\zz{``e.g. r=20'' sounds fishy. why 20? can we just say r is a small constant?} 
Thus, Theorem \ref{embed_bound} indicates that, if an edge connects nodes $i$ and $j$ in the clean graph, then the Euclidean distance between the weighted spectral embeddings of these two nodes will be small, which motivates us to build a k-nearest neighbor (kNN) graph as $\mathcal{G}_{base}$ to incorporate those clean edges. 
%The connection between kNN and TSVD is provided in Appendix~\ref{knn_svd}.

Concretely, we first obtain the weighted spectral embedding matrix $V$ of the input adversarial graph $\mathcal{G}_{adv}$ to represent that of the underlying clean graph $\mathcal{G}_{clean}$, as $V$ consists of dominant singular components that are shared by $\mathcal{G}_{adv}$ and $\mathcal{G}_{clean}$~\citep{entezari2020all}. We then leverage $V$ to construct a kNN graph, where each node is connected to its $k$ most similar nodes based on the Euclidean distance between their spectral embeddings. Note that $V$ can be further concatenated with node feature matrix for constructing the kNN graph. A thorough discussion on incorporating node feature information is available in Appendix~\ref{node_feat}. %Note that we choose a relatively large $k$ (e.g., $k=50\sim100$) to make it more likely for the constructed kNN graph to contain those important edges. Moreover, a n
In this work, we exploit an approximate kNN algorithm for constructing the graph, which has $O(|\mathcal{V}|\log|\mathcal{V}|)$ complexity and thus can scale to very large graphs~\citep{malkov2018efficient, cheng2021spade, chengspade}. 
By choosing a proper $k$ (e.g., $k=50$), $\mathcal{G}_{base}$ is likely to cover edges in the underlying clean graph. Thus, $\mathcal{G}_{base}$ can serve as a reasonable search space for identifying critical edges in the next step.

%\input{table/dataset}
%\vspace{-5pt}
\subsection{Graph Refinement via Edge Pruning}
%\vspace{-5pt}
\label{refine}
For the second phase of \name shown in Figure \ref{figure:overview}, we refine $G_{base}$ by aggressively pruning  noncritical edges from $G_{base}$, such that the refined graph only preserves the most important edges that contribute most to the log-likelihood $F$ in Equation \ref{pgm}. %To this end, \name leverages the notion of probabilistic graphical model (PGM) to determine whether an edge is critical or not. Specifically, let $V \in R^{n \times r}$ be the weighted spectral embedding matrix on the input adversarial graph, where $n$ denotes the number of nodes, $r$ the number of eigenpairs, we can have the following objective function by replacing $X$ with $V$ in Equation \ref{lasso}:

%To identify those critical edges in a scalable way, we do not explicitly compute the optimal $\Theta$. Instead, we exploit Equation \ref{pgm} to guide the processing of identifying critical edges. Specifically, we consider an edge $(i,j)$ to be critical if inserting it (or increasing its edge weight) to the underlying graph will increase the log-likelihood $F$ in Equation \ref{pgm}. 

To identify critical (noncritical) edges that can most effectively increase (decrease) $F$, we exploit the update of $\Theta$ based on gradient ascent: $\Theta \leftarrow \Theta + \eta \frac{\partial F}{\partial \Theta}$, where $\eta$ is the step size. As mentioned in Section \ref{pgm_work}, $\Theta$ is constrained to be $L + \frac{I}{\sigma^{2}}$, which means the off-diagonal elements in $\Theta$ correspond to negative of edge weights in the underlying graph, i.e., $\Theta_{i,j} = -w_{i,j}$. Thus, the update of $\Theta_{i,j}$ during gradient ascent can be viewed as:
%\vspace{-5pt}
\begin{equation}\label{ga_edge}
\Theta_{i,j} \leftarrow \Theta_{i,j} + \eta (\frac{\partial F}{\partial \Theta})_{i,j}=\Theta_{i,j} - \eta \frac{\partial F}{\partial w_{i,j}}
%\vspace{-3pt}
\end{equation}
Equation \ref{ga_edge} means that, if $\frac{\partial F}{\partial w_{i,j}}$ is large and positive, $\Theta_{i,j}$ will become more negative, which corresponds to increasing the edge weight in the underlying graph. Similarly, if $\frac{\partial F}{\partial w_{i,j}}$ is small and negative, $\Theta_{i,j}$ will be less negative, corresponding to decreasing the edge weight. In other words, the edge weight $w_{i,j}$ with a large (small) $\frac{\partial F}{\partial w_{i,j}}$ should be increased (decreased) to maximize the log-likelihood $F$, meaning the corresponding edge is critical (noncritical). Thus, we can identify the critical edges once we know $\frac{\partial F}{\partial w_{i,j}}$.
By setting $\alpha=0$ in Equation \ref{pgm} (as \name naturally produces a sparse graph) and taking the partial derivative with respect to an edge weight $w_{i,j}$, we have:
\begin{equation}\label{edge_grad}
\frac{\partial F}{\partial w_{i,j}} = \sum\limits_{k=1}^n \frac{1}{\lambda_k + 1/\sigma^2}\frac{\partial \lambda_k}{\partial w_{i,j}}-\frac{\|V^Te_{i,j}\|_2^2}{r}
\end{equation}
where $\lambda_k, \forall k=1,2,...,n$ are the Laplacian eigenvalues of $\mathcal{G}_{base}$ (the initial graph for edge pruning), $e_{i,j}=e_i-e_j$, and $e_i$ denotes the vector with all zero entries except for the $i$-th entry being $1$.
\begin{theorem}[\citet{feng2021sgl}]
 \label{feng}
 Let $\lambda_k$ and $u_k$ be the $k$-th eigenvalue and the corresponding eigenvector of the Laplacian matrix, respectively. The spectral perturbation $\delta \lambda_k$ due to the increase of an edge weight $w_{i,j}$ can be estimated by
  %\vspace{-5pt}
 %\begin{equation}\label{delta_spectral}
$\delta \lambda_k = \delta w_{i,j}(u_k^Te_{i,j})^2$.
%\end{equation}
\end{theorem}
%\vspace{-5pt}
The proof for Theorem \ref{feng} is available in~\citet{feng2021sgl}. According to Theorem \ref{feng} and Equation \ref{edge_grad}, we can estimate $\frac{\partial F}{\partial w_{i,j}} \approx \|U^Te_{i,j}\|_2^2 - \frac{1}{r}\|V^Te_{i,j}\|_2^2$, 
%in the following:
%\vspace{-5pt}
%\begin{equation}\label{approx_grad}
%\frac{\partial F}{\partial w_{i,j}} \approx \|U^Te_{i,j}\|_2^2 - \frac{1}{r}\|V^Te_{i,j}\|_2^2
%\end{equation}
where $U=[\frac{u_1}{\sqrt{\lambda_1+1/\sigma^2}}, ..., \frac{u_r}{\sqrt{\lambda_r+1/\sigma^2}}]$, $\lambda_{i}$ is the $i$-th smallest Laplacian eigenvalue of $\mathcal{G}_{base}$, and $u_{i}$ is the corresponding eigenvector. Consequently, an edge $(i,j)$ is critical if $\|U^Te_{i,j}\|_2^2 \gg \frac{1}{r}\|V^Te_{i,j}\|_2^2$. 
As $V$ and $U$ are the spectral embeddings on the input adversarial graph and the base graph, respectively, we define the \textbf{spectral embedding distortion} $s_{i,j} = \frac{\|U^Te_{i,j}\|_2^2}{\|V^Te_{i,j}\|_2^2}$ to measure the edge importance.
%\begin{equation}\label{score}
%$s_{i,j} = \frac{\|U^Te_{i,j}\|_2^2}{\|V^Te_{i,j}\|_2^2}$.
%\end{equation}
%Based on Equation \ref{score}, 
Consequently, we prune edges in the base graph $\mathcal{G}_{base}$ that have small spectral embedding distortion, i.e., $s_{i,j} < \gamma$, where $\gamma$ is a hyperparameter to control the sparsity of the refined graph. We further provide a strategy to simplify the distortion metric for edge pruning in Appendix~\ref{sim_embed_diff}. Hence, the refined base graph $\mathcal{G}_{base}'$ largely recovers the underlying clean graph structure from the input adversarial graph. Since $\mathcal{G}_{base}'$ is constructed by only leveraging the top few dominant singular components of $\mathcal{G}_{adv}$, it ignores the high-rank adversarial components and thus robust to adversarial attacks. As a result, we can train a given GNN model on $\mathcal{G}_{base}'$ to improve its robustness, 
which is the last phase of \name.

\subsection{Complexity of \name}
\label{comp}
%\vspace{-5pt}
The first phase of \name requires $O(r|\mathcal{E}|)$ time for computing top $r$ Laplacian eigenpairs~\citep{baglama2005augmented}, and $O(|\mathcal{V}|\log|\mathcal{V}|)$ time for kNN graph construction~\citep{malkov2018efficient}. The second phase involves $O(rk|\mathcal{V}|)$ time for computing spectral embeddings and edge pruning on the kNN graph. Thus, the overall time complexity for graph purification is $O(r(|\mathcal{E}|+k|\mathcal{V}|)+|\mathcal{V}|\log|\mathcal{V}|)$, where $|\mathcal{V}|$ ($|\mathcal{E}|$) denotes the number of nodes (edges) in the adversarial graph, and $k$ is the averaged node degree in the kNN graph. Our systematic approach of choosing $r$ and the space complexity 
analysis are in Appendix \ref{algo}.
%\vspace{-5pt}

%% file: figure/overview.tex
\begin{figure*}[ht!]
\begin{center}
	\includegraphics[width=0.9\textwidth]{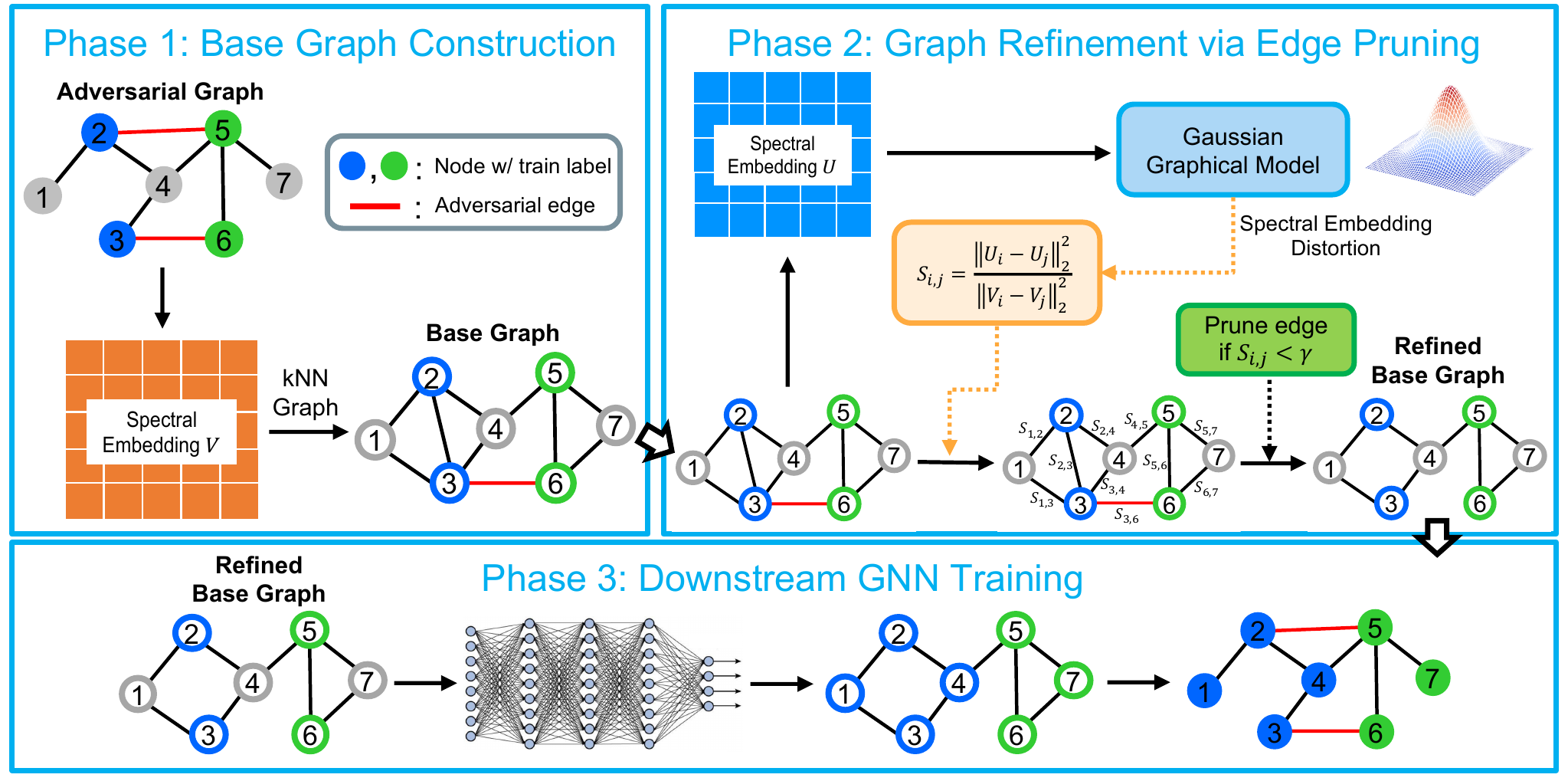}
	\caption{An overview of the three major phases of \name. \protect\label{figure:overview}}
\end{center}
%\vspace{-12pt}
\end{figure*}

%% file: experiments.tex
%\input{table/dataset}
%\input{table/nettack}
%\input{table/hetero}
%\input{table/meta}
\section{Experiments}
We have conducted comparative evaluation of \name against state-of-the-art defense GNN models under targeted attack (Nettack)~\citep{zugner2018adversarial} and non-targeted attack (Metattack)~\citep{zugner2019adversarial} 
%with different perturbation budgets, 
on both homophilic and heterophilic datasets. Besides, we also evaluate \name robustness against adaptive attacks. In addition, we further show the scalability of \name by comparing its run time with prior defense methods and evaluating \name on ogbn-products, which consists of more than $2$ million nodes~\citep{hu2020open}. Finally, we conduct ablation studies to understand the effectiveness of \name kernels.
%is over $100\times$ larger than the datasets used in ~\citet{entezari2020all, zhang2020gnnguard, jin2020graph}. 
%Finally, we show the runtime comparison between \name and prior defense methods.
%perform ablation studies to understand the effectiveness of each kernel in \name. 

%\vspace{-15pt}
\input{table/dataset}
\input{table/nettack1}
\input{table/meta1}
\textbf{Experimental Setup.} 
Table \ref{statistics} shows the statistics of the datasets used in our experiments. We follow~\citet{zhu2020beyond} to compute the homophily score per dataset (lower score means more heterophilic). More details of datasets are available in Appendix \ref{data}. We choose as baselines two state-of-the-art defense methods based on graph purification: TSVD~\citep{entezari2020all} and Pro-GNN~\citep{jin2020graph}. Besides, we evaluate training based defense methods GCN-LFR~\citep{chang2021not} and GNNGuard~\citep{zhang2020gnnguard} on homophilic and heterophilic graphs, respectively. Moreover, we use GCN~\citep{kipf2016semi} and GPRGNN~\citep{chien2021adaptive} as the backbone GNN models for defense on homophilic datasets (i.e., Cora and Pubmed). As GCN performs poorly on heterophilic datasets~\citep{zhu2020beyond, lim2021large}, we choose GPRGNN as the backbone model on Chameleon and Squirrel datasets. Due to the space limit, we provide defense results with H2GCN~\citep{zhu2020beyond} as the backbone model in Appendix~\ref{h2gcn}. 
%More details of the backbone models are available in Appendix \ref{gnn}. 
For all baselines, we tune their hyperparameters against adversarial attacks with a small perturbation, and keep the same hyperparameters for larger adversarial perturbations.
%As ProGNN training is very slow, we replace in its implementation. 
Detailed hyperparameter settings of baselines and \name are available in Appendix~\ref{hyper}.
Our hardware information is provided in Appendix~\ref{hardware}.

%\vspace{-7pt}
\subsection{Robustness of \name}
%\vspace{-5pt}
%\subsubsection{Defense on Homophilic Graphs}
%\label{defend_target}
\textbf{Defense on homophilic graphs}. We first evaluate the model robustness on homophilic graphs against the targeted attack (Nettack) and the non-targeted attack (Metattack). 
Specifically, Nettack aims to fool a GNN model to misclassify some target nodes with a few structure (edge) perturbations. The goal of Metattack is to drop the overall accuracy of the whole test set with a given perturbation ratio budget (i.e., the number of adversarial edges over the number of total edges). 
Due to the space limit, we only show defense results under Nettack and Metattack with $5$ perturbed edges per target node and $20\%$ perturbation ratio, respectively. Results with other perturbation budgets are in Appendix~\ref{more_results}.
%We choose the same set of target nodes as in \citep{jin2020graph} and the attack budget varies from $0$ to $5$ perturbations per target node. 

Table \ref{table:nettack1} reports the average accuracy over $10$ runs on Cora and Pubmed. 
%and two heterophilic datasets (i.e., Chameleon and Squirrel).
It shows that \name, with either a backbone GNN model (GCN or GPRGNN), outperforms defense baselines in terms of both clean and adversarial accuracy in most cases. %indicating that \name recovers the key clean graph structure that contributes to GNN training. 
%Moreover, \name consistently achieves the highest adversarial robustness compared with prior defense methods. 
%and $12.44\%$ over defense methods
%on homophilic and heterophilic datasets, respectively. 
We attribute the large accuracy improvement to \name's strengths in recovering key structures of the clean graph while ignoring the high-rank adversarial components during graph purification. Moreover, as both TSVD and ProGNN involve dense matrices during GNN training, they run out of GPU memory even on Pubmed, a graph with only 20k nodes. In contrast, \name is not only robust to adversarial attacks, but also scalable to large graphs, as empirically shown in Section \ref{large_graphs}.\\

\vspace{-10pt}
\textbf{Defense on heterophilic graphs}. 
%We further evaluate model robustness on heterophilic graphs, i.e., Chameleon and Squirrel.
%against a strong non-targeted attack, i.e., Metattack,
%~\citep{zugner2019adversarial}, 
%whose goal is to drop the overall accuracy of the whole test set with a given perturbation ratio budget (i.e., the number of adversarial edges over the number of total edges). 
We report the averaged accuracy over $10$ runs on heterophilic graphs in Table \ref{table:meta1}, 
%with perturbation ratio in $\{0\%, 10\%, 20\%\}$. As shown in Table \ref{table:meta}, 
which shows that all defense baselines fail to defend GPRGNN on heterophilic graphs and even degrade the accuracy of the vanilla GPRGNN by a large margin. The reason why ProGNN performs poorly is that it follows the graph homophily assumption for improving GNN robustness, which contradicts the property of heterophilic graphs. For the TSVD-based defense method, the low-rank graph generated by TSVD contains negative edge weights, which degrade the performance of GPRGNN for adapting its graph filter on heterophilic graphs~\citep{chien2021adaptive}. Although \citet{zhang2020gnnguard} have shown GNNGuard can improve model robustness on synthetic heterophillic graphs, our results indicate that it fails to defend GNN models on realistic heterohilic graphs. We attribute it to that the quality of graphlet degree vectors used in GNNGuard 
%(heterophily setting) 
is degraded by structural perturbations induced via adversarial attacks. In contrast, \name largely recovers the clean graph structure based on Theorem~\ref{embed_bound} without the assumption on whether adjacent nodes have similar attributes. In other words, \name will produce a heterophilic graph if the underlying clean graph is heterophilic, which is further confirmed in Appendix~\ref{homo_score}. Consequently,
%; it produces a graph without negative edge weights and ignores high-rank adversarial components, 
\name improves accuracy over defense baselines by up to $10.23\%$ (i.e., $43.64\%-33.41\%$ on Squirrel under Nettack) on heterophilic graphs.
%on heterophilic datasets.%enables \name to protect GPRGNN on heterophilic datasets. 
%As a result, \name-GPRGNN achieves adversarial accuracy improvement over the unvaccinated GPRGNN and two defense baselines by a margin up to $6.76\%$ and $13.27\%$, respectively.
%\name gains up to $2.84\%$ and $13.27\%$ accuracy improvement over the defense baselines on homophilic and heterophilic datasets, respectively, which further confirms that \name is resistant to adversarial attacks.
%can indeed preserve the critical graph structure that contributes to GNN training, while ignoring the adversarial components. 
%It is worth noting that TSVD and ProGNN involve dense matrices during GNN training. Both methods run out of GPU memory even on Pubmed, whose graph only contains 20k nodes. In contrast, \name is not only robust to adversarial attacks, but also scalable to large graphs, as empirically shown in Section \ref{large_graphs}.
%averaged accuracy over $10$ runs on Cora, Citeseer, and Pubmed datasets .

%\subsubsection{Defense Against Adaptive Attack}
%\label{adapt_attack}
\textbf{Defense against adaptive attacks}. 
%To show \name is indeed effective for improving model robustness, we further show the results of defense against adaptive attacks. 
As \name is non-differentiable during kNN graph construction, it is difficult to optimize a specific loss function for adaptive attack. Instead, we adopt an attack called LowBlow from~\cite{entezari2020all} (based on Metattack), which deliberately perturbs low-rank singular components in the graph spectrum, yet violates the unnoticeable condition (i.e., preserving node degree distribution after attacking). Since LowBlow has cubic complexity for computing the full set of adjacency eigenpairs, we only show results on the small graph Cora in Table \ref{table:adapt}, which indicates \name still achieves the highest adversarial accuracy under LowBlow, while all low-rank defense baselines perform even worse than vanilla GPRGNN model. The reason lies in that the kNN graph (with a relatively large $k$) in \name is less vulnerable to the perturbations of weighted spectral embeddings (i.e., low-rank singular components)~\citep{wang2018analyzing}, compared to prior low-rank defense methods. 
\input{table/merge}

%Moreover, the graph refinement step can further remove some harmful edges based on PGM.

%As shown in~\cite{entezari2020all}, the low-rank attack typically violates the unnoticeable condition (i.e., similar node degree distribution). The intuition is that the low-rank singular components capture the key global and local graph structure.
%Specifically, given a clean graph adjacency matrix $A$ and the corresponding adversarial adjacency $A'$, 
%Note that this low-rank attack is 

%\input{figure/runtime}
%\input{table/large_graphs}
%\vspace{-7pt}
\subsection{Scalability of \name}
\label{large_graphs}
\input{table/large_graphs}
%\input{figure/runtime}
%\vspace{-5pt}
To demonstrate the scalability of \name, we first compare the run time of \name with prior low-rank defense methods with GPRGNN as the backbone GNN model. As shown in Figure \ref{figure:runtime_comp}, the TSVD defense method is slower than \name since it produces a dense adjacency matrix that slows down the GNN training. Moreover, ProGNN is extremely slow as it jointly learns the low-rank graph structure and the robust GNN model, which requires performing TSVD for every epoch. In contrast, \name can efficiently produce a sparse graph for downstream GNN training, leading to end-to-end runtime speedup over prior methods by up to $14.7\times$.

In addition, we further evaluate the robustness of \name on two large datasets: ogbn-arxiv and ogbn-products, under powerful and scalable attacks proposed by~\cite{geisler2021robustness}. As we run out of GPU memory when performing the PR-BCD attack, we choose the more scalable version GR-BCD that has less memory usage. We use GCN as the backbone model since it outperforms GPRGNN on large graphs. As TSVD and ProGNN run out of memory on these two datasets, we choose GNNGuard, GCNJaccard~\cite{wu2019adversarial}, and Soft Median GDC~\cite{geisler2021robustness} as baselines. Table \ref{grbcd} shows \name achieves comparable clean accuracy compared to GCN, and drastically improves the adversarial accuracy over defense baselines by up to $16.13\%$. Moreover, we also evaluate the run time of \name on the large graphs. Concretely, the end-to-end run time of GARNET is $40$ mins and $4$ hours on ogbn-arxiv and ogbn-products, respectively, which is $3 \times$ faster than the most competitive baseline GNNGuard that takes more than $2$ hours on ogbn-arxiv and $11$ hours on ogbn-products. We provide potential ways to further accelerate \name in Appendix~\ref{runtime_garnet}.

%This means \name can also defend GNN models on large datasets.
%we further evaluate the robustness of \name with GPRGNN as the backbone model on two large datasets: ogbn-arxiv and ogbn-products. Given that existing strongest attacks (Nettack and Metattack) are not scalable to large graphs, we leverage a less powerful yet more scalable attacking algorithm called DICE~\citep{waniek2018hiding}, which randomly connects (disconnects) nodes from different (same) classes, to perturb the graph structure. To have a challenging defense scenario, we consider a $70\%$ perturbation ratio for DICE. 

%while keeping the overall graph size the same. Specifically, we use DICE to first randomly delete $30\%$ edges linking nodes from the same class and then randomly insert $30\%$ edges linking nodes from different classes. 
%Note that we only use training labels for DICE. 
%In regard to baselines, w
%Note that both TSVD and ProGNN run out of memory on these two datasets. Table \ref{dice} shows that \name-GPRGNN slightly degrades the clean accuracy but improves the adversarial accuracy over vanilla GPRGNN by a margin of $6.41\%$ and $6.05\%$ on ogbn-arxiv and ogbn-products, respectively. This means \name can also defend GNN models on large datasets.
%\subsection{Runtime Comparison}
%\label{runtime}

%\vspace{-14pt}
\subsection{Ablation Analysis of \name}
\label{ablation}
\input{figure/abla_refine}
%\vspace{-7pt}
Figure \ref{figs:abla_refine} shows the comparison of GARNET results with and without graph refinement. When only constructing the base graph, GARNET achieves better adversarial accuracy than the vanilla GNN model,
which confirms our Theorem~\ref{embed_bound}
that the base graph construction can successfully recover clean graph edges. The graph refinement step further improves GARNET accuracy ($\sim\!\!2\%$ increase) since some 
noncritical or even harmful edges are removed based on PGM. Due to the space limitation, the ablation studies of \name on the kNN graph and edge pruning are available in Appendix \ref{abla}.

%\vspace{-7pt}
\subsection{Visualization}
\label{visual}
%\vspace{-7pt}
We visualize the local structure (within 2-hop neighbors) of a target node (randomly picked) on Cora in Figure \ref{figure:visual_1255}. By comparing Figures~\ref{figure:adv_1255} and \ref{figure:garnet_1255}, it is clear that \name effectively removes most of the adversarial edges induced by Nettack that connect nodes with different labels~\cite{jin2020graph}. As a result, it is trivial for the backbone GNN model to correctly predict the target node since the surrounding nodes share the same label as the target node in \name graph. This explains why \name substantially improves the adversarial accuracy of GNN models. More visualizations are available in Appendix~\ref{g_visual}. 
%Apart from graph visualization, we further quantitatively measure the quality of clean graph recovery by \name and provide the results in Appendix~\ref{acc_recover}.
\input{figure/visual_fig.tex}
%We provide more visualization results on other target nodes in Appendix.

%% file: table/dataset.tex
\begin{table*}[t!]
\centering
\caption{Statistics of datasets used in our experiments.}
\label{statistics}
\scalebox{0.9}{
%\begin{adjustbox}{width=\columnwidth,center}
\begin{tabular}[t]{llcrrcr}\toprule
\textbf{Dataset}            &   \textbf{Type}  &  \textbf{Homophily Score} &  \textbf{Nodes}   & \textbf{Edges}   & \textbf{Classes}   & \textbf{Features}    \\ \midrule
Cora  &   Homophily   &   $0.80$  &   $2,485$   & $5,069$  & $7$   & $1,433$    \\
%Citeseer &   Homophily  &  $0.74$   &  $2,110$   &  $3,668$   &  $6$   &  $3,703$   \\
Pubmed &  Homophily    &  $0.80$ &  $19,717$   &  $44,324$   &   $3$  & $500$  \\
Chameleon & Heterophily & $0.23$ &  $2,277$   &  $62,792$   &   $5$  & $2,325$  \\
Squirrel &  Heterophily    &  $0.22$ &  $5,201$   &  $396,846$   &   $5$  & $2,089$  \\
ogbn-arxiv & Homophily & $0.66$ &  $169,343$   &  $1,166,243$   &   $40$  & $128$  \\
ogbn-products & Homophily & $0.81$ &  $2,449,029$   &  $61,859,140$   &   $47$  & $100$  \\
\bottomrule
\end{tabular}
}
%\end{adjustbox}
\end{table*}

%% file: table/nettack1.tex
\begin{table*}[ht!]
  \begin{center}
    \caption{Averaged node classification accuracy (\%) $\pm$ std under targeted attack (Nettack) and non-targeted attack (Metattack) on homophilic graphs --- We bold and underline the first and second highest accuracy of each backbone GNN model, respectively. $OOM$ means out of memory.}
    %\vspace{-5pt}
    \label{table:nettack1}
    \begin{adjustbox}{width=\columnwidth,center}
    {\renewcommand{\arraystretch}{1.1}
    \setlength\tabcolsep{1.5 pt}
    \begin{NiceTabular}{l|cc|cc|cc|cc}
      %\toprule % <-- Toprule here
      %Dataset & GCN & GPR & GCNSVD & GPRSVD & ProGCN & ProGPR & \name GCN & \name GPR \\
      %$\alpha$ & $\beta$ & $\gamma$ \\
      %\midrule % <-- Midrule here
      \toprule
 &  \multicolumn{2}{c}{Cora (Nettack)} & \multicolumn{2}{c}{Cora (Metattack)}
 &  \multicolumn{2}{c}{Pubmed (Nettack)} & \multicolumn{2}{c}{Pubmed (Metattack)}\\
\cmidrule(r){2-3} \cmidrule(r){4-5} \cmidrule(r){6-7} \cmidrule(r){8-9}
Model   &  Clean   &  Adversarial  &  Clean   &  Adversarial  &  Clean   &  Adversarial &  Clean   &  Adversarial  \\
\midrule
      GCN-Vanilla & $\underline{80.96} \pm 0.95$ & $55.66 \pm 1.95$ & $\mathbf{81.35} \pm 0.66$ & $56.28 \pm 1.19$ & $87.26 \pm 0.51$ & $66.67 \pm 1.34$ & $\mathbf{87.16} \pm 0.09$ & $77.20 \pm 0.27$ \\
      GCN-TSVD & $72.65 \pm 2.29$ & $60.30 \pm 2.25$ & $73.86 \pm 0.53$ & $62.44 \pm 1.16$ & $87.03 \pm 0.48$ & $\underline{79.56} \pm 0.48$ & $84.53 \pm 0.08$ & $\underline{84.30} \pm 0.08$ \\
      GCN-ProGNN & $80.54 \pm 1.21$ & $\underline{65.38} \pm 1.65$ & $78.56 \pm 0.36$ & $\underline{72.28} \pm 1.67$ & $\mathbf{88.14} \pm 1.44$ & $71.89 \pm 1.56$ & $84.62 \pm 0.11$ & $83.89 \pm 0.32$ \\
      GCN-LFR & $80.07 \pm 0.95$ & $53.73 \pm 2.17$ & $77.23 \pm 2.61$ & $65.38 \pm 3.71$ & $87.20 \pm 1.24$ & $68.49 \pm 2.44$ & $81.91 \pm 0.26$ & $78.32 \pm 0.69$ \\
      GCN-GARNET & $\mathbf{81.08} \pm 2.05$ & $\mathbf{67.04} \pm 2.05$ & $\underline{79.64} \pm 0.75$ & $\mathbf{73.89} \pm 0.91$ & $\underline{87.96} \pm 0.58$ & $\mathbf{86.12} \pm 0.86$ & $\underline{85.37} \pm 0.20$ & $\mathbf{85.14} \pm 0.23$ \\
      \midrule
      GPR-Vanilla & $\mathbf{83.04} \pm 2.05$ & $62.89 \pm 1.95$ & $\mathbf{83.05} \pm 0.42$ & $74.27 \pm 2.11$ & $\underline{90.05} \pm 0.73$ & $\underline{76.99} \pm 1.16$ & $\mathbf{87.35} \pm 0.13$ & $\underline{84.18} \pm 0.15$ \\
      GPR-TSVD & $81.68 \pm 1.78$ & $63.52 \pm 3.27$ & $81.61 \pm 0.54$ & $\underline{78.50} \pm 1.20$ & $OOM$ & $OOM$ & $OOM$ & $OOM$ \\
      GPR-ProGNN & $82.04 \pm 1.33$ & $\underline{63.74} \pm 2.57$ & $82.04 \pm 0.90$ & $76.29 \pm 1.46$ & $OOM$ & $OOM$ & $OOM$ & $OOM$ \\
      %GPR-LFR & $XXX$ & $XXX$ & $XXX$ & $XXX$ & $XXX$ & $XXX$ & $XXX$ & $XXX$ \\
      GPR-GARNET & $\underline{82.77} \pm 1.89$ & $\mathbf{71.45} \pm 2.73$ & $\underline{82.67} \pm 1.89$ & $\mathbf{81.34} \pm 0.79$ & $\mathbf{90.99} \pm 0.52$ & $\mathbf{89.52} \pm 0.45 $ & $\underline{86.86} \pm 0.57$ & $\mathbf{85.69} \pm 0.26$ \\
      \bottomrule % <-- Bottomrule here
    \end{NiceTabular}}
    \end{adjustbox}
  \end{center}
  \vspace{-10pt}
\end{table*}

%% file: table/meta1.tex
\begin{table*}[ht!]
  \begin{center}
    \caption{Averaged node classification accuracy (\%) $\pm$ std on heterophilic graphs --- We bold and underline the first and second highest accuracy, respectively. The backbone GNN model is GPRGNN.}
    %\vspace{-5pt}
    \label{table:meta1}
    \begin{adjustbox}{width=\columnwidth,center}
    {\renewcommand{\arraystretch}{1.1}
    \setlength\tabcolsep{1.5 pt}
    \begin{NiceTabular}{l|cc|cc|cc|cc}
      %\toprule % <-- Toprule here
      %Dataset & GCN & GPR & GCNSVD & GPRSVD & ProGCN & ProGPR & \name GCN & \name GPR \\
      %$\alpha$ & $\beta$ & $\gamma$ \\
      %\midrule % <-- Midrule here
      \toprule
 &  \multicolumn{2}{c}{Chameleon (Nettack)} & \multicolumn{2}{c}{Chameleon (Metattack)}
 &  \multicolumn{2}{c}{Squirrel (Nettack)} & \multicolumn{2}{c}{Squirrel (Metattack)}\\
\cmidrule(r){2-3} \cmidrule(r){4-5} \cmidrule(r){6-7} \cmidrule(r){8-9}
Model   &  Clean   &  Adversarial  &  Clean   &  Adversarial  &  Clean   &  Adversarial &  Clean   &  Adversarial  \\
\midrule
      Vanilla & $\underline{71.46} \pm 1.92$ & $\underline{66.26} \pm 1.71$ & $\mathbf{61.36} \pm 1.00$ & $\underline{53.20} \pm 0.88$ & $\underline{41.36} \pm 2.87$ & $\underline{39.45} \pm 2.36$ & $\underline{39.51} \pm 1.64$ & $\underline{35.22} \pm 1.20$ \\
      TSVD & $62.12 \pm 3.04$ & $60.37 \pm 2.86$ & $47.29 \pm 1.63$ & $45.12 \pm 1.34$ & $32.98 \pm 2.36$ & $31.20 \pm 1.84$ & $31.36 \pm 1.87$ & $23.91 \pm 1.40$ \\
      ProGNN & $58.80 \pm 1.72$ & $57.07 \pm 1.82$ & $48.39 \pm 0.68$ & $46.69 \pm 0.61$ & $31.81 \pm 1.72$ & $27.27 \pm 1.87$ & $31.64 \pm 2.87$ & $29.36 \pm 3.61$ \\
      GNNGuard & $64.87 \pm 2.62$ & $62.21 \pm 1.94$ & $58.01 \pm 1.57 $ & $49.89 \pm 1.34$ & $34.17 \pm 2.33$ & $33.41 \pm 1.82$ & $37.46 \pm 0.56$ & $32.69 \pm 0.59 $ \\
      GARNET & $\mathbf{72.89} \pm 2.65$ & $\mathbf{71.83} \pm 2.11$ & $\underline{61.11} \pm 2.46$ & $\mathbf{59.96} \pm 0.84$ & $\mathbf{44.91} \pm 1.53$ & $\mathbf{43.64} \pm 1.53$ & $\mathbf{43.43} \pm 1.14$ & $\mathbf{41.97} \pm 1.02$ \\
      \bottomrule % <-- Bottomrule here
    \end{NiceTabular}}
    \end{adjustbox}
  \end{center}
  \vspace{-15pt}
\end{table*}

%% file: table/merge.tex
%\vspace{-50pt}
\begin{table}
%\vspace{-15pt}
	\begin{minipage}{0.49\linewidth}
	\vspace{-15pt}
		\caption{Averaged accuracy (\%) $\pm$ std on Cora under Metattack and LowBlow with $20\%$ perturbation ratio. We use GPRGNN as the backbone GNN model.}
    \vspace{4pt}
    \label{table:adapt}
		\centering
		\begin{adjustbox}{width=0.9\columnwidth,center}
		\begin{tabular}{c|cc}
      \toprule
 %& \multicolumn{2}{c}{Metattack}\\
 %& Metattack
 %& LowBlow\\
%\cmidrule(r){2-3}
Model   &  Metattack   &  LowBlow  \\
\midrule
      Vanilla & $74.27 \pm 2.11$ & $74.77 \pm 0.71$ \\
      TSVD & $78.50 \pm 1.20$ & $26.03 \pm 2.76$ \\
      ProGNN & $76.29 \pm 1.46$ & $69.88 \pm 1.61$ \\
      %LFR & $XXX$ & $XXX$ & $XXX$ & $XXX$ \\
      %GNNGuard & $XXX$ & $XXX$ & $XXX$ & $XXX$ \\
      \name & $\mathbf{81.34} \pm 0.79$ & $\mathbf{77.71} \pm 0.95$ \\
      \bottomrule % <-- Bottomrule here
    \end{tabular}
    \end{adjustbox}
	\end{minipage}\hfill
	%\begin{minipage}{0.46\linewidth}
         \begin{minipage}{0.46\linewidth}
		\centering
		\centerline{\includegraphics[width=0.8\columnwidth]{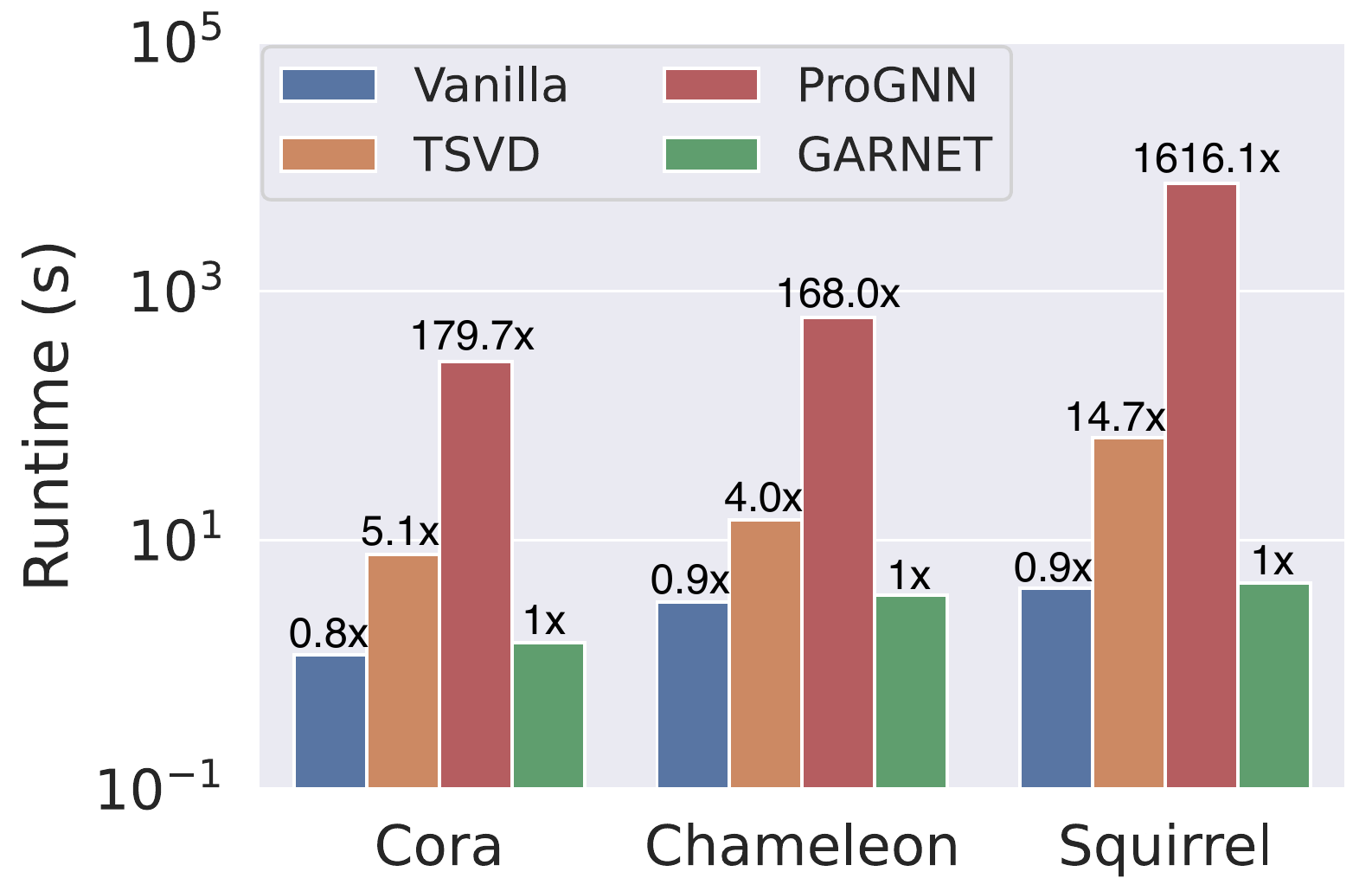}}
        %{figure/runtime_comp.pdf}}
		\captionof{figure}{End-to-end runtime comparison of \name and baseline methods.}
\label{figure:runtime_comp}
	\end{minipage}
	\vspace{-15pt}
\end{table}

%% file: table/large_graphs.tex
\iffalse
\begin{table}[ht]
\centering
\caption{Averaged clean and adversarial accuracy (\%) $\pm$ std 
%$\pm$ std as well as adversarial accuracy (\%) $\pm$ std 
under non-targeted attack (DICE) with $70\%$ perturbation ratio. 
%--- We bold the highest accuracy.
%$OOM$ indicates out of memory.
}
\label{dice}

\scalebox{0.99}{
\begin{adjustbox}{width=\columnwidth,center}
\begin{tabular}[t]{lclcl}
\toprule
 &  \multicolumn{2}{c}{\textbf{ogbn-arxiv-DICE}} & \multicolumn{2}{c}{\textbf{ogbn-products-DICE}}\\
\cmidrule(r){2-3} \cmidrule(r){4-5} 
\textbf{Method}   & Clean & Adversarial   & Clean & Adversarial \\
\midrule
%GCN          & $\mathbf{71.09} \pm 0.19$ & $60.24 \pm 0.23$ & $43.24$ & $\underline{75.64} \pm 0.21$ & $65.74 \pm 0.50$ & $14.34$   \\
%GCN-Jaccard          & $67.72 \pm 0.19$ & $58.66 \pm 0.11$ & $90.32$ & $72.93 \pm 0.09$ & $64.51 \pm 0.10$ & $37.95$  \\
%GCN-SVD          & $OOM$ & $OOM$ & $NA$ & $OOM$ & $OOM$ & $NA$  \\
GPRGNN         & $\textbf{68.04} \pm 0.23$ & $56.28 \pm 0.37$ & $\mathbf{72.64} \pm 0.09$ & $64.33 \pm 0.12$  \\
%Pro-GNN          & $OOM$ & $OOM$ & $OOM$ & $OOM$ & $OOM$ & $OOM$  \\

%\midrule
\name          & $67.15 \pm 0.42$ & $\mathbf{62.69} \pm 0.22$ & $71.35 \pm 0.11$ & $\mathbf{70.38} \pm 0.17$  \\
\bottomrule

\end{tabular}
\end{adjustbox}
}
\end{table}
%\fixme{need to solve the runtime issue on products}
\fi

%\renewcommand{\thetable}{\Alph{table}}
\begin{table}[ht!]
%\vspace{-5pt}
\centering
\caption{Averaged accuracy (\%) $\pm$ std 
%$\pm$ std as well as adversarial accuracy (\%) $\pm$ std 
under GR-BCD attack. 
%--- We bold the highest accuracy.
%$OOM$ indicates out of memory.
}
%\vspace{8pt}
\label{grbcd}
%\vspace{-5pt}
%\scalebox{0.99}{
\begin{adjustbox}{width=\columnwidth,center}
\begin{tabular}[t]{lcccccc}
\toprule
&  \multicolumn{3}{c}{ogbn-arxiv} 
&  \multicolumn{3}{c}{ogbn-products}
\\
\cmidrule(r){2-4}
\cmidrule(r){5-7}
Model   & Clean & $25\%$ Ptb.   & $50\%$ Ptb. 
& Clean & $25\%$ Ptb.   & $50\%$ Ptb. 
\\
\midrule
GCN         & $\textbf{70.74} \pm 0.26$ & $45.18 \pm 0.25$ & $39.12 \pm 0.27$ 
& $\underline{75.68} \pm 0.20$ & $64.70 \pm 0.43$ & $62.71 \pm 0.44$  
\\
GNNGuard         & $68.78 \pm 0.32$ & $\underline{47.46} \pm 0.11$ & $\underline{41.18} \pm 0.12$ 
& $74.82 \pm 0.11$ & $\underline{66.76} \pm 0.23$ & $\underline{63.22} \pm 0.26$ 
\\
GCNJaccard        & $67.77 \pm 0.18$ & $46.27 \pm 0.11$ & $40.84 \pm 0.19$ 
& $72.95 \pm 0.08$ & $60.90 \pm 0.18$ & $58.84 \pm 0.20$
\\
Soft Median GDC        & $69.75 \pm 0.03$ & $45.31 \pm 0.06$ & $40.11 \pm 0.06$ 
& $66.31 \pm 0.03$ & $60.59 \pm 0.05$ & $59.73 \pm 0.05$ 
\\
GARNET          & $\underline{69.91} \pm 0.29$ & $\textbf{61.32} \pm 0.20$ & $\textbf{60.88} \pm 0.13$ 
& $\textbf{76.05} \pm 0.19$ & $\textbf{75.03} \pm 0.14$ & $\textbf{74.97} \pm 0.24$  
\\
%\midrule
%\midrule
%&  \multicolumn{3}{c}{ogbn-products} \\
%\cmidrule(r){2-4} 
%ogbn-products   & Clean & $25\%$ Ptb.   & $50\%$ Ptb. \\
%\midrule
%GCN         & $75.68 \pm 0.20$ & $64.70 \pm 0.43$ & $62.71 \pm 0.44$  \\
%GNNGuard         & $74.82 \pm 0.11$ & $67.43 \pm 0.23$ & $64.67 \pm 0.26$  \\
%GARNET          & $\textbf{76.05} \pm 0.19$ & $\textbf{75.03} \pm 0.14$ & $\textbf{74.97} \pm 0.24$  \\
\bottomrule

\end{tabular}
\end{adjustbox}
%}
%\vspace{-20pt}
\end{table}

%% file: figure/abla_refine.tex
\begin{figure}[ht]
\begin{center}
%\vspace{-8pt}
\centerline{\includegraphics[width=0.9\columnwidth]{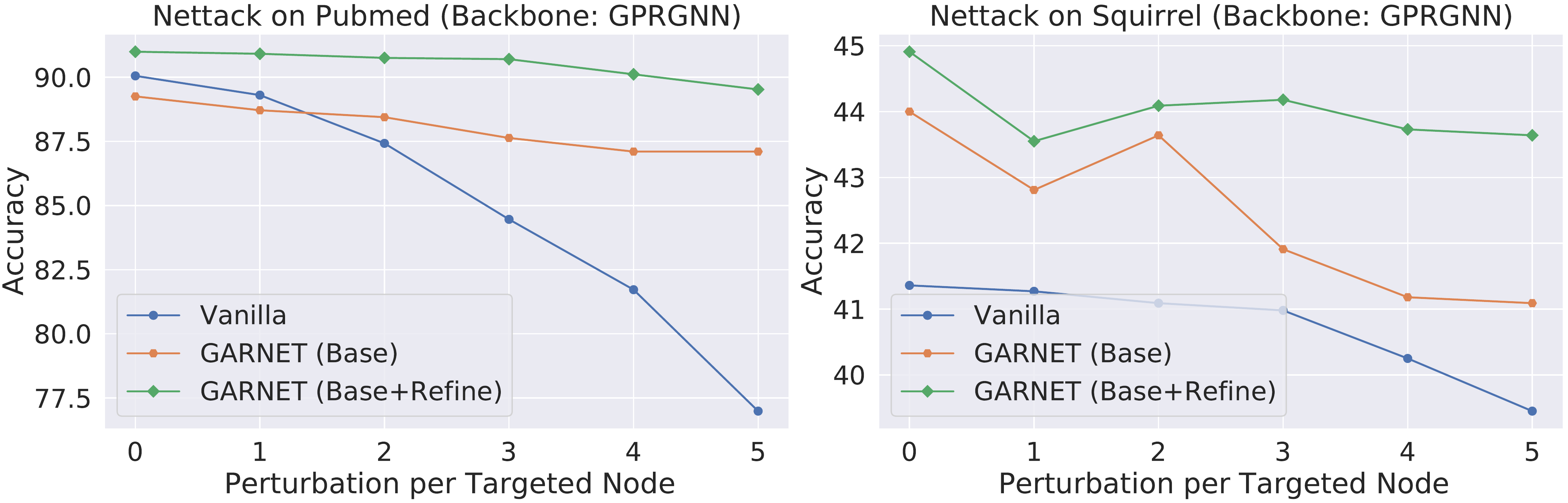}}
%\vspace{-13pt}
\caption{Ablation study of GARNET on graph refinement.}
\label{figs:abla_refine}
\end{center}
\vspace{-20pt}
\end{figure}

%% file: figure/visual_fig.tex
\begin{figure*}[ht]% [hpbt] what you need
    \centering
    \vspace{-10pt}
     \subfigure[]{%
        \includegraphics[width=0.32\textwidth]{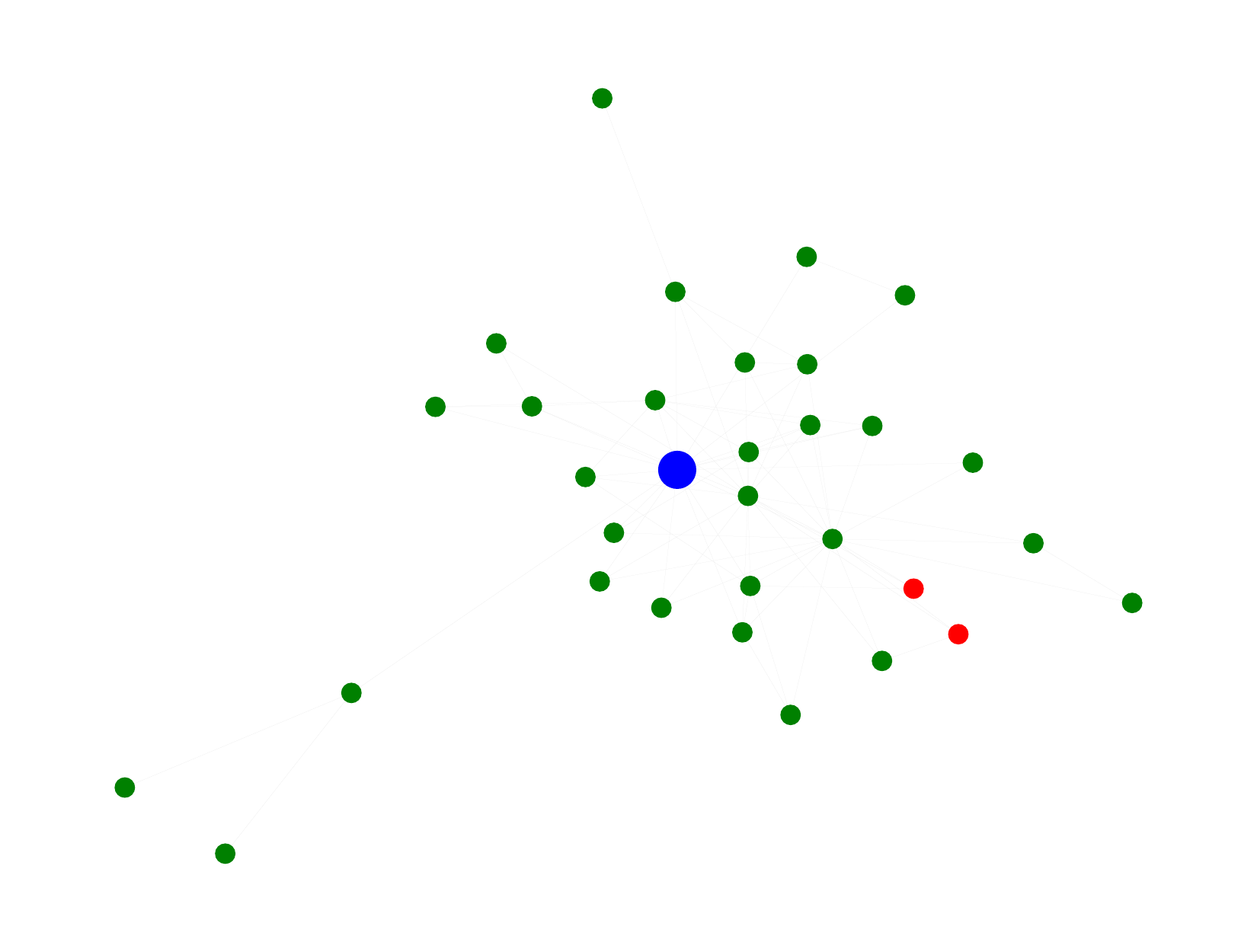}
        \label{figure:clean_1255}}
    %\hspace{0.1\textwidth}
    \subfigure[]{%{0.3\textwidth}
        \includegraphics[width=0.32\textwidth]{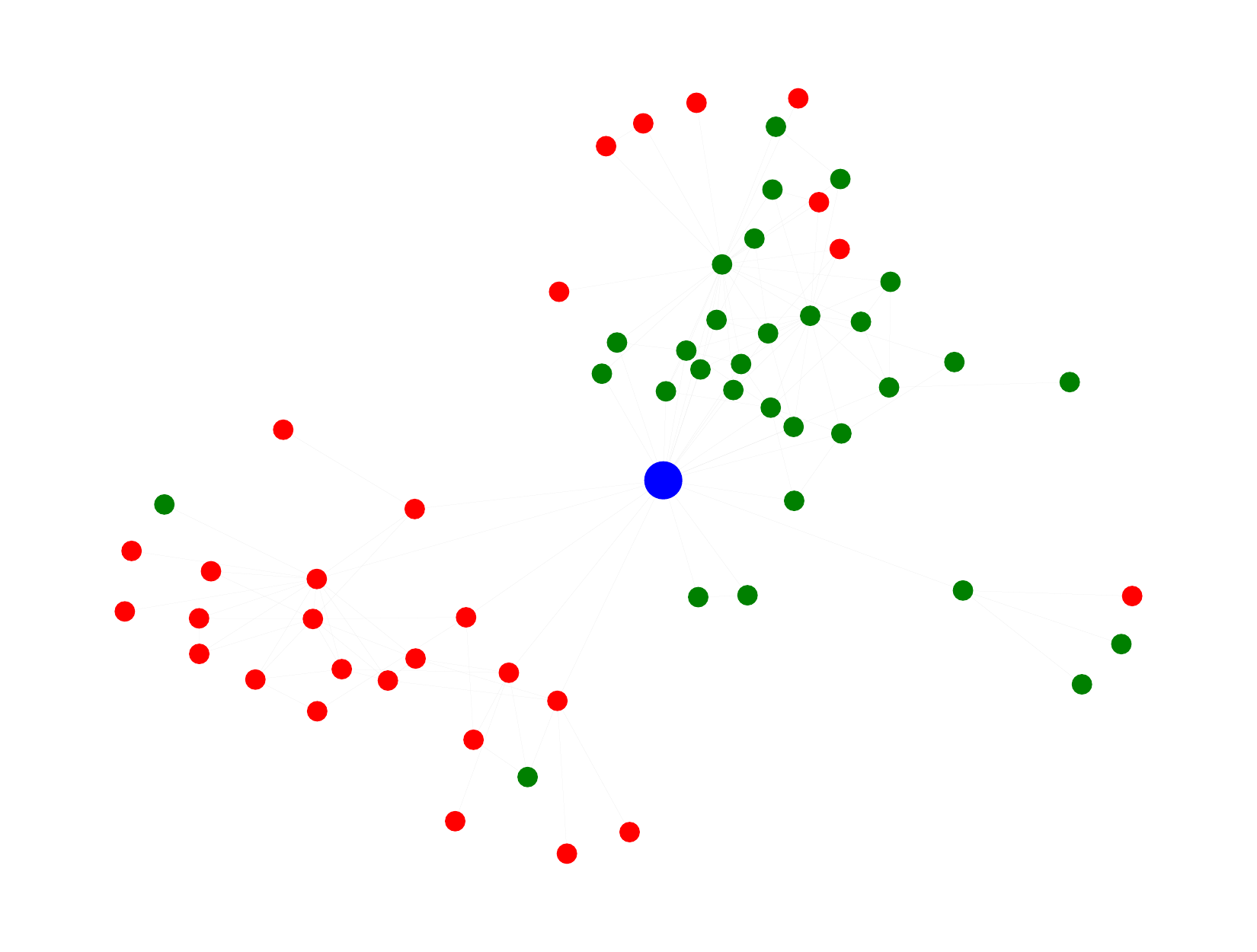}
        \label{figure:adv_1255}}
     \subfigure[]{%{0.3\textwidth}
        \includegraphics[width=0.32\textwidth]{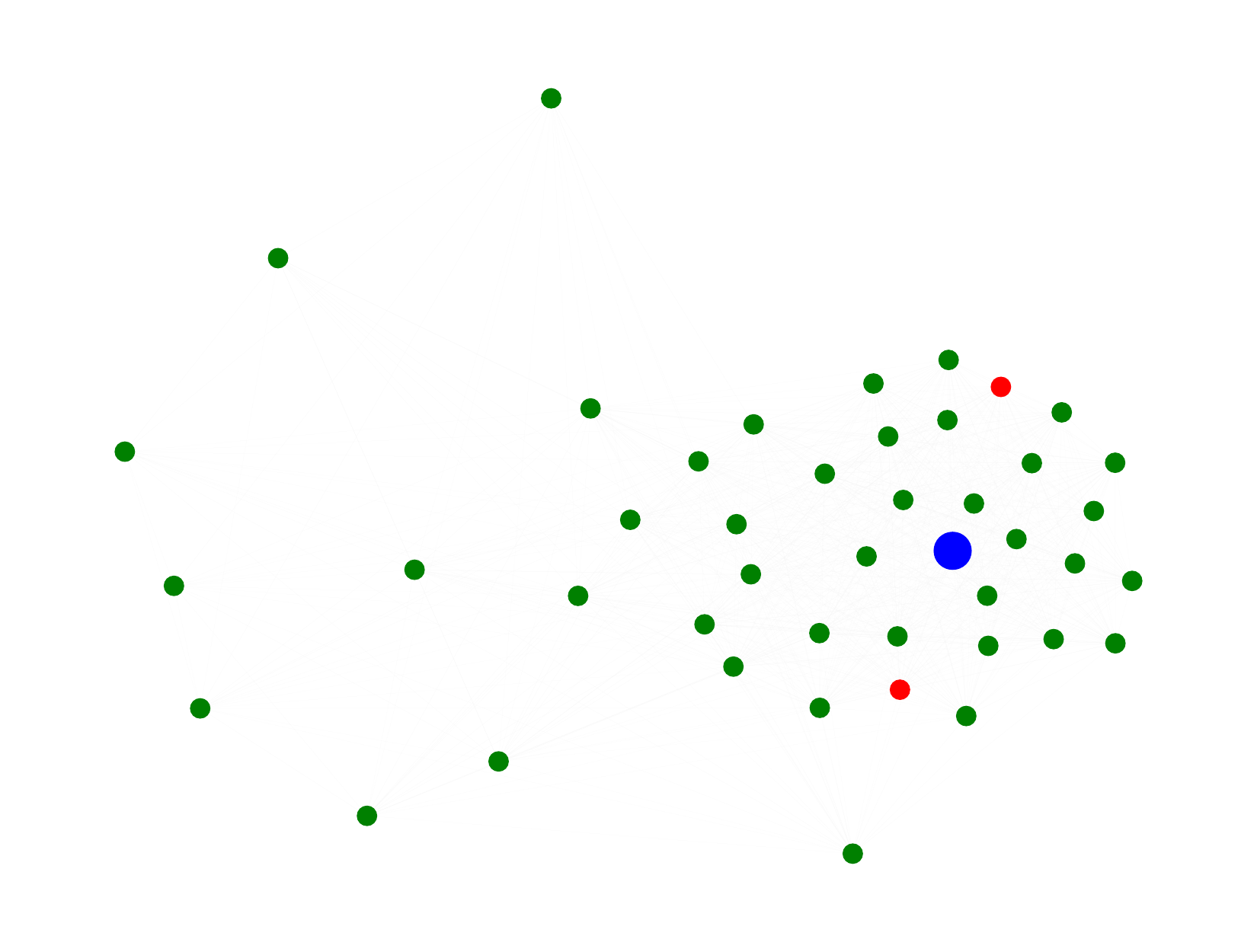}
        \label{figure:garnet_1255}}
    %\vspace{-6pt}
    \caption{Visualizations on the same target node (marked in blue) as well as its 1-hop and 2-hop neighbors. Neighbor nodes are marked in green if they have the same label as the target node, and red otherwise. (a) clean graph. (b) adversarial graph. (c) adversarial graph purified by \name.}
    \vspace{-15pt}
    \label{figure:visual_1255}
    %\vspace{-1.6cm}
\end{figure*}

%% file: conclusions.tex
%\vspace{-8pt}
\section{Conclusions}
\label{conclude}
%\vspace{-5pt}
This work introduces \name, a spectral approach to robust and scalable graph neural networks by combining spectral embedding and the probabilistic graphical model. \name first uses weighted spectral embedding to construct a base graph, which is then refined by pruning uncritical edges based on the graphical model. Results show that \name not only outperforms state-of-the-art defense models, 
%on homophilic and heterophilic graphs under both targeted and non-targeted adversarial attacks, 
but also scales to large graphs with millions of nodes.  An interesting direction for future work is to incorporate high-order structural information (e.g., motifs) to further boost model robustness.

%% file: acknowledgements.tex
\section*{Acknowledgements}
This work is supported in part by NSF grants \#2021309, \#2205572, \#2212370, and \#2212371, and a Facebook Research Award.

%% file: appendix.tex
\section{Proof for Proposition \ref{theory_embed}}
\label{proof_1}
\begin{proof}
As the graph is undirected, we can perform eigendecomposition on both $A_{norm}$ and $L_{norm}$ to obtain their real eigenvalues and the corresponding eigenvectors. Let ${\lambda}_i$, $\hat{\lambda}_i$, and ${\sigma}_i, i=1,2,...,r$ denote the $r$ smallest eigenvalues of $L_{norm}$, $r$ largest eigenvalues of $A_{norm}$, and $r$ largest singular values of $A_{norm}$, respectively. Since $A_{norm} = I - L_{norm}$, $A_{norm}$ and $L_{norm}$ share the same set of eigenvectors while their eigenvalues satisfy: $\hat{\lambda}_i = 1 - {\lambda}_i, i=1,2,...,r$. Moreover, since we assume that the $r$ largest magnitude eigenvalues of $A_{norm}$ are non-negative, we have ${\sigma}_i = \left| \hat{\lambda}_i \right| = \hat{\lambda}_i, i=1,2,...,r$. Thus, we have:
\begin{align*}
  VV^T &= \left[ {{v}_1} ,..., {{v}_r} \right] 
  \begin{bmatrix}
    { \left| 1-{\lambda_1} \right|} & & \\
    & \ddots & \\
    & & {\left| 1-{\lambda_r} \right|}
  \end{bmatrix}
  \left[ {{v}_1} ,..., {{v}_r} \right]^T \\
  &= \left[ {{v}_1} ,..., {{v}_r} \right] 
  \begin{bmatrix}
    { \left| \hat{\lambda_1} \right|} & & \\
    & \ddots & \\
    & & {\left| \hat{\lambda_r} \right|}
  \end{bmatrix}
  \left[ {{v}_1} ,..., {{v}_r} \right]^T \\
  &= \left[ {{v}_1} ,..., {{v}_r} \right] 
  \begin{bmatrix}
    { \sigma_1 } & & \\
    & \ddots & \\
    & & { \sigma_r }
  \end{bmatrix}
  \left[ {{v}_1} ,..., {{v}_r} \right]^T \\
  &= \hat{A}
\end{align*}
%Since $V_{i,:}$ is defined as the weighted spectral embedding of node $i$ and $\hat{A}_{i,j} = V_{i,:}V_{j,:}^T$, $\hat{A}_{i,j}$ corresponds to the dot product score between the weighted spectral embeddings of node $i$ and $j$, which completes the proof of the theorem.
\end{proof}

\section{Proof for Theorem \ref{embed_bound}}
\label{proof_2}
\begin{proof}
Since the weighted embedding matrix $V$ is defined as $V \overset{\mathrm{def}}{=}\left[ {\sqrt{ \left| 1-{\lambda_1} \right|}}{{v}_1} ,..., {\sqrt{ \left| 1-{\lambda_r} \right|}}{{v}_r} \right]$, where ${\lambda}_1, {\lambda}_2, ..., {\lambda}_r$ and $v_1, v_2, ..., v_r$ are the top $r$ smallest eigenvalues and the corresponding eigenvectors of normalized graph Laplacian matrix $L_{norm} = I - {D}^{-\frac{1}{2}} A {D}^{-\frac{1}{2}}$, we have:
\begin{align*}
  \sum_{(i,j) \in \mathcal{E}} \|V_i - V_j\|_2^2  &= \sum_{k=1}^r \sum_{(i,j) \in \mathcal{E}} \left| 1-{\lambda_k} \right| (v_{k,i}-v_{k,j})^2 \\
  &= \sum_{k=1}^r \left| 1-{\lambda_k} \right| \sum_{(i,j) \in \mathcal{E}} (v_{k,i}-v_{k,j})^2 \\
  &= \sum_{k=1}^r \left| 1-{\lambda_k} \right| v_k^TL_{norm}v_k \\
  &= \sum_{k=1}^r ( 1-{\lambda_k} ) \lambda_k \\
  &\leq \sum_{k=1}^r 0.25 \\
  &= 0.25 r
\end{align*}
\end{proof}

The fourth equation above is based on Courant-Fischer Theorem~\citep{golub2013matrix} with the assumption that $\lambda_r \leq 1$ and the Laplacian eigenvectors are normalized (i.e., $\|v_k\|_2=1, \forall k=1,...,r$). The inequality is derived by arithmetic mean-geometric mean (AM-GM) inequality.

\section{Dataset Details}
\label{data}
%\input{table/dataset}
%Table \ref{statistics} shows the statistics of the datasets used in our experiments. We follow~\citet{zhu2020beyond} to compute the homophily score per dataset (lower score means more heterophilic). 
As in \citet{jin2020graph}, we extract the largest connected components of the original Cora and Pubmed datasets~\citep{yang2016revisiting} for the adversarial evaluation, with the same train/validation/test split. For Chameleon and Squirrel~\citep{rozemberczki2021multi}, we keep the same split setting as \citet{chien2021adaptive}. Finally, we follow the split setting of Open Graph Benchmark (OGB) \citep{hu2020open} on ogbn-arxiv and ogbn-products. Note that all data used in our experiments do not contain personally identifiable information or offensive content.

In addition, we follow \citet{jin2020graph} for the selection of target nodes on Cora and Pubmed under Nettack. For the Chameleon and Squirrel datasets under Nettack, we choose target nodes that have degrees within the range of $\left[20, 50 \right]$ and $\left[20, 140 \right]$, respectively. In regard to non-targeted attacks (i.e., Metattack), we choose nodes in the test set as target nodes for all datasets. We implement all the adversarial attacks based on the DeepRobust library~\citep{li2020deeprobust}.
 
\section{Hyperparameters Settings} 
\label{hyper}

\subsection{Backbone GNN Models}

\textbf{GCN.} We choose the GCN hyperparameters based on the DeepRobust library~\citep{li2020deeprobust}.

\textbf{GPRGNN.} We follow the hyperparameter settings provided at \href{https://github.com/jianhao2016/GPRGNN}{github.com/jianhao2016/GPRGNN} with slightly different dropout rates (chosen from {$0.3, 0.5, 0.7$}) and learning rates (chosen from {$0.01, 0.05, 0.1$}). Specifically, we provide the complete choices of dropout rates and learning rates across all datasets and attack settings below:

\noindent $\bullet$ Cora-Nettack: dropout of $0.5$ and learning rate of $0.01$.

\noindent $\bullet$ Cora-Metattack: dropout of $0.5$ and learning rate of $0.01$.

\noindent $\bullet$ Pubmed-Nettack: dropout of $0.5$ and learning rate of $0.01$.

\noindent $\bullet$ Pubmed-Metattack: dropout of $0.5$ and learning rate of $0.01$.

\noindent $\bullet$ Chameleon-Nettack: dropout of $0.5$ and learning rate of $0.05$.

\noindent $\bullet$ Chameleon-Metattack: dropout of $0.3$ and learning rate of $0.05$.

\noindent $\bullet$ Squirrel-Nettack: dropout of $0.5$ and learning rate of $0.1$.

\noindent $\bullet$ Squirrel-Metattack: dropout of $0.5$ and learning rate of $0.1$.

\textbf{H2GCN.} We train a three-layer model in full batch, with a learning rate of $0.01$, dropout of $0.5$, hidden dimension of $64$, and $300$ epochs for both Chameleon and Squirrel datasets.

\subsection{Defense Baselines}

\textbf{TSVD.} We use the same r eigenvectors in TSVD as those used in GARNET, which is shown in Table~\ref{table:hyperpara}.

\textbf{GCNJaccard.} We choose the GCNJaccard hyperparameters based on the DeepRobust library~\citep{li2020deeprobust}.

\textbf{GNNGuard.} We set edge pruning threshold (the only hyperparameter in GNNGuard) to be $P_0=0.1$.

\textbf{Soft Median GDC.} We strictly follow the hyperparameter setting suggested by \citet{geisler2021robustness}. In particular, we choose temperature $T$ of $5.0$ for soft median, $\alpha$ of 0.1 (0.15) and $k$ of 64 (32) for GDC on ogbn-arxiv (ogbn-products).

\textbf{ProGNN.} We find out its performance is very sensitive to hyperparameters. Thus we strictly follow the tuned hyperparameters available at \href{https://github.com/ChandlerBang/Pro-GNN/tree/master/scripts}{github.com/ChandlerBang/Pro-GNN/scripts}. As GCN-ProGNN training is very slow on Pubmed (estimated time is $30$ days for $10$ runs), we follow the suggestion from ProGNN authors to replace ``svd'' with ``truncated svd'' in the ProGNN implementation.

%\vspace{-5pt}
\subsection{\name}
%\vspace{-5pt}
\input{table/hyperpara}
%\vspace{-10pt}
We run all GNN training with a full batch way and show the hyperparameters of \name on different datasets under Nettack ($1$ perturbation per node), Metattack ($10\%$ perturbation ratio), and GR-BCD ($25\%$ perturbation ratio) in Table \ref{table:hyperpara}. Note that we provide our strategy of choosing $r$ in Appendix~\ref{algo}, which avoids conducting hyperparameter tuning on $r$ per dataset. Besides, we set the prior data variance $\sigma^2$ to be positive infinity for all graphs (i.e., ignore self-loops of $\Theta$ in Equation~\ref{pgm}). Last but not least, we refer readers to our official configuration files at \href{https://github.com/cornell-zhang/GARNET/tree/main/configs}{github.com/cornell-zhang/GARNET/configs} for the detailed choices of the threshold $\gamma$ for edge pruning. 
%For the adaptive label propagation kernel, we do not manually tune hyperparameters but instead leverage Optuna~\citep{akiba2019optuna} to optimize the hyperparameters such as the scale $\beta$ for error correction based on the validation set, since this kernel is fast to compute. 
%In addition, we run all GNN training with a full batch way.

%\vspace{-10pt}
\section{Hardware Information}
\label{hardware}
%For ogbn-arxiv and ogbn-products, we run experiments on a Linux machine with an Intel Xeon Silver $4214$ CPU ($8$ cores @ $2.20$GHz) and 8 NVIDIA RTX A6000 GPUs ($48$ GB memory per GPU). For the rest of the datasets, 
We conduct all experiments on a Linux machine with an Intel Xeon Gold $5218$ CPU ($8$ cores @ $2.30$GHz), 8 NVIDIA RTX $2080$ Ti GPU ($11$ GB memory per GPU), and 1 RTX A$6000$ GPU ($48$ GB memory).

\section{Complexity Analysis of \name}
\label{algo}

\subsection{Time Complexity -- Choice of \texorpdfstring{$r$} -}
We choose $r$ based on the number of classes per dataset, which depends on the downstream task rather than number of nodes in the graph. Specifically, suppose $\lambda_r$ is the $r$-th largest eigenvalue, an appropriate $r$ is chosen if there is a large gap between $\lambda_r$ and $\lambda_{r+1}$ (i.e., a large eigengap) in the graph spectrum. According to~\cite{von2007tutorial}, the eigengap is highly related to the number of clusters in the graph. In this work, we approximate $r$ by $r \approx 10c$ to cover the large eigengap, where $c$ denotes the number of classes/clusters. As shown in Tables \ref{statistics} and \ref{table:hyperpara}, the number of classes in small (large) graphs is around $5$ ($50$), so we use $r=50$ ($r=500$) in experiments. As a result, GARNET has the near-linear time complexity $O(r(|E|+k|V|)+|V|log|V|) = O(c(|E|+k|V|)+|V|log|V|)$.

\subsection{Space Complexity}
\name involves forming a sparse kNN graph by building hierarchical navigable small world (HNSW) graphs~\citep{malkov2018efficient} that contain $O(\left|V\right|\log\left|V\right|)$ nodes in total and each node connects to a fixed number of neighbors. Thus, the space complexity of storing the HNSW graphs is  $O(\left|V\right|\log\left|V\right|)$. In addition, \name also needs to store the input adversarial graph and the produced kNN graph. As a result, the total space complexity of \name is $O(\left|V\right|(\log\left|V\right|+k) + |\mathcal{E}|)$, where $|\mathcal{V}|$ and $|\mathcal{E}|$ denote the number of nodes and edges in the adversarial graph, respectively, and $k$ is the averaged node degree in the kNN graph.

Apart from the complexity analysis, we further provide the algorithms of \name and TSVD below for comparison. For \name algorithm, the embedding matrix $V$ at line $3$ can be further concatenated with the node feature matrix, which may improve the quality of $\mathcal{G}_{base}$ as discussed in Appendix~\ref{node_feat}. Moreover, lines $5$ and $6$ are optional as illustrated in Appendix~\ref{sim_embed_diff}.

\input{table/algorithm}

\section{Ablation Study}
\label{abla}
\subsection{Choice of \texorpdfstring{$k$} f for kNN Graph Construction}
\label{k_val}
\input{figure/knn}
To evaluate the sensitivity of \name to k nearest-neighbor (kNN) graph construction, we evaluate the adversarial accuracy of \name with different $k$ values for constructing kNN graphs. Figure \ref{figure:knn} shows that the accuracy of \name does not change too much when varying $k$ value within the range of $\left[10, 100 \right]$, indicating a relatively large $k$ (e.g., $k\ge10$) can enable the kNN graph to incorporate most of edges in the underlying clean graph. Consequently, the performance of \name is relatively robust to the choice of $k$ for kNN graph construction. As the peak performance is typically achieved in $\left[30, 80 \right]$, we recommend choosing $k=30\sim80$ for building the kNN graph in practice.

\subsection{Choice of \texorpdfstring{$\gamma$} f for edge pruning}
\label{edge_prune}
\input{figure/edge_prune}
Apart from the choice of $k$ for kNN graph construction, another critical hyperparameter of \name is the threshold $\gamma$ that determines whether an edge should be pruned in the base graph. Thus, we further evaluate the effect of $\gamma$ on the performance of \name. Specifically, we pick $\gamma$ in the set of $\{0.001, 0.005, 0.01, 0.05, 0.1\}$ and evaluate the corresponding adversarial accuracy of \name under Nettack with $1$ perturbation per target node. As shown in Figure \ref{figure:edge_p}, the improper choice of $\gamma$ may degrade the adversarial accuracy of \name by $3\%$. 
%However, picking $\gamma$ around $0.01$ can always achieve a reasonable adversarial accuracy on different datasets with different backbone GNN models. As a result, we recommend choosing $\gamma$ in the range of $[0.005, 0.05]$ in practice. In addition, Figure \ref{figure:edge_p} also shows that GPRGNN-\name is more robust to the changes of $\gamma$ than GCN-\name, which is due to the reason that the adaptive graph filter in GPRGNN can adapt to the graph structure after edge pruning.
It is worth noting that the proper value of $\gamma$ depends on the specific method adopted for measuring spectral embedding distortion. In other words, the optimal $\gamma$ will be different if we use the simplified version of spectral embedding distortion discussed in Appendix~\ref{sim_embed_diff}. Thus, we refer readers to our official configuration files at \href{https://github.com/cornell-zhang/GARNET/tree/main/configs}{github.com/cornell-zhang/GARNET/configs} for the proper choices of $\gamma$ based on the simplified distortion metric. 

\section{Backbone GNN Models for Defense}
\label{gnn}

As \name can be integrated with any existing GNN models to improve their adversarial accuracy, we choose two popular GNN models as the backbone model in our experiments: GCN and GPRGNN~\citep{kipf2016semi, chien2021adaptive}. As the GCN model implicitly assumes the underlying graph is homophilic, it performs poorly on heterophilic graphs~\citep{zhu2020beyond}. In contrast, GPRGNN can work on both homophilic and heterophilic datasets, due to its learned graph filter that can adapt to the homophily/heterophily property of the underlying graph. Thus, we choose GPRGNN as the backbone model for evaluation on heterophilic datasets. In addition, we also show the defense results with the H2GCN~\cite{zhu2020beyond} as backbone model in Appendix~\ref{h2gcn}.
%In addition, we add C\&S~\cite{huang2020combining} to GPRGNN on homophilic graphs (i.e., Cora and Pubmed) for all the defense baselines as well as \name.

\section{Defense Results with Various Perturbation Budgets}
\label{more_results}
\input{table/nettack}
\input{table/meta}

We provide additional defense results under Nettack and Metattack with various perturbations in Tables \ref{table:nettack} and \ref{table:meta} respectively. The results indicate that GARNET outperforms prior defense methods in most cases.

%\section{Broader Impact}
%\label{impact}
% \citet{zugner2018adversarial} have shown that graph adversarial attacks can drastically degrade the performance of GNN models for downstream applications. For instance, an attacker can attack a GNN-based recommender system on Facebook social network or Amazon co-purchasing network, via creating a fake account and make some connections to other users or items. Those connections can be viewed as adversarial edges in the graph. As a result, the attacker can deliberately enforce a GNN model to recommend some irrelevant or even harmful contents to other users. 
% Thus, improving adversarial robustness of GNN models has the potential for positive societal benefit. 
 
% We hope that this paper provides insight on the robustness and scalablity limitations of prior defense methods. Moreover, we believe that the proposed \name can largely overcome these two limitations and produce a robust GNN model against adversarial attacks on large-scale graph datasets. Nevertheless, we have to admit that \name may potentially provide the attacker with some hints about developing a even more powerful and scalable adversarial attack than all existing attacks, which is a possible negative consequence.
 
\section{Defense on H2GCN}
\label{h2gcn}

\input{table/h2gcn}
We provide the results of combining GARNET with H2GCN~\cite{zhu2020beyond} on heterophilic graphs in Table \ref{table:h2gcn}, which shows that GARNET achieves the highest accuracy in most cases and improves the accuracy on all perturbed graphs by a large margin compared to the vanilla H2GCN as well as H2GCN-TSVD. The results further confirm that GARNET is able to improve robustness of different backbone GNN models.

\section{Broader Impact}
\label{impact}
\citet{zugner2018adversarial} have shown that graph adversarial attacks can drastically degrade the performance of GNN models for downstream applications. For instance, an attacker can attack a GNN-based recommender system on Facebook social network or Amazon co-purchasing network, via creating a fake account and make some connections to other users or items. Those connections can be viewed as adversarial edges in the graph. As a result, the attacker can deliberately enforce a GNN model to recommend some irrelevant or even harmful contents to other users. 
Thus, improving adversarial robustness of GNN models has the potential for positive societal benefit. 
 
We hope that this paper provides insight on the robustness and scalablity limitations of prior defense methods. Moreover, we believe that the proposed \name can largely overcome these two limitations and produce a robust GNN model against adversarial attacks on large-scale graph datasets. Nevertheless, we have to admit that \name may potentially provide the attacker with some hints about developing an even more powerful and scalable adversarial attack than all existing attacks, which is a possible negative consequence.

%\section{Graph Rank Comparison}
%\label{rank}

\section{Discussion on Node Features}
\label{node_feat}
\subsection{Graph Construction with Node Features}
\input{table/node_feat.tex}
As \name purifies the adversarial graph by building a kNN graph based on dominant singular components, a natural question is whether the kNN graph constructed from node features can also achieve similar performance. We answer this question by comparing the results of \name graph and the node feature graph in Table~\ref{table:node_feat}. Note that the clean and adversarial accuracy are the same on the graph constructed from node features, since node features are unchanged after graph adversarial attack. Besides, we only show results on homophilic graphs as the kNN graph constructed from node features naturally falls into this category. Table~\ref{table:node_feat} shows that the node feature graph performs much worse than \name graph. This further confirms that the method proposed in this work is critical to improve the robustness of GNN models.

Apart from constructing the kNN graph purely from node features, we can also concatenate node features with dominant singular components for kNN graph construction, which may further improve the accuracy of \name. Note that we only adopt this notion for homophilic graphs, as this approach implicitly assumes that the graph is homophilic (nodes with similar features are adjacent in the kNN graph). The results in Table~\ref{table:node_feat} indicate that augmenting \name with node features can further improve the accuracy in several cases.

\subsection{Defense Against Node Feature Attack}
\name can be extended to handle node feature attack, although this paper mainly focuses on defending against graph structure attack, which we believe is more challenging than defending node feature attack due to the discrete nature. Specifically, we can perform TSVD to obtain the low-rank approximation of the node feature matrix, which can remove high-rank adversarial components in node features~\cite{entezari2020all}. The low-rank feature matrix is then concatenated to the weighted spectral embeddings to produce the kNN base graph. In this way, the downstream GNN model will be able to aggregate neighbors whose features are less perturbed during message passing.

\section{Additional Visualization Results}
\label{g_visual}
\input{figure/addition_visual.tex}
We visualize more target nodes and their local structures in Figure~\ref{figure:visual}, which reveals that \name consistently improves the quality of adversarial graph by removing adversarial edges that connect nodes with different labels. As a result, the adversarial accuracy of backbone GNN models can be largely improved once they are trained on the \name graph. 
%More importantly, compared to the clean graph, there are more neighbors with the same label as the target node in the \name graph, which makes the GNN model less likely to make a wrong prediction on the target node. This explains why \name can also improve clean accuracy in a few cases. 

\section{Acceleration of GARNET on Large Graphs}
\label{runtime_garnet}
%Apart from the runtime comparison of \name and defense baselines on small graphs in Figure~\ref{figure:runtime_comp}, we further evaluate the run time of \name on large (OGB) graphs. Concretely, the end-to-end run time of GARNET is 40 mins and 4 hours on ogbn-arxiv and ogbn-products, respectively, which is $3 \times$ faster than the most competitive baseline GNNGuard that takes more than 2 hours on ogbn-arxiv and 11 hours on ogbn-products. 
There are two major kernels in \name: (1) weighted spectral embedding (i.e., computing the top $r$ singular components or Laplacian eigenpairs), (2) kNN graph construction. For accelerating weighted spectral embedding, we can leverage the notion of multi-level graph coarsening~\cite{deng2020graphzoom, cheng2021spade} so that we only need to perform TSVD on the coarsest graph. To speedup the process of kNN graph construction, we can exploit Faiss to enable performing kNN on GPU~\cite{johnson2019billion}.

\section{Homophily Score of \name Graph}
\label{homo_score}
\input{table/garnet_homo_score.tex}
We follow~\citet{zhu2020beyond} to compute the homophily score per dataset (lower score means more heterophilic). As shown in Table~\ref{table:garnet_homo_score}, the \name graph is homophilic (heterophilic) if the corresponding clean graph is homophilic (heterophilic), which further confirms Theorem~\ref{embed_bound} that our approach can effectively recover the clean graph structure. As a result, \name supports both homophilic and heterophilic graphs.

\section{Accuracy of Clean Graph Recovery}
\label{acc_recover}
\input{table/edge_recover.tex}

Apart from visualizing \name graph in Figures~\ref{figure:visual_1255} and \ref{figure:visual}, we further quantify how well \name recovers the clean graph structure. Concretely, given a target node, we first extract nodes within its 2-hop neighbors in the clean graph and \name graph (under Metattack with $20\%$ perturbation ratio), respectively. By denoting the extracted nodes by $N_{clean}$ for clean graph and $N_{garnet}$ for \name graph, we define the recall score and precision score as follows:
$$Recall  = \frac{|N_{clean} \cap N_{garnet}|}{|N_{clean}|}$$
$$Precision  = \frac{|N_{clean} \cap N_{garnet}|}{|N_{garnet}|}$$
Table~\ref{table:edge_recover} shows the averaged recall and precision over $5$ nodes on Cora and Chameleon graphs. The results show that the recall scores are very high for both graphs, which indicates \name is able to accurately recover clean graph structure. The relatively low precision scores indicate that \name also introduces new edges to the graph (i.e., $|N_{garnet}| > |N_{clean}|$). We argue that those new edges are likely to connect spectrally similar nodes that are far away in the original clean graph, which enables \name to also incorporate global structural information. This explains why \name can sometimes outperform vanilla GNN models on clean heterophilic graphs (shown in Table~\ref{table:meta1}), where global structural information is very critical for node prediction.

\section{Further Discussion on Graph Recovery with PGM}
\label{eff_resist}
Intuitively, if we use more (clean) Laplacian eigenpairs (i.e., a larger $r$) for constructing the embedding matrix $V$ based on Definition~\ref{def}, the optimal solution for Equation~\ref{pgm} (i.e., $\Theta^*$) will be closer to the actual clean graph. In this section, we confirm this intuition based on graph resistance distances between node pairs. Specifically, consider the following expression for calculating effective-resistance distances between nodes $p$ and $q$ using all Laplacian eigenvalues/eigenvectors except $\lambda_1=0$: 

$$\sum_{i=2}^{|V|} \frac{(u_i^Te_{p,q})^2}{\lambda_i}$$ 

\citet{feng2021sgl} has shown that the effective-resistance distance between any node pair on the learned graph $\Theta^*$ (when $\sigma$ approaches infinity) will fully match the Euclidean distance between the corresponding data samples (i.e., rows in weighted spectral embedding matrix $V$ in our case). Moreover, it can be shown that the Euclidean distance between the data samples in our case will match the effective-resistance distance on the original graph when $r = |\mathcal{V}|$ (with proper normalization on Laplacian eigenvectors). As a result, the resistance distances on the learned graph $\Theta^*$ will be the same as the ones on the original graph when $r=|\mathcal{V}|$. Moreover, using a larger $r$ value will lead to a more accurate estimation of the learned (clean) graph. 

In practice, if $r$ satisfies that $\lambda_r \ll \lambda_{r+1}$, dropping the terms with much larger eigenvalues (i.e., $\lambda_{r+1}$, $\lambda_{r+2}$, ..., $\lambda_{|\mathcal{V}|}$) will not significantly impact the approximation accuracy. A proper $r$ can be effectively determined based on the strategy proposed in Appendix~\ref{algo}. We leave the theoretical guarantee of other metrics for graph comparison to our future work.

\section{Connection Between kNN Graph and TSVD Graph}
\label{knn_svd}
Apart from the motivation of constructing a kNN graph as $\mathcal{G}_{base}$ based on Theorem~\ref{embed_bound}, we further motivate the kNN graph construction from the perspective of improving the scalability of TSVD-based defense methods. Concretely, as previous TSVD-based methods produce a dense (low-rank) adjacency matrix $\hat{A}$, they involve dense matrices during GNN training, which has quadratic time/space complexity and thus cannot scale to large graphs. A potential solution is to sparsify $\hat{A}$ by preserving the top $k$ largest elements per row. However, na\"ively selecting the largest elements of each row in $\hat{A}$ requires forming/storing $\hat{A}$ first, which still has quadratic time/space complexity. In contrast, we leverage the (approximate) kNN algorithm to construct the sparsified $\hat{A}$ by taking as input the weighted spectral embedding $V$ (note that ${\hat{A}} = VV^T$ based on Proposition~\ref{theory_embed}). Consequently, our kNN graph construction step can also be viewed as a scalable way of sparsifying the dense adjacency matrix $\hat{A}$ generated by TSVD. Moreover, Theorem~\ref{embed_bound} theoretically guarantees that the sparsified graph serves as a reasonable $\mathcal{G}_{base}$ for edge pruning.

\section{Simplified Version of Spectral Embedding Distortion}
\label{sim_embed_diff}

To obtain $s_{i,j} = \frac{\|U^Te_{i,j}\|_2^2}{\|V^Te_{i,j}\|_2^2}$, we have to compute both the top $r$ Laplacian eigenpairs of $\mathcal{G}_{base}$ (for constructing $U$) and those of $\mathcal{G}_{adv}$ (for constructing $V$). However, as $\mathcal{G}_{base}$ is relatively dense (a large $k$ used for kNN graph construction), computing its Laplacian eigenpairs is time-consuming, especially on large graphs. Fortunately, ~\citet{von2007tutorial} has shown that the top $r$ Laplacian eigenvectors (corresponding to smallest eigenvalues) only vary a little across nodes in a dense graph (i.e., they are smooth over the graph), which means $\|U^Te_{i,j}\|_2$ is very similar across different edges in $\mathcal{G}_{base}$. Consequently, the term $\|V^Te_{i,j}\|_2$ becomes the dominant factor in $s_{i,j}$ to identify (non)critical edges: a large value of $\|V^Te_{i,j}\|_2$ means a small value of $s_{i,j}$, indicating the corresponding edge is noncritical. As we empirically find out that exploiting $\|V^Te_{i,j}\|_2$ to prune noncritical edges does not degrade the accuracy of \name, we adopt this simplified version of spectral embedding distortion in our implementation.

%% file: table/hyperpara.tex
\begin{table}[ht]
\centering
%\vspace{20pt}
\caption{Summary of hyperparameters in \name --- We denote the number of eigenpairs for spectral embedding and nearest neighbors for base graph construction
%, and the threshold for edge pruning 
by $r$ and $k$
%, and $\gamma$
, respectively.}
\label{table:hyperpara}
\scalebox{0.5}{
\begin{adjustbox}{width=0.9\columnwidth,center}
\begin{tabular}[t]{lccccccccc}
\toprule
\textbf{Dataset}   & $r$   & $k$     \\
\midrule
Cora-Nettack   &  $50$ &  $30$ \\
Cora-Metattack   &  $50$ &  $30$ \\
Pubmed-Nettack   &  $50$ &  $50$ \\
Pubmed-Metattack   &  $50$ &  $50$ \\
Chameleon-Nettack   &  $50$ &  $50$ \\
Chameleon-Metattack   &  $50$ &  $50$ \\
Squirrel-Nettack   &  $50$ &  $50$ \\
Squirrel-Metattack   &  $50$ &  $50$ \\
ogbn-arxiv-GRBCD   &  $500$ &  $50$ \\
ogbn-products-GRBCD   &  $500$ &  $50$ \\
\bottomrule
\end{tabular}
\end{adjustbox}
}
\end{table}

%% file: table/algorithm.tex
\SetAlFnt{\large}
\SetAlCapFnt{\large}
\SetAlCapNameFnt{\large}
\IncMargin{1em}
\begin{algorithm}[H]
\LinesNumbered
\KwIn{Adversarial graph $\mathcal{G}_{adv}$; node feature matrix $X \in R^{n \times d}$; prior data variance $\sigma^2$; truncated svd rank $r$; kNN graph $k$; threshold for edge pruning $\gamma$; a GNN model for defense.}
\KwOut{Node embedding matrix $Z \in R^{n \times c}$}
%\SetAlgoLined
    $M, S = eigs(\mathcal{G}_{adv}, r$); \\
    $V = M \sqrt{\left|I-S\right|}$; \\
    $V = concat(V, X)$; \Comment{optional} \\
    $\mathcal{G}_{base} = kNN\_graph(V, k)$; \\
    $M', S'= eigs(\mathcal{G}_{base}, r$); \Comment{optional} \\
    $U = M' / \sqrt{S'+I/\sigma^2}$; \Comment{optional} \\
    \For{$e_{i,j} \in \mathcal{G}_{base}$}{
        \If{$\frac{\|U_i-U_j\|_2^2}{\|V_i-V_j\|_2^2} < \gamma$ (or simplified version: $\|V_i-V_j\|_2^2 > \gamma'$)}
        {
            Prune $e_{i,j}$ from $\mathcal{G}_{base}$; \\
        }
    }
   %\FOR{edge $e_{i,j}$ {\bfseries in} $\mathcal{G}_{base}$}
   %\IF{$\frac{\|U_i-U_j\|_2^2}{\|V_i-V_j\|_2^2} < \gamma$}
    $Z$ = GNN($\mathcal{G}_{base}'$, $X$);
 \caption{\name based adversarial defense (this work)}
 \label{algo_table}
\end{algorithm}

\begin{algorithm}[H]
\LinesNumbered
\KwIn{Adversarial graph $\mathcal{G}_{adv}$; node feature matrix $X \in R^{n \times d}$; truncated svd rank $r$; a GNN model for defense.}
\KwOut{Node embedding matrix $Z \in R^{n \times c}$}
%\SetAlgoLined
   $U, S, V = TSVD(\mathcal{G}_{adv}, r$); \\
   $A_{\mathcal{G}_{tsvd}} = USV^T$; \\
   $Z$ = GNN($\mathcal{G}_{tsvd}$, $X$);
   \caption{Truncated SVD based adversarial defense (prior work)}
   \label{algo_tsvd}
\end{algorithm}

%% file: figure/knn.tex
\begin{figure}[ht!]
\begin{center}
	\includegraphics[width=\textwidth]{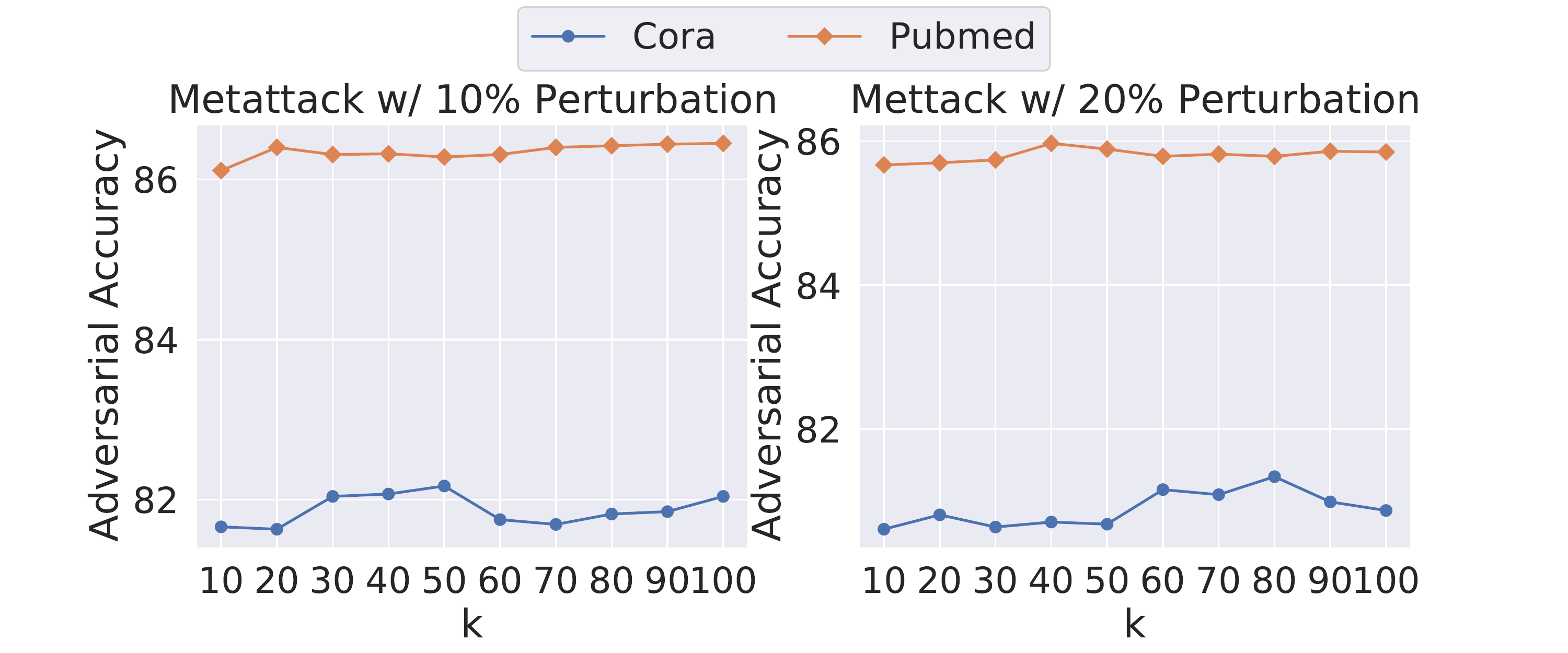}
	\caption{Ablation study of \name on $k$ for kNN graph construction. \label{figure:knn}}
\end{center}
\vspace{-12pt}
\end{figure}

%% file: figure/edge_prune.tex
\begin{figure}[htb]
\begin{center}
	\includegraphics[width=\textwidth]{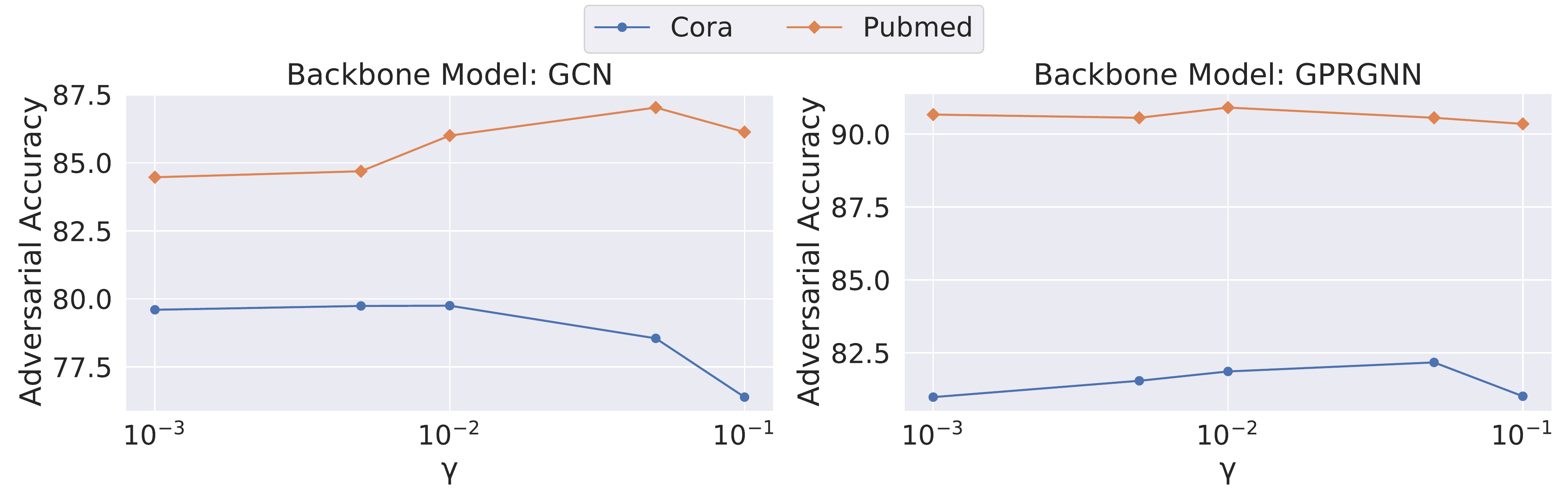}
	\caption{Ablation study of \name on threshold $\gamma$ for edge pruning. \label{figure:edge_p}}
\end{center}
\vspace{-12pt}
\end{figure}

%% file: table/nettack.tex
\begin{table*}[ht!]
  \begin{center}
    \caption{Averaged node classification accuracy (\%) $\pm$ std under targeted attack (Nettack) with different perturbation ratio --- We denote the evaluated dataset by its name with the number of perturbations (e.g., Cora-0 means the clean Cora graph and Cora-1 denotes there is $1$ adversarial edge perturbation per target node). As GCN is not designed for heterophilic graphs, we only show results of defense methods with GPRGNN as the backbone model on Chameleon and Squirrel. We bold and underline the first and second highest accuracy of each backbone GNN model, respectively. $OOM$ means out of memory.}
    %\vspace{8pt}
    \label{table:nettack}
    \begin{adjustbox}{width=\columnwidth,center}
    \begin{NiceTabular}{c|cccc|cccc}
      %\toprule % <-- Toprule here
      %Dataset & GCN & GPR & GCNSVD & GPRSVD & ProGCN & ProGPR & \name GCN & \name GPR \\
      %$\alpha$ & $\beta$ & $\gamma$ \\
      %\midrule % <-- Midrule here
      \toprule
 &  \multicolumn{4}{c}{GCN} & \multicolumn{4}{c}{GPRGNN}\\
\cmidrule(r){2-5} \cmidrule(r){6-9}
Dataset   &  Vanilla   &  TSVD  &  ProGNN   &  \name  &  Vanilla   &  TSVD  &  ProGNN &  \name  \\
\midrule
      Cora-0 & $\underline{80.96} \pm 0.95$ & $72.65 \pm 2.29$ & $80.54 \pm 1.21$ & $\mathbf{81.08} \pm 2.05$ & $\textbf{83.04} \pm 2.05$ & $81.68 \pm 1.78$ & $82.04 \pm 1.33$ & $\underline{82.77} \pm 1.89$ \\
      Cora-1 & $70.06 \pm 0.81$ & $71.36 \pm 1.63$ & $\mathbf{81.65} \pm 0.59$ & $\underline{79.75} \pm 2.35$ & $\underline{81.68} \pm 2.18$ & $79.36 \pm 2.23$ & $80.56 \pm 1.71$ & $\mathbf{82.17} \pm 1.95$ \\
      Cora-2 & $68.60 \pm 1.81$ & $70.66 \pm 2.76$ & $\mathbf{79.83} \pm 1.10$ & $\underline{79.69} \pm 1.50$ & $74.34 \pm 2.41$ & $\underline{76.26} \pm 2.34$ & $76.12 \pm 2.43$ & $\mathbf{78.55} \pm 2.11$ \\
      Cora-3 & $65.04 \pm 3.31$ & $68.20 \pm 1.93$ & $\underline{72.08} \pm 1.20$ & $\mathbf{74.42} \pm 2.06$ & $70.96 \pm 2.00$ & $70.90 \pm 3.89$ & $\underline{73.74} \pm 2.73$ & $\mathbf{79.40} \pm 1.35$ \\
      Cora-4 & $61.69 \pm 1.48$ & $65.34 \pm 3.46$ & $\underline{67.83} \pm 1.87$ & $\mathbf{69.60} \pm 2.67$ & $65.90 \pm 1.61$ & $65.51 \pm 3.27$ & $\underline{68.94} \pm 3.25$ & $\mathbf{72.77} \pm 2.16$ \\
      Cora-5 & $55.66 \pm 1.95$ & $60.30 \pm 2.25$ & $\underline{65.38} \pm 1.65$ & $\mathbf{67.04} \pm 2.05$ & $62.89 \pm 1.95$ & $63.52 \pm 3.27$ & $\underline{63.74} \pm 2.57$ & $\mathbf{71.45} \pm 2.73$ \\
      \midrule
      Pubmed-0 & $87.26 \pm 0.51$ & $87.03 \pm 0.48$ & $\mathbf{88.14} \pm 1.44$ & $\underline{87.96} \pm 0.58$ & $\underline{90.05} \pm 0.73$ & $OOM$ & $OOM$ & $\mathbf{90.99} \pm 0.52$ \\
      Pubmed-1 & $84.29 \pm 0.68$ & $\underline{86.46} \pm 0.28$ & $85.75 \pm 1.23$ & $\mathbf{87.03} \pm 0.68$ & $\underline{89.30} \pm 0.54$ & $OOM$ & $OOM$ & $\mathbf{90.91} \pm 0.47$ \\
      Pubmed-2 & $82.17 \pm 0.67$ & $\underline{83.68} \pm 0.46$ & $81.23 \pm 1.21$ & $\mathbf{86.92} \pm 0.45$ & $\underline{87.42} \pm 0.28$ & $OOM$ & $OOM$ & $\mathbf{90.75} \pm 0.55$ \\
      Pubmed-3 & $81.13 \pm 0.53$ & $\underline{81.34} \pm 0.68$ & $80.65 \pm 1.39$ & $\mathbf{86.50} \pm 0.45$ & $\underline{84.46} \pm 0.53$ & $OOM$ & $OOM$ & $\mathbf{90.70} \pm 0.37$ \\
      Pubmed-4 & $75.48 \pm 0.52$ & $\underline{82.41} \pm 0.54$ & $78.46 \pm 1.11$ & $\mathbf{86.44} \pm 0.64$ & $\underline{81.72} \pm 0.72$ & $OOM$ & $OOM$ & $\mathbf{90.11} \pm 0.57$ \\
      Pubmed-5 & $66.67 \pm 1.34$ & $\underline{79.56} \pm 0.48$ & $71.89 \pm 1.56$ & $\mathbf{86.12} \pm 0.86$ & $\underline{76.99} \pm 1.16$ & $OOM$ & $OOM$ & $\mathbf{89.52} \pm 0.45$ \\
      \midrule
      Chameleon-0 & \Block{6-4}{\diagbox{}{}} & & & & $\underline{71.46} \pm 1.92$ & $62.12 \pm 3.04$ & $58.80 \pm 1.72$& $\mathbf{72.89} \pm 2.65$ \\
      Chameleon-1 & & & & & $\underline{71.02} \pm 1.57$ & $61.34 \pm 2.93$ & $58.05 \pm 1.90$& $\mathbf{72.68} \pm 1.89$ \\
      Chameleon-2 & & & & & $\underline{70.71} \pm 1.12$ & $61.09 \pm 2.80$ & $57.44 \pm 1.67$& $\mathbf{72.20} \pm 2.31$ \\
      Chameleon-3 & & & & & $\underline{70.30} \pm 1.28$ & $60.98 \pm 2.82$ & $57.19 \pm 1.83$& $\mathbf{72.17} \pm 2.07$ \\
      Chameleon-4 & & & & & $\underline{69.87} \pm 1.29$ & $60.85 \pm 3.31$ & $57.44 \pm 1.63$& $\mathbf{72.06} \pm 2.94$ \\
      Chameleon-5 & & & & & $\underline{66.26} \pm 1.71$ & $60.37 \pm 2.86$ & $57.07 \pm 1.82$& $\mathbf{71.83} \pm 2.11$ \\
      \midrule
      %\cmidrule{1} \cmidrule{6-9}
      Squirrel-0 & \Block{6-4}{\diagbox{}{}} & & & & $\underline{41.36} \pm 2.87$ & $32.98 \pm 2.36$ & $31.81 \pm 1.72$& $\mathbf{44.91} \pm 1.53$ \\
      Squirrel-1 & & & & & $\underline{41.27} \pm 3.16$ & $32.63 \pm 0.87$ & $30.54 \pm 2.45$& $\mathbf{43.55} \pm 1.79$ \\
      Squirrel-2 & & & & & $\underline{41.09} \pm 2.14$ & $32.05 \pm 1.05$ & $30.73 \pm 2.13$& $\mathbf{44.09} \pm 2.35$ \\
      Squirrel-3 & & & & & $\underline{40.98} \pm 2.72$ & $32.00 \pm 1.66$ & $30.25 \pm 1.98$& $\mathbf{44.18} \pm 2.26$ \\
      Squirrel-4 & & & & & $\underline{40.25} \pm 2.82$ & $31.45 \pm 1.38$ & $29.09 \pm 2.33$& $\mathbf{43.73} \pm 1.62$ \\
      Squirrel-5 & & & & & $\underline{39.45} \pm 2.36$ & $31.20 \pm 1.84$ & $27.27 \pm 1.87$& $\mathbf{43.64} \pm 1.53$ \\
      \bottomrule % <-- Bottomrule here
    \end{NiceTabular}
    \end{adjustbox}
  \end{center}
  \vspace{-8pt}
\end{table*}

%% file: table/meta.tex
\begin{table*}[ht]
  \begin{center}
    \caption{Averaged node classification accuracy (\%) $\pm$ std under non-targeted attack (Metattack) with different perturbation ratio --- We denote the evaluated dataset by its name with the perturbation ratio (e.g., Cora-0 means the clean Cora graph and Cora-10 denotes there are $10\%$ adversarial edges). As GCN is not designed for heterophilic graphs, we only show results of defense methods with GPRGNN as the backbone model on Chameleon and Squirrel. We bold and underline the first and second highest accuracy of each backbone GNN model, respectively. $OOM$ means out of memory.}
    %\vspace{8pt}
    \label{table:meta}
    \begin{adjustbox}{width=\columnwidth,center}
    \begin{NiceTabular}{c|cccc|cccc}
      %\toprule % <-- Toprule here
      %Dataset & GCN & GPR & GCNSVD & GPRSVD & ProGCN & ProGPR & \name GCN & \name GPR \\
      %$\alpha$ & $\beta$ & $\gamma$ \\
      %\midrule % <-- Midrule here
      \toprule
 &  \multicolumn{4}{c}{GCN} & \multicolumn{4}{c}{GPRGNN}\\
\cmidrule(r){2-5} \cmidrule(r){6-9}
Dataset   &  Vanilla   &  TSVD  &  ProGNN   &  \name  &  Vanilla   &  TSVD  &  ProGNN   &  \name  \\
\midrule
      Cora-0 & $\mathbf{81.35} \pm 0.66$ & $73.86 \pm 0.53$ & $78.56 \pm 0.36$ & $\underline{79.64} \pm 0.75$ & $\mathbf{83.05} \pm 0.42$ & $81.61 \pm 0.54$ & $82.04 \pm 0.90$ & $\underline{82.67} \pm 1.89$ \\
      Cora-10 & $69.50 \pm 1.46$ & $69.45 \pm 0.69$ & $\mathbf{77.90} \pm 0.69$ & $\underline{77.78} \pm 0.53$ & $80.37 \pm 0.65$ & $\underline{81.08} \pm 0.52$ & $80.31 \pm 1.23$ & $\mathbf{82.17} \pm 0.69$ \\
      Cora-20 & $56.28 \pm 1.19$ & $62.44 \pm 1.16$ & $\underline{72.28} \pm 1.67$ & $\mathbf{73.89} \pm 0.91$ & $74.27 \pm 2.11$ & $\underline{78.50} \pm 1.20$ & $76.29 \pm 1.46$ & $\mathbf{81.34} \pm 0.79$ \\
      \midrule
      Pubmed-0 & $\mathbf{87.16} \pm 0.09$ & $84.53 \pm 0.08$ & $84.62 \pm 0.11$& $\underline{85.37} \pm 0.20$ & $\mathbf{87.35} \pm 0.13$ & $OOM$ & $OOM$ & $\underline{86.86} \pm 0.57$ \\
      Pubmed-10 & $81.16 \pm 0.13$ & $\underline{84.56} \pm 0.10$ & $84.09 \pm 0.12$& $\mathbf{85.22} \pm 0.13$ & $\underline{85.52} \pm 0.14$ & $OOM$ & $OOM$ & $\mathbf{86.24} \pm 0.20$ \\
      Pubmed-20 & $77.20 \pm 0.27$ & $\underline{84.30} \pm 0.08$ & $83.89 \pm 0.32$& $\mathbf{85.14} \pm 0.23$ & $\underline{84.18} \pm 0.15$ & $OOM$ & $OOM$ & $\mathbf{85.69} \pm 0.26$ \\
      \midrule % <-- Midrule here
      Chameleon-0 & \Block{3-4}{\diagbox{}{}} & & & & $\mathbf{61.36} \pm 1.00$ & $47.29 \pm 1.63$ & $48.39 \pm 0.68$ & $\underline{61.11} \pm 2.46$ \\
      Chameleon-10 & & & & & $\underline{57.55} \pm 1.26$ & $47.07 \pm 1.21$ & $47.80 \pm 0.91$& $\mathbf{60.96} \pm 1.22$ \\
      Chameleon-20 & & & & & $\underline{53.20} \pm 0.88$ & $45.12 \pm 1.34$ & $46.69 \pm 0.61$ & $\mathbf{59.96} \pm 0.84$ \\
      \midrule
      %\midrule
      Squirrel-0 & \Block{3-4}{\diagbox{}{}} & & & & $\underline{39.51} \pm 1.64$ & $31.36 \pm 1.87$ & $31.64 \pm 2.87$ & $\mathbf{43.43} \pm 1.14$ \\
      Squirrel-10 & & & & & $\underline{38.27} \pm 0.83$ & $28.25 \pm 1.66$ & $30.33 \pm 3.29$& $\mathbf{42.62} \pm 1.09$ \\
      Squirrel-20 & & & & & $\underline{35.22} \pm 1.20$ & $23.91 \pm 1.40$ & $29.36 \pm 3.61$& $\mathbf{41.97} \pm 1.02$ \\
      \bottomrule % <-- Bottomrule here
    \end{NiceTabular}
    \end{adjustbox}
  \end{center}
  \vspace{-10pt}
\end{table*}

%% file: table/h2gcn.tex
\begin{table*}[ht!]
  \begin{center}
    \caption{Averaged node classification accuracy (\%) $\pm$ std on heterophilic graphs --- We bold and underline the first and second highest accuracy, respectively. The backbone GNN model is H2GCN.}
    %\vspace{-5pt}
    \label{table:h2gcn}
    \begin{adjustbox}{width=\columnwidth,center}
    {\renewcommand{\arraystretch}{1.1}
    \setlength\tabcolsep{1.5 pt}
    \begin{NiceTabular}{l|cc|cc|cc|cc}
      \toprule
 &  \multicolumn{2}{c}{Chameleon (Nettack)} & \multicolumn{2}{c}{Chameleon (Metattack)}
 &  \multicolumn{2}{c}{Squirrel (Nettack)} & \multicolumn{2}{c}{Squirrel (Metattack)}\\
\cmidrule(r){2-3} \cmidrule(r){4-5} \cmidrule(r){6-7} \cmidrule(r){8-9}
Model   &  Clean   &  Adversarial  &  Clean   &  Adversarial  &  Clean   &  Adversarial &  Clean   &  Adversarial  \\
\midrule
      Vanilla & $\underline{78.43} \pm 2.09$ & $62.20 \pm 1.99$ & $\mathbf{68.45} \pm 0.57$ & $52.73 \pm 1.72$ & $\mathbf{55.36} \pm 2.91$ & $29.55 \pm 3.09$ & $\mathbf{61.23} \pm 0.71$ & $\underline{44.84} \pm 0.89$ \\
      TSVD & $67.07 \pm 1.15$ & $\underline{63.17} \pm 1.61$ & $61.75 \pm 1.09$ & $\underline{54.06} \pm 1.66$ & $32.45 \pm 1.87$ & $\underline{31.64} \pm 2.09$ & $46.66 \pm 1.71$ & $40.56 \pm 1.41$ \\
      GARNET & $\mathbf{78.78} \pm 1.84$ & $\mathbf{76.10} \pm 1.92$ & $\underline{66.63} \pm 1.05$ & $\mathbf{61.12} \pm 0.59$ & $\underline{54.09} \pm 1.73$ & $\mathbf{53.27} \pm 1.50$ & $\underline{59.67} \pm 0.83$ & $\mathbf{50.08} \pm 1.92$ \\
      \bottomrule % <-- Bottomrule here
    \end{NiceTabular}}
    \end{adjustbox}
  \end{center}
\end{table*}

%% file: table/node_feat.tex
\begin{table*}[ht!]
  \begin{center}
    \caption{Averaged node classification accuracy (\%) $\pm$ std under targeted attack (Nettack) and non-targeted attack (Metattack) on Cora and Pubmed --- We bold and underline the first and second highest accuracy, respectively. ``NodeFeat'' denotes the kNN graph constructed from node features is used for GNN training. ``GARNET w/ NodeFeat'' denotes the kNN graph constructed from the concatenation of dominant singular components and node features. The backbone model is GCN.}
    %\vspace{-5pt}
    \label{table:node_feat}
    \begin{adjustbox}{width=\columnwidth,center}
    {\renewcommand{\arraystretch}{1.1}
    \setlength\tabcolsep{1.5 pt}
    \begin{NiceTabular}{c|cc|cc|cc|cc}
      \toprule
 &  \multicolumn{2}{c}{Cora (Nettack)} & \multicolumn{2}{c}{Cora (Metattack)}
 &  \multicolumn{2}{c}{Pubmed (Nettack)} & \multicolumn{2}{c}{Pubmed (Metattack)}\\
\cmidrule(r){2-3} \cmidrule(r){4-5} \cmidrule(r){6-7} \cmidrule(r){8-9}
Model   &  Clean   &  Adversarial  &  Clean   &  Adversarial  &  Clean   &  Adversarial &  Clean   &  Adversarial  \\
\midrule
      Vanilla & $80.96 \pm 0.95$ & $55.66 \pm 1.95$ & $\underline{81.35} \pm 0.66$ & $56.28 \pm 1.19$ & $87.26 \pm 0.51$ & $66.67 \pm 1.34$ & $\mathbf{87.16} \pm 0.09$ & $77.20 \pm 0.27$ \\
      NodeFeat & $52.65 \pm 2.69$ & $52.65 \pm 2.69$ & $56.44 \pm 1.04$ & $56.44 \pm 1.04$ & $83.01 \pm 0.99$ & $83.01 \pm 0.99$ & $78.66 \pm 0.15$ & $78.66 \pm 0.15$ \\
      GARNET & $\underline{81.08} \pm 2.05$ & $\mathbf{67.04} \pm 2.05$ & $79.64 \pm 0.75$ & $\mathbf{73.89} \pm 0.91$ & $\underline{87.96} \pm 0.58$ & $\underline{86.12} \pm 0.86$ & $85.37 \pm 0.20$ & $\mathbf{85.14} \pm 0.23$ \\
      GARNET w/ NodeFeat & $\mathbf{83.73} \pm 1.17$ & $\underline{64.35} \pm 2.98$ & $\mathbf{81.93} \pm 0.44$ & $\underline{71.97} \pm 0.95$ & $\mathbf{88.76} \pm 0.40$ & $\mathbf{86.56} \pm 0.62$ & $\underline{86.16} \pm 0.16$ & $\underline{84.88} \pm 0.34$ \\
      \bottomrule % <-- Bottomrule here
    \end{NiceTabular}}
    \end{adjustbox}
  \end{center}
  %\vspace{-5pt}
\end{table*}

%% file: figure/addition_visual.tex
\begin{figure*}[ht]% [hpbt] what you need
    \centering
    %\vspace{-10pt}
    \subfigure[]{%
        \includegraphics[width=0.32\textwidth]{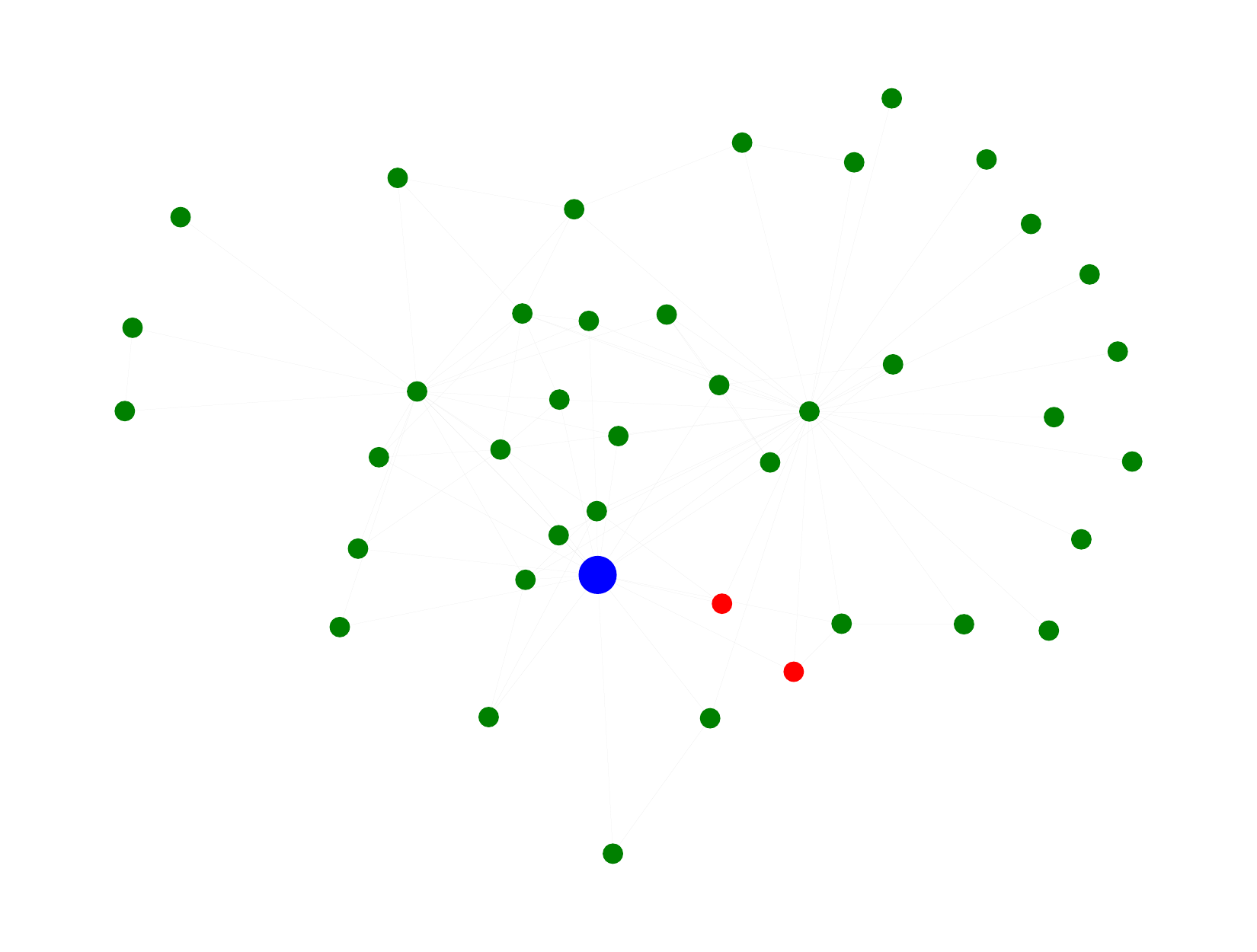}}
    \subfigure[]{%
        \includegraphics[width=0.32\textwidth]{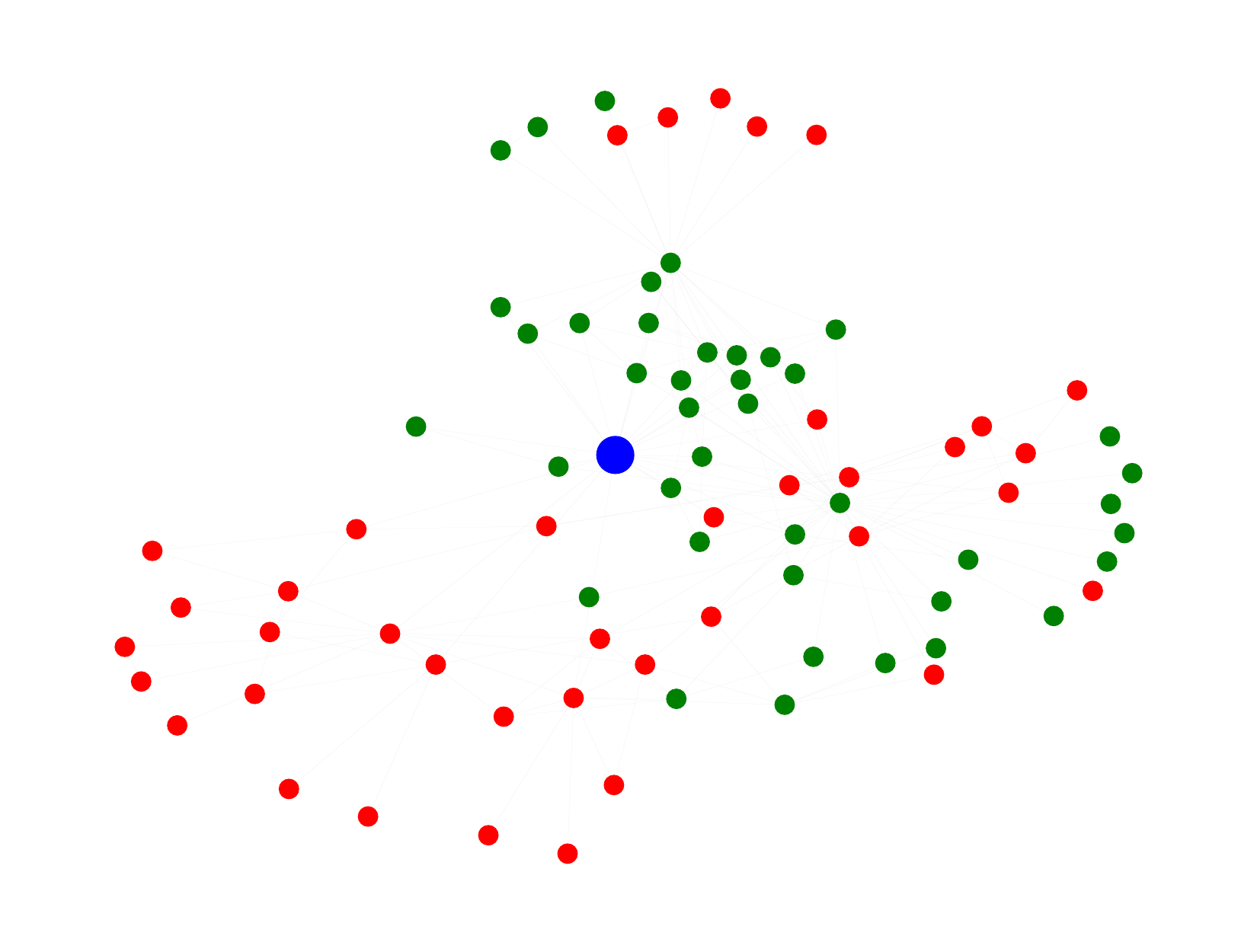}}
    \subfigure[]{%
        \includegraphics[width=0.32\textwidth]{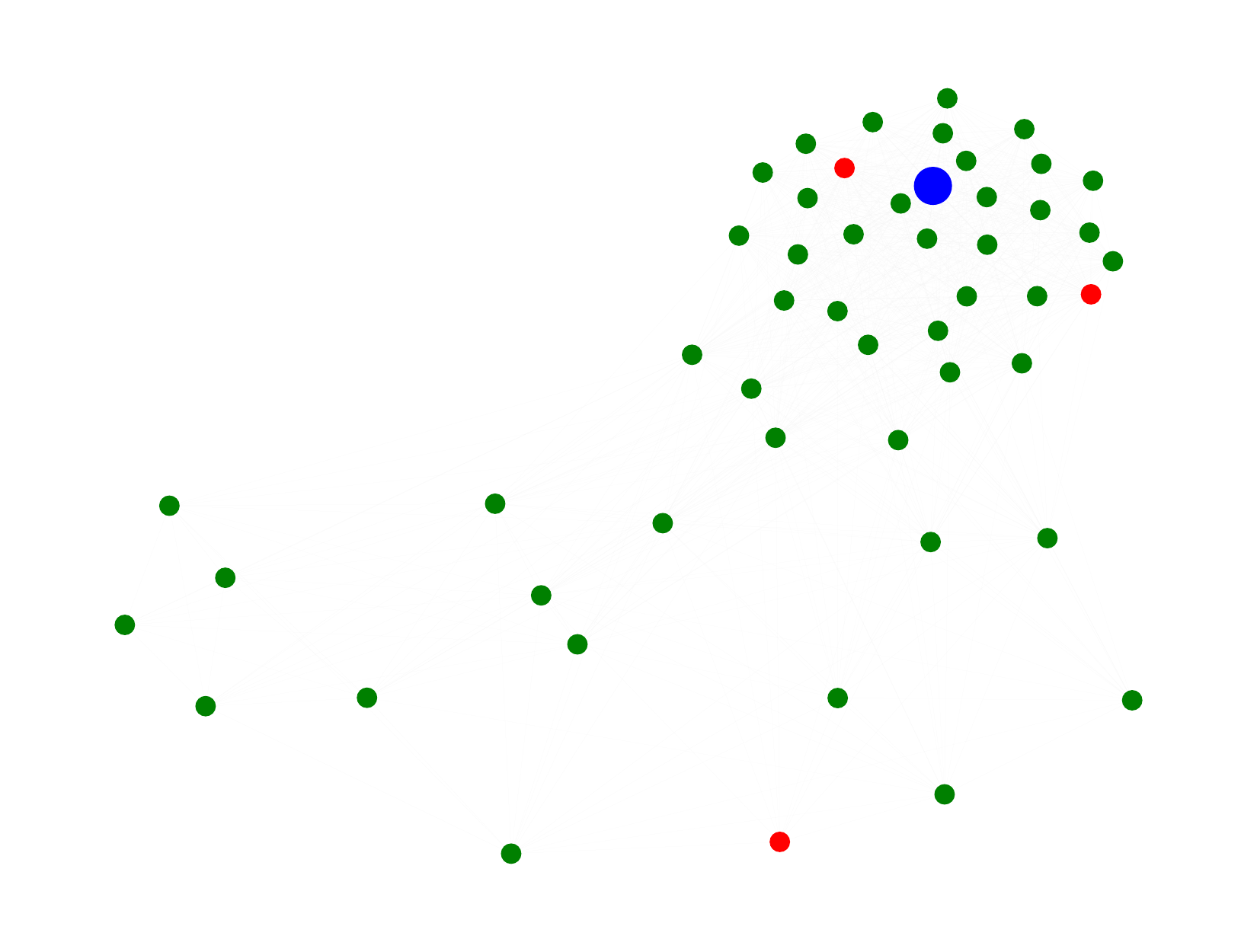}}
    \hfill
    \subfigure[]{%
        \includegraphics[width=0.32\textwidth]{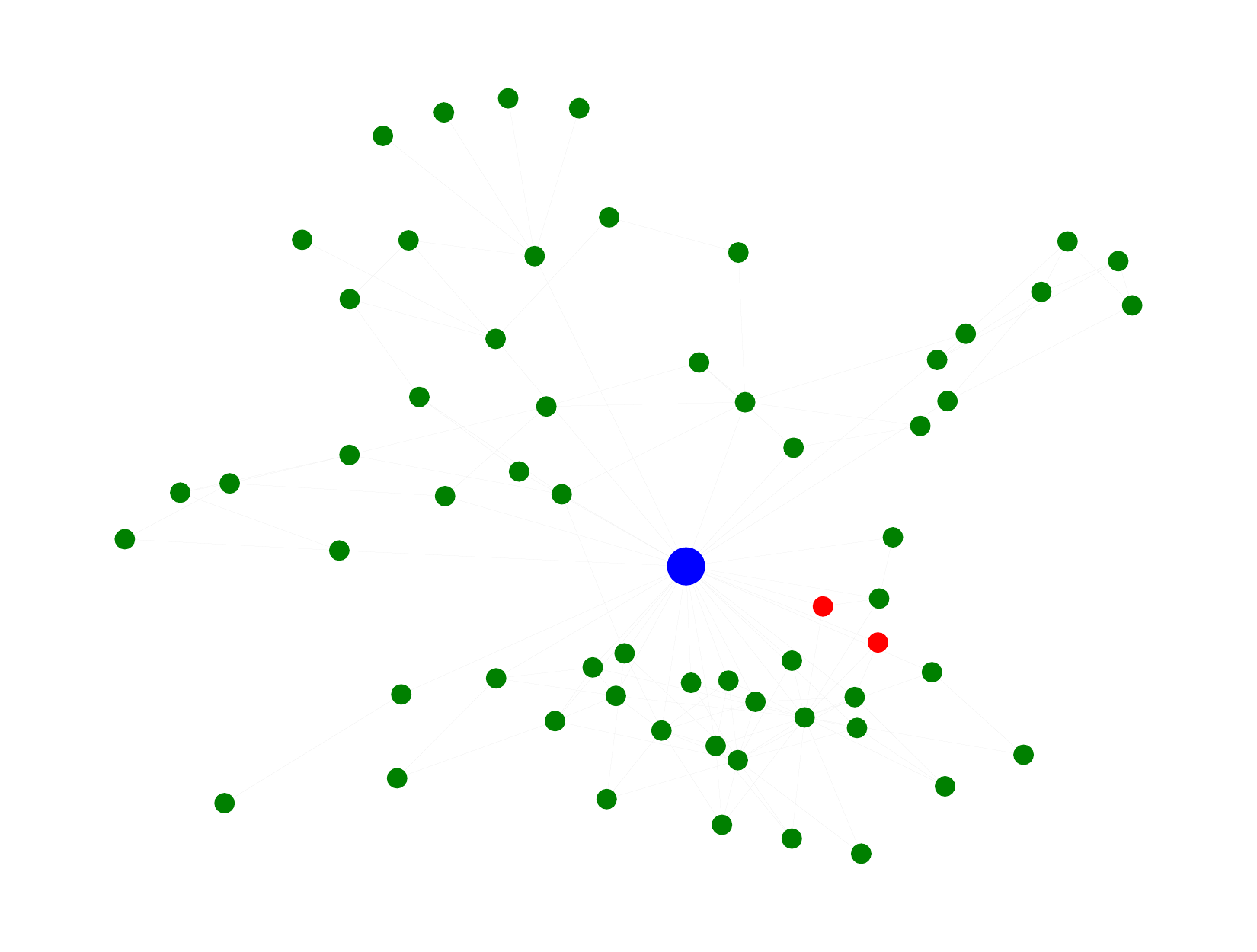}}
    \subfigure[]{%
        \includegraphics[width=0.32\textwidth]{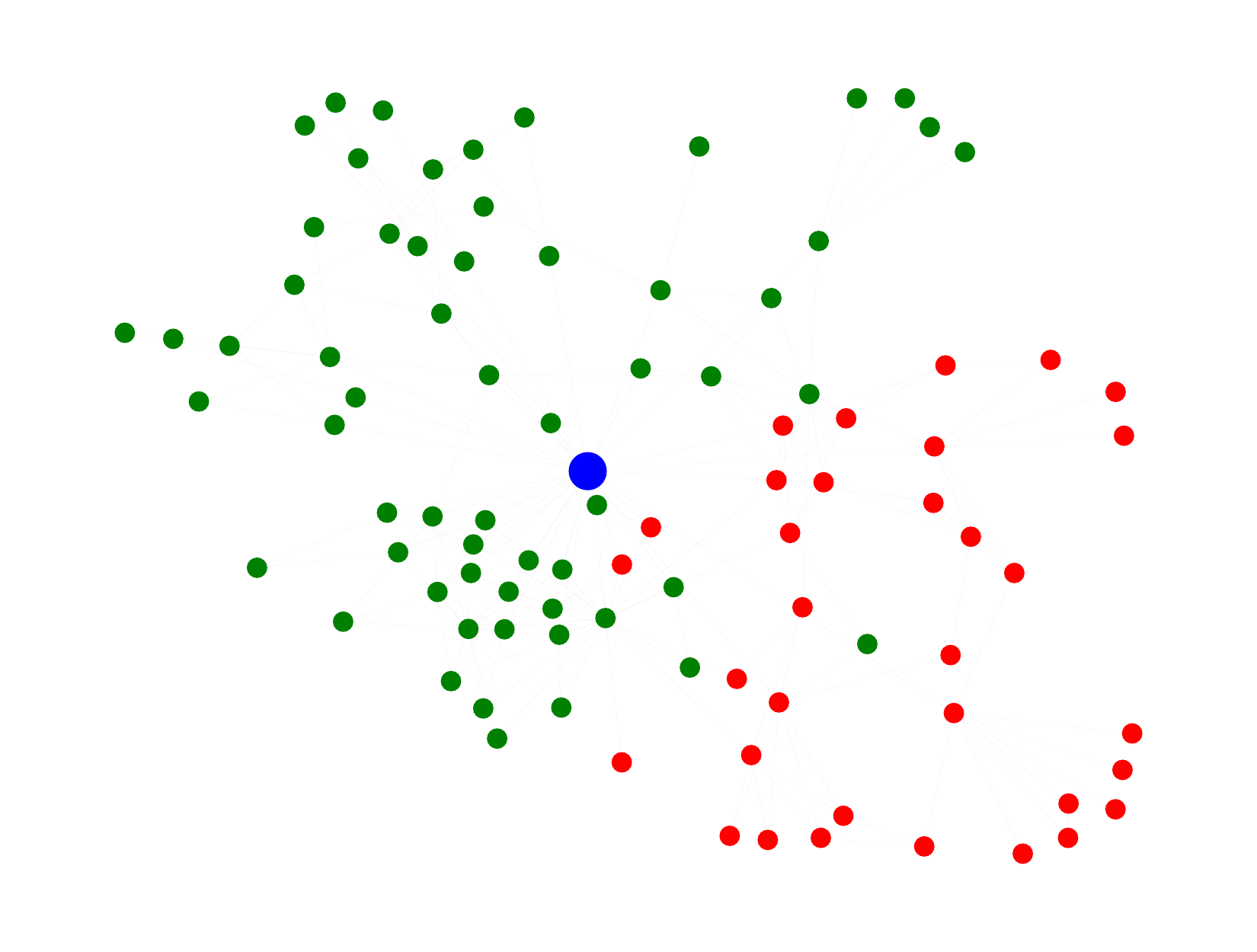}}
    \subfigure[]{%
        \includegraphics[width=0.32\textwidth]{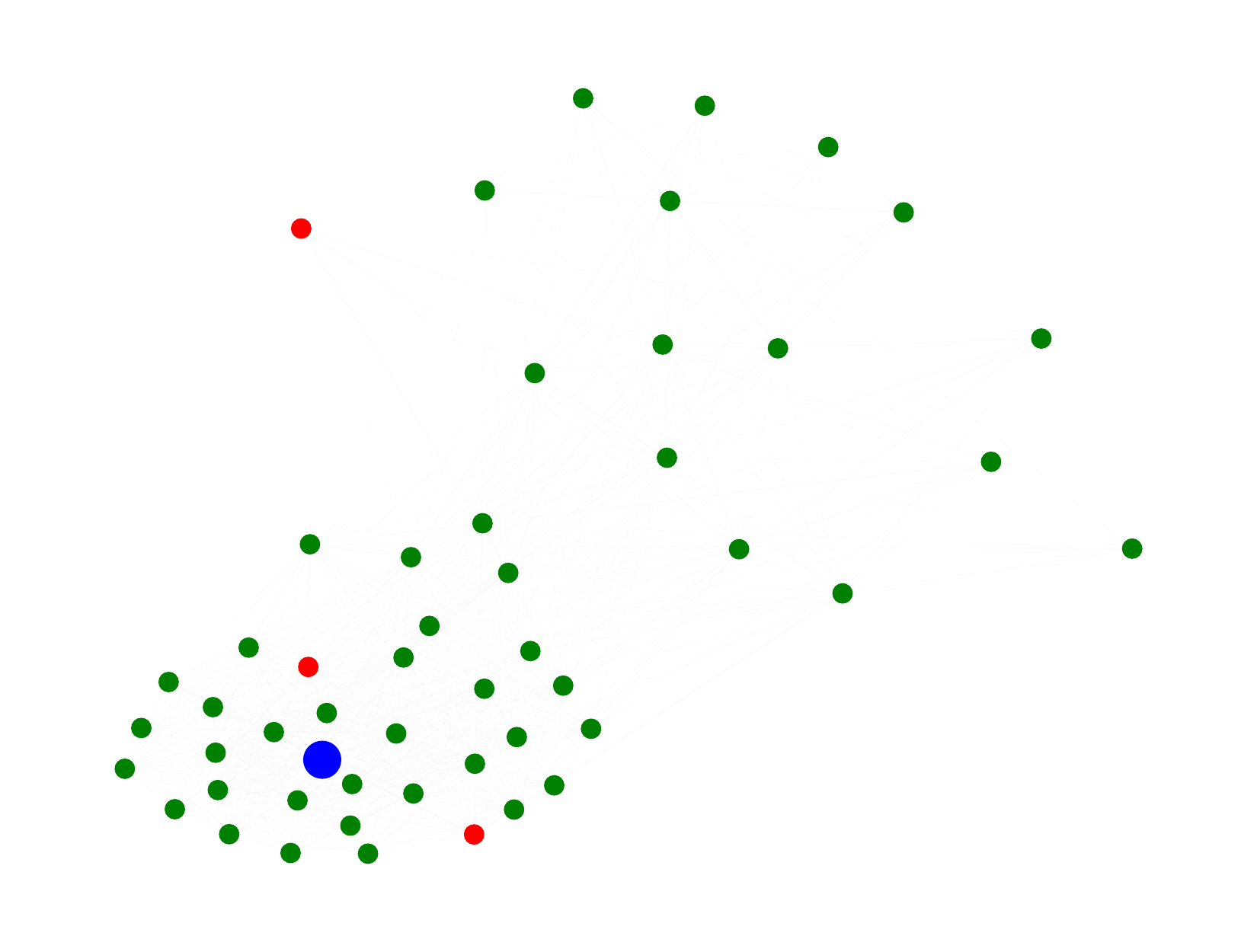}}
    \hfill
    \subfigure[]{%
        \includegraphics[width=0.32\textwidth]{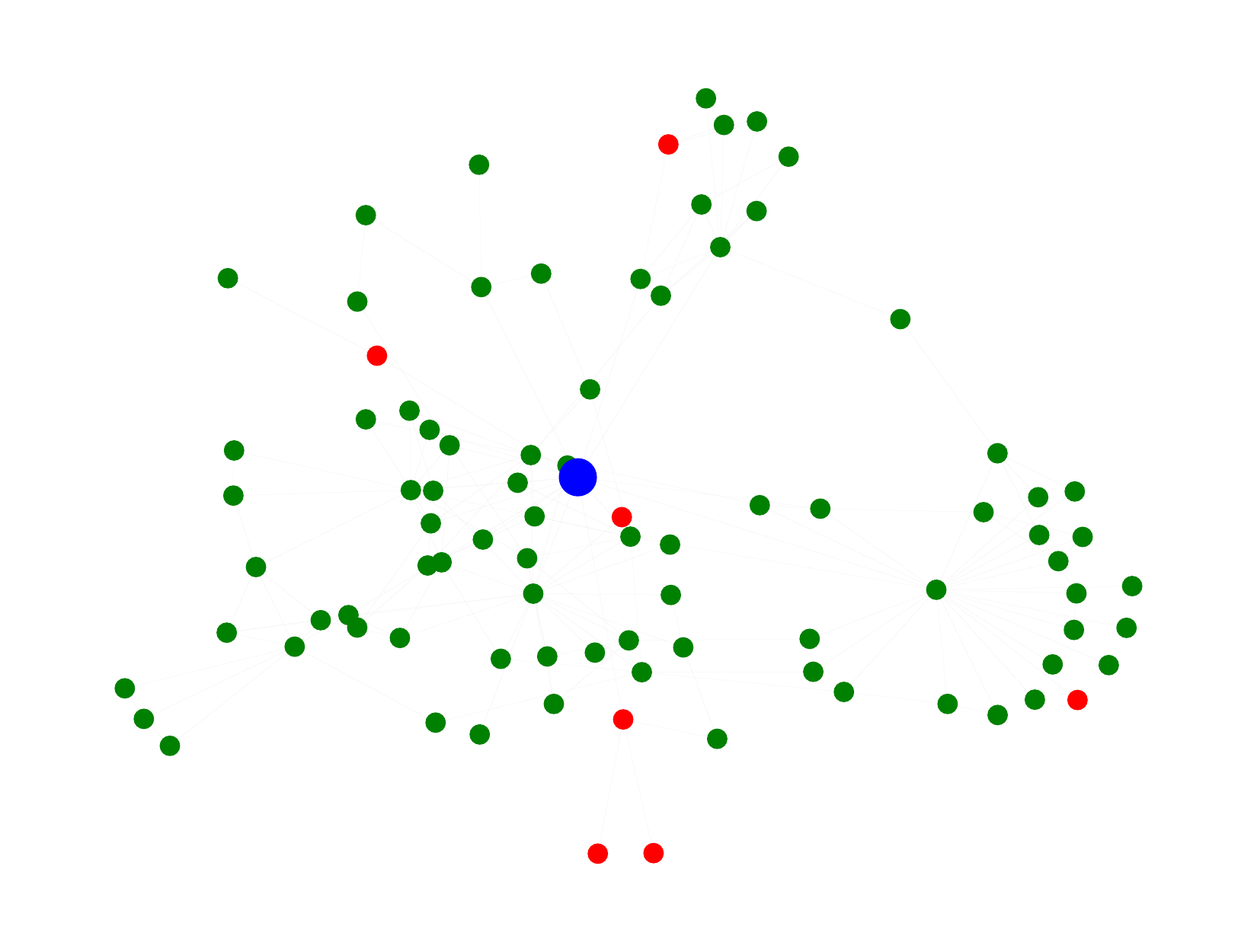}}
    \subfigure[]{%
        \includegraphics[width=0.32\textwidth]{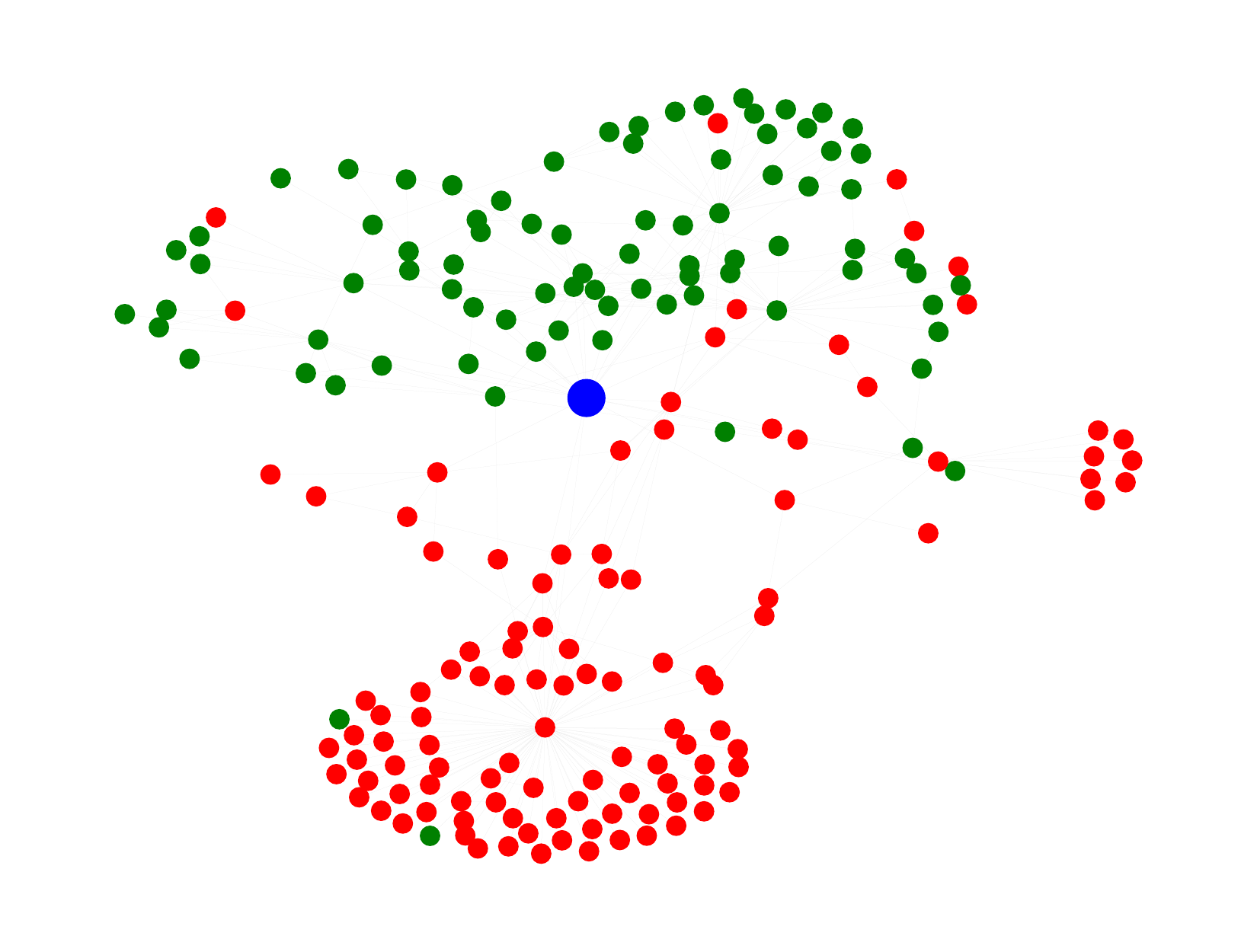}}
    \subfigure[]{%
        \includegraphics[width=0.32\textwidth]{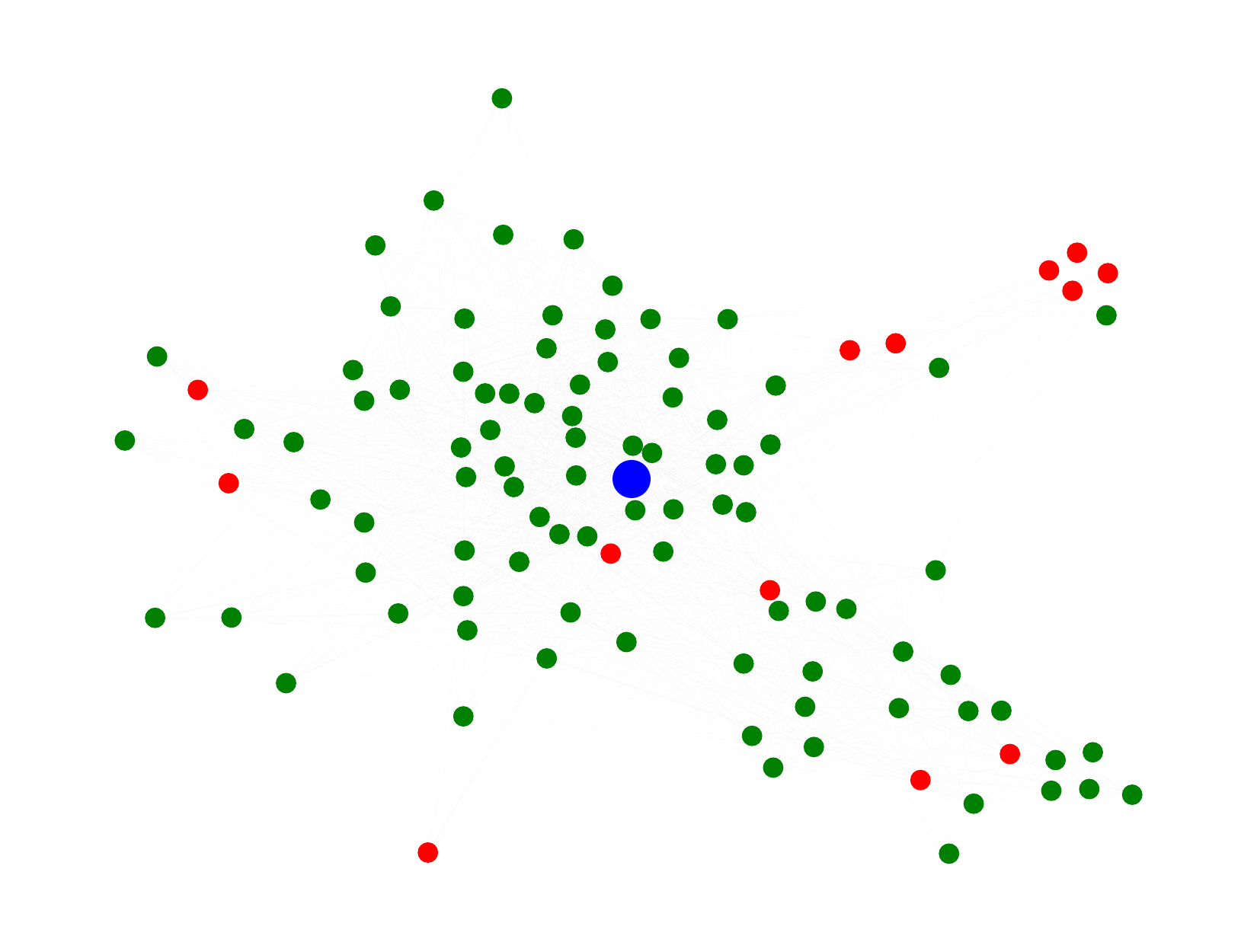}}
    \hfill
    \subfigure[]{%
        \includegraphics[width=0.32\textwidth]{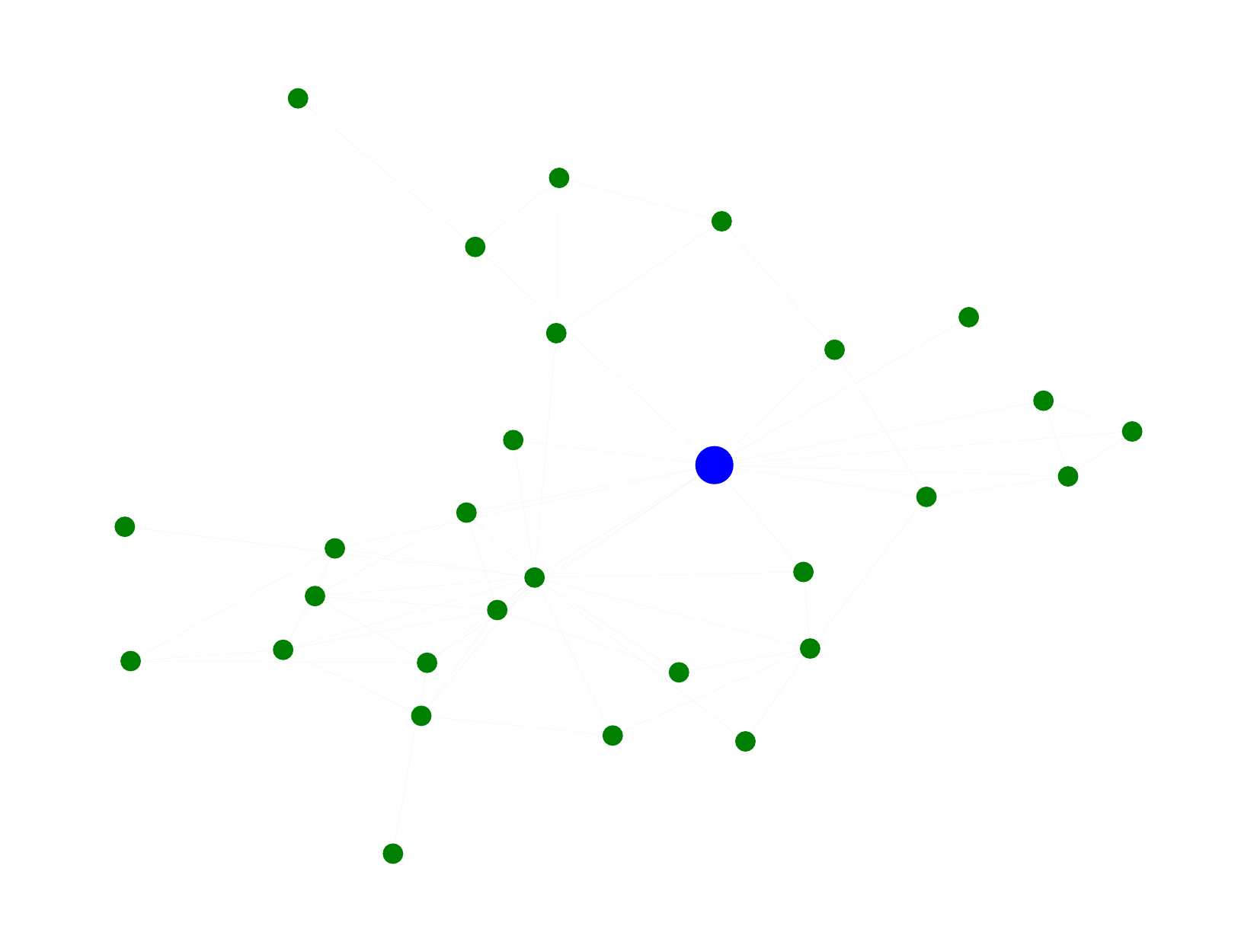}}
    \subfigure[]{%
        \includegraphics[width=0.32\textwidth]{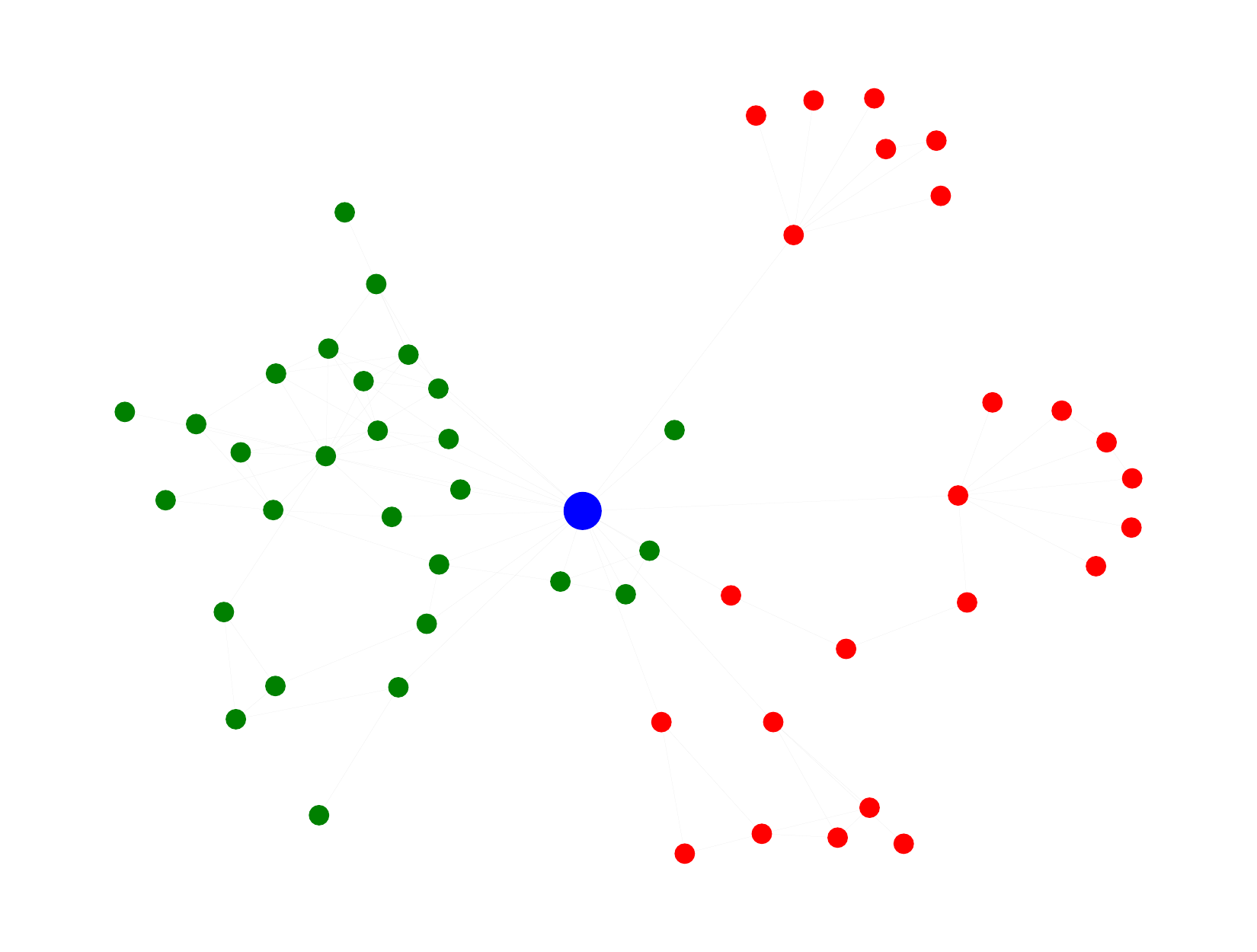}}
    \subfigure[]{%
        \includegraphics[width=0.32\textwidth]{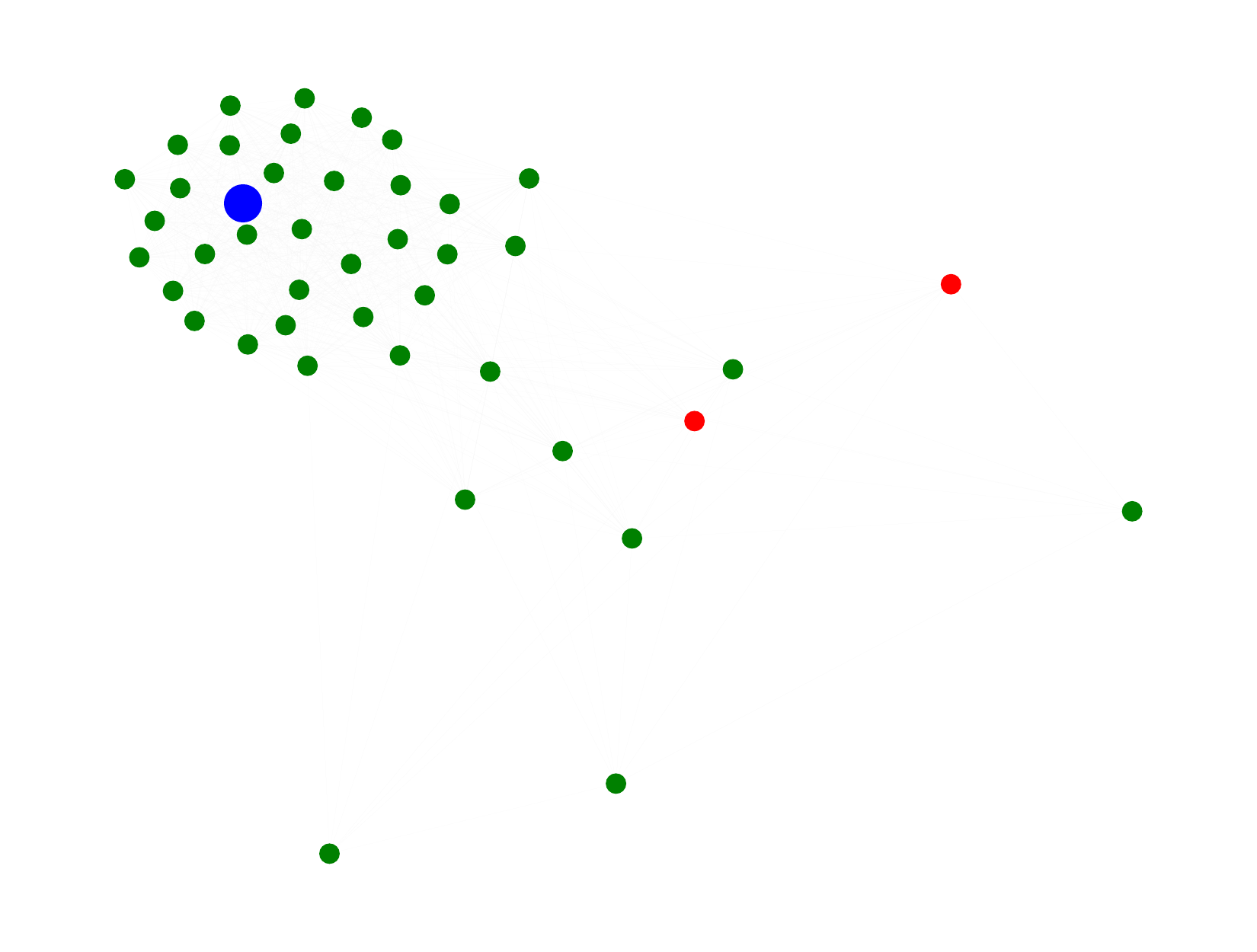}}
    %\vspace{-6pt}
    \caption{Cora visualizations on a target node (marked in blue) as well as its 1-hop and 2-hop neighbors. Neighbor nodes are marked in green if they have the same label as the target node, and red otherwise. Note that the three graphs in the same row share the same target node (randomly picked), while graphs in different rows focus on different target nodes. \textbf{Left}: clean graph. \textbf{Middle}: adversarial graph. \textbf{Right}: adversarial graph purified by \name.}
    %\vspace{-15pt}
    \label{figure:visual}
    %\vspace{-1.6cm}
\end{figure*}

%% file: table/garnet_homo_score.tex
\begin{table*}[ht!]
  \begin{center}
    \caption{Graph homophily score.}
    %\vspace{-5pt}
    \label{table:garnet_homo_score}
    \begin{adjustbox}{width=0.75\columnwidth,center}
    %{\renewcommand{\arraystretch}{1.1}
    %\setlength\tabcolsep{1.5 pt}
    \begin{NiceTabular}{lcc|cc}
      \toprule
 &  \multicolumn{2}{c}{Homophilic graphs} & \multicolumn{2}{c}{Heterophilic graphs}\\
\cmidrule(r){2-3} \cmidrule(r){4-5}
Dataset   &  Cora   &  Pubmed  &  Chameleon  &  Squirrel \\
\midrule
      Clean graph & $0.80$ & $0.80$ &  $0.23$ &  $0.22$ \\
      \name graph & $0.75$ & $0.72$ &  $0.25$ &  $0.26$ \\
      \bottomrule % <-- Bottomrule here
    \end{NiceTabular}
    \end{adjustbox}
  \end{center}
  %\vspace{-5pt}
\end{table*}

%% file: table/edge_recover.tex
\begin{table*}[ht!]
  \begin{center}
    \caption{Averaged recall and precision of clean structure recovery over $5$ (randomly picked) nodes.}
    %\vspace{-5pt}
    \label{table:edge_recover}
    \begin{adjustbox}{width=0.65\columnwidth,center}
    %{\renewcommand{\arraystretch}{1.1}
    %\setlength\tabcolsep{1.5 pt}
    \begin{NiceTabular}{lcc}
      \toprule
      &  Recall   &  Precision  \\
\midrule
      Cora (homophilic graph) & $0.94$ & $0.65$ \\
      Chameleon (heterophilic graph) & $0.87$ & $0.59$ \\
      \bottomrule % <-- Bottomrule here
    \end{NiceTabular}
    \end{adjustbox}
  \end{center}
  %\vspace{-5pt}
\end{table*}